\pdfoutput=1

\documentclass{article} 
\usepackage{iclr2027_conference,times}

\usepackage{amsmath,amsfonts,bm}

\def\eqref#1{equation~\ref{#1}}

\def\1{\bm{1}}

\DeclareMathAlphabet{\mathsfit}{\encodingdefault}{\sfdefault}{m}{sl}
\SetMathAlphabet{\mathsfit}{bold}{\encodingdefault}{\sfdefault}{bx}{n}

\PassOptionsToPackage{hyphens}{url} 
\usepackage{hyperref}
\usepackage{url}

\usepackage{enumitem}
\usepackage{amsmath, amssymb}
\usepackage{xcolor,colortbl}
\usepackage{soul}
\usepackage{booktabs}
\usepackage{comment}
\usepackage{subcaption}
\usepackage{multirow}

\usepackage{tikz}
\usepackage{pgfplots}
\usepackage{amsfonts}
\usepackage{wrapfig}

\usetikzlibrary{shapes.geometric, arrows.meta, positioning, fit, backgrounds, decorations.pathreplacing, shadows, calc, shapes.misc}

\newcommand{\finding}[3]{%
    \par\smallskip\noindent
    \colorbox{gray!12}{\parbox{\dimexpr\linewidth-2\fboxsep}{%
        \textbf{Finding #1: #2}\par\smallskip #3}}%
    \par\smallskip
}

\newcommand{\hypothesisbox}[1]{%
    \par\smallskip\noindent
    \colorbox{gray!12}{\parbox{\dimexpr\linewidth-2\fboxsep}{%
        \textbf{Central Hypothesis.}\quad\textit{#1}}}%
    \par\smallskip
}

\newcommand{\lm}[1]{\texttt{#1}}
\newcommand{\data}[1]{\textsf{#1}}
\newcommand{\affilsup}[1]{\rlap{\textsuperscript{\normalfont#1}}}

\title{Judge Circuits Explain Format-Induced \\Inconsistency in LLM-as-a-Judge}

\author{
    Nils Feldhus\affilsup{5,3}
    \qquad 
    Tanja Baeumel\affilsup{3,8}
    \qquad
    Elena Golimblevskaia\affilsup{4}
    \qquad
    Qianli Wang\affilsup{1}
    \\
    \textbf{Van Bach Nguyen}\affilsup{7}
    \qquad
    \textbf{Aaron Louis Eidt}\affilsup{1,4}
    \qquad
    \textbf{Selin Kahvecioglu}\affilsup{1}
    \\
    \textbf{Christopher Ebert}\affilsup{3}
    \quad
    \textbf{Wojciech Samek}\affilsup{1,2,4}
    \qquad
    \textbf{Jing Yang}\affilsup{1,2}
    \\
    \textbf{Vera Schmitt}\affilsup{1,3,6,8}
    \qquad
    \textbf{Sebastian M\"oller}\affilsup{1,3}
    \qquad
    \textbf{Simon Ostermann}\affilsup{3,8}
    \\
    $^1$ \small{Technische Universit\"at Berlin}
    \qquad
    $^2$ \small{BIFOLD – Berlin Institute for the Foundations of Learning and Data }
    \\
    $^3$ \small{German Research Center for Artificial Intelligence (DFKI)}
    \qquad
    $^4$ \small{Fraunhofer Heinrich Hertz Institute}
    \\
    $^5$ \small{CLCG, University of Groningen}
    \qquad
    $^6$ \small{Johannes Gutenberg University Mainz}
    \\
    $^7$ \small{Marburg University}
    \qquad
    $^8$ \small{Centre for European Research in Trusted AI (CERTAIN)}
    \\
    \footnotesize{ Correspondence: \texttt{n.feldhus@rug.nl} }
}

\iclrfinalcopy 
\begin{document}

\maketitle
\lhead{Preprint. Under review.}

\begin{abstract}
    \textbf{LLM-as-a-judge} has become the dominant paradigm for grading model outputs at scale, yet the same model assigns systematically different scores when its output format changes (e.g., a $1$--$5$ rating vs.~a True/False label).
    Existing diagnoses of these format-induced inconsistencies stop at the input-output level.
    Using Position-aware Edge Attribution Patching (PEAP), we causally investigate the internal mechanism in five open-weight instruction-tuned models (Gemma-3, Qwen2.5, Llama-3.1) across five judgment tasks.
    We find that judgments across structured understanding and open-ended preference tasks share a sparse \textit{Latent Evaluator} sub-graph in the mid-to-late layers;
    zero-ablating it collapses judgment while damaging knowledge probes substantially less than a random ablation of equal size in architecturally modular models.
    By structurally decoupling abstract judging from output formatting, we provide a mechanistic account of format-induced inconsistency on the open-weight models we study:
    a continuous judgment signal computed in the shared trunk is mapped through fragile, format-specific terminal branches. The judgment itself can therefore be read out independently of the requested output format.
    Our findings imply that benchmark comparisons of judge reliability across formats partly measure the fragile formatting stage, and can understate the quality of the underlying evaluation.

\end{abstract}

\section{Introduction}
\label{sec:intro}

The LLM-as-a-Judge (LaaJ) paradigm is now widespread across NLP for evaluation tasks such as benchmark scoring, reward modeling, and content moderation -- automating quality assessment without a human in the loop \citep{zheng-2023-judging-laaj, calderon-2025-alternative-annotator-test, gao-2025-llm-based-nlg-evaluation, li2024llmsasjudgescomprehensivesurveyllmbased}.
However, the reliability of LLMs as automated judges is heavily contested.
\citet{lee-2025-evaluating-consistency-llm-evaluators} document a dissociation -- relative preferences are often consistent, but absolute ratings are not -- and isolate two specific failure modes: self-consistency across repeated evaluations, and inter-scale consistency across different rating formats. 
The latter failure mode is the phenomenon this paper explains, and we can exhibit concretely: 
on the \data{STS-B} minimal pair in Figure~\ref{fig:worked_example}, Gemma-3-12B rates the semantic similarity of a matched sentence pair (\textit{a boy jumping into a lake}, described twice) at $3$ of $5$ under the rating format, yet answers ``No'' when the same instance is posed as a binary similar/dissimilar classification.
The content and the model are identical; only the requested output format changed, and the model's commitment changed with it.

Even large proprietary models show both failure modes, undermining the reproducibility of any LaaJ-driven leaderboard, reward, or safety judgment.
\citet{eshuijs-2025-short-circuiting-shortcuts} corroborate this from a different angle, showing that models frequently exploit shallow classification shortcuts (response length, sentiment polarity) and skip the integration of input and target aspects that holistic evaluation requires.
Comparable inconsistency and calibration failures hold for judges of $<70$B parameters \citep{girrbach-2025-reference-free-rating}.

No prior work has causally investigated the internal computational mechanisms underlying LLM judgment, a necessary step toward understanding and improving LaaJ reliability.
We recast the diagnostic question from \textit{``does the model judge consistently?''} to \textit{``where in the computational pathway does format-induced inconsistency originate?''}, 
and hypothesize that LaaJ implements judgment via two architecturally separable sub-systems -- a shared evaluation core and a format-specific output router -- and that the consistency failures in \citet{lee-2025-evaluating-consistency-llm-evaluators} localize to the latter.

\begin{wrapfigure}{r}{0.65\textwidth}
    \centering
    \includegraphics[width=\linewidth]{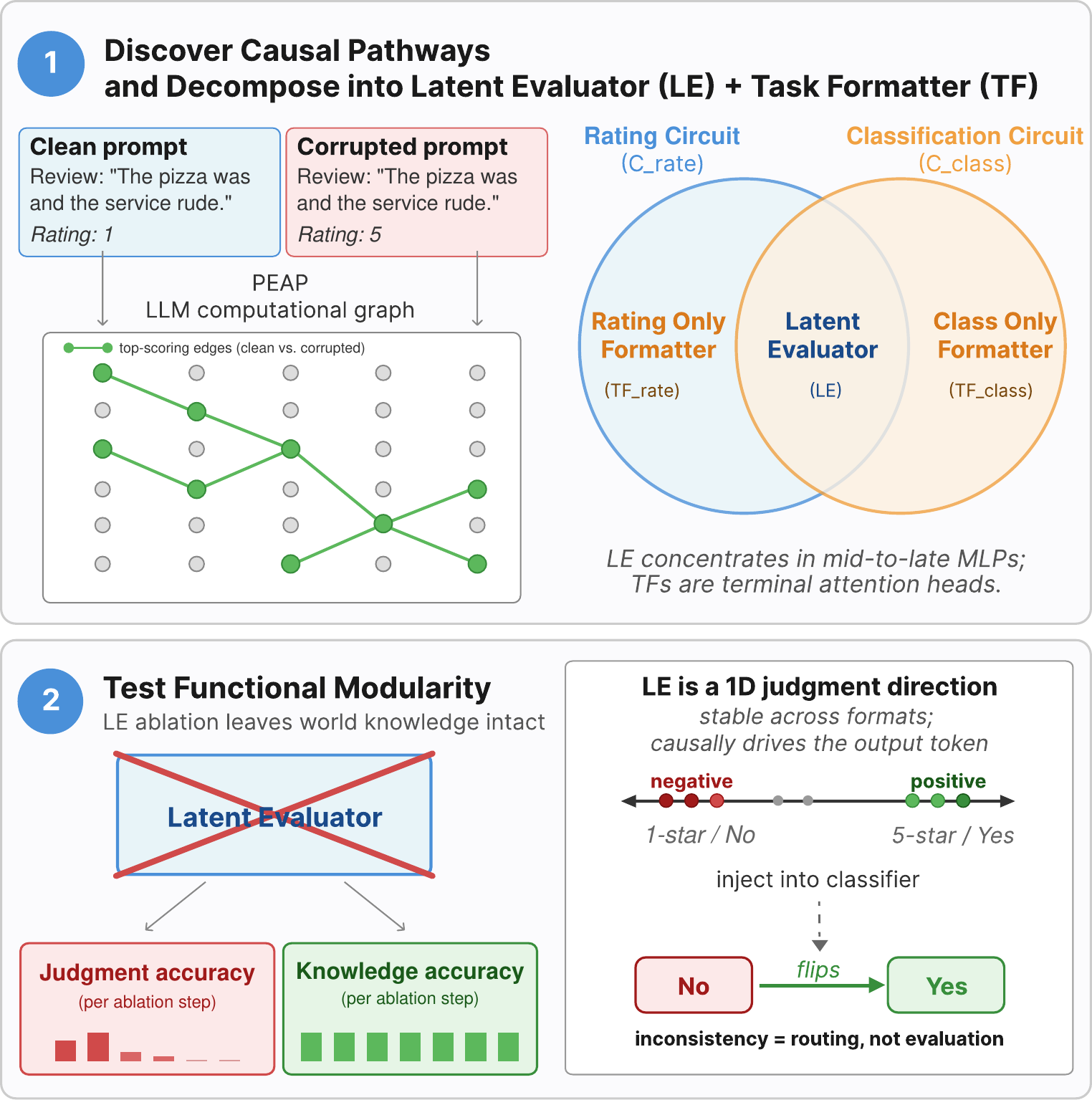}
    \caption{
        Overview on a \data{Yelp} minimal pair: (\textbf{1}) PEAP traces the cross-token causal edges that drive the difference between the clean and corrupted run (green) into a shared \textit{Latent Evaluator} sub-circuit ($\mathcal{C}_{\text{LE}} := \mathcal{C}_{\text{rate}} \cap \mathcal{C}_{\text{class}}$), leaving format-specific Task Formatters ($\mathcal{C}_{\text{TF,rate}}, \mathcal{C}_{\text{TF,class}}$) in the two circuits' disjoint parts. \\ 
        (\textbf{2}) Zero-ablation (red~$\boldsymbol\times$) separates evaluation from world knowledge: as more LE edges are ablated (bars), judgment accuracy on the rating tasks collapses while knowledge-probe accuracy (MMLU, factual QA) drops substantially less than under a random ablation of equal size; Boundless DAS (BDAS) finds a 1D judgment direction in the LE's activation space; Task Formatters in terminal layers map that judgment scalar to the target token.
    }
    \label{fig:overview}
\end{wrapfigure}

To test this, we use Position-aware Edge Attribution Patching (PEAP) \citep{haklay-2025-position-aware-automated-circuit-discovery}, which handles the cross-token edges needed for judge circuits whose inputs span separated linguistic spans (e.g., premise vs.~hypothesis) while remaining linear-in-edges to compute.
Drawing on the intermediate-variable \citep{lepori-2024-uncovering-intermediate-variables} and formal/functional dissociation \citep{hanna-2026-formal-dissociated} literature, we cross-validate every circuit with three independent causal probes -- cumulative edge patching, subspace steering, and cross-format activation transfer -- which converge on the same Latent Evaluator components and guard against non-identifiability \citep{miller-2024-transformer-circuit-evaluation-metrics, meloux-2025-everything-everywhere-all-at-once} (Figure~\ref{fig:overview}).

\noindent
\textbf{Contributions}:
\begin{enumerate}[topsep=1pt, partopsep=0pt, itemsep=0pt, leftmargin=0.7cm]
\renewcommand{\labelenumi}{(\theenumi)}
    \item \textbf{We show that LLM judgment is computed by highly sparse, cross-task circuits} sharing a generalized \textit{Latent Evaluator} in mid-to-late layers, recoverable at top-$k \leq 200$ edges.

    \item \textbf{We show that judgment modularity is architecture-dependent.}
        On modular models, zero-ablating the Latent Evaluator collapses judgment while damaging the knowledge probes we test substantially less than a size- and type-matched random-ablation control;
        Gemma-3-12B is the only model whose Latent Evaluator damages the verification probes beyond that control, while a smaller entanglement with pure factual recall appears in both Gemma-3 scales (\S \ref{sec:modularity_zero_ablation}).

    \item \textbf{We provide a mechanistic explanation of inter-format LLM evaluator inconsistency}, localizing it to the format-specific output routing while the underlying evaluation stays stable.
\end{enumerate}

Together, these results suggest that LaaJ format inconsistency is a routing problem, so fixes can target the formatter and leave the model's judgment competence untouched. \footnote{Code and data available at: \url{https://github.com/nfelnlp/JuCi}}

\section{Experimental Setup}

\hypothesisbox{An LLM-as-a-judge implements judgment via two architecturally separable sub-systems -- a shared evaluation core and a format-specific output router.}
We test this in three steps: \S\ref{sec:discovery} discovers the candidate sub-circuits; \S\ref{sec:modular} probes whether the shared core is functionally isolated; \S\ref{sec:inconsistency} causally validates the split via cross-format activation transfer.

A \textit{judgment task} in our setting asks the model to assign a quality, preference, or correctness score to a candidate text given the input it conditions on.
The output is a scalar rating or a categorical verdict over the candidate. 
Our pipeline operates on contrastive minimal-pair prompts (Figure~\ref{fig:overview});
the rating-vs-classification decomposition into a \textit{Latent Evaluator} and format-specific \textit{Task Formatters} is introduced in \S\ref{sec:contrastive}.

\paragraph{Data}
\label{sec:setup_data}
We select five datasets that together span the three dimensions of evaluation that LaaJ is deployed for: (i)~structured linguistic correctness (\data{CoLA}, \data{MultiNLI}, \data{STS-B}), (ii)~preference / quality judgment (\data{RewardBench}), and (iii)~subjective sentiment (\data{Yelp}).
\begin{itemize}[noitemsep,leftmargin=*]
    \item \data{CoLA} (linguistic acceptability) (\citeauthor{warstadt-2019-cola}): fluency and grammaticality as quality criteria.
    \item \data{MultiNLI} (natural language inference) \citep{williams-2018-mnli}: entailment / neutral / contradiction between a hypothesis and a premise.
    \item \data{STS-B} (sentence semantic similarity) \citep{cer-etal-2017-semeval}: semantic equivalence between pairs.
    \item \data{RewardBench} (preference evaluation) \citep{lambert-2025-rewardbench}: the canonical testbed for open-ended LLM-as-a-judge capabilities.
    \item \data{Yelp} (sentiment, 1--5 star reviews) \citep{zhang-2015-character}: a subjective, user-written evaluation domain with a natural ordinal scale.
\end{itemize}

\paragraph{Models}
\label{sec:setup_models}
We evaluate five instruct-tuned models from three families: Gemma-3 (12B-it, 27B-it) \citep{gemma-team-2025-gemma-3}, Qwen2.5 (7B-Instruct, 14B-Instruct) \citep{qwen-2025-qwen-25}, and Llama-3.1-8B-Instruct \citep{grattafiori-2024-llama-3}, accessed via TransformerLens \citep{nanda-2022-transformerlens}.
We cap the minimal-pair subset at $|S| = 500$ for \data{MNLI}; \data{CoLA}, \data{STS-B}, \data{RewardBench}, and \data{Yelp} have $100$ -- $200$ valid semantic pairs each.
Within-task circuit IoU is comparable across these subset sizes (split-half reliability, App.~\ref{app:split_half}); 
the compute constraints behind the cap and our backward-pass tracing budget are deferred to App.~\ref{app:minimal_pairs}.

\paragraph{Prompt design}
\label{sec:setup_prompts}
For each dataset we construct contrastive \textit{minimal pairs}: a clean prompt (correct rating) and a corrupted prompt (incorrect rating) with matched token lengths for PEAP attribution\footnote{For \data{MNLI}, pairs are drawn from the {entailment, contradiction} subset so that clean and corrupted prompts have semantically opposed ground truth (App.~\ref{app:minimal_pairs}).}.
Half the pairs assign the higher rating to the clean prompt and half to the corrupted prompt, so per-edge attributions are symmetric by construction (\S\ref{sec:method_patching}). 
We format every input as a $1$--$5$ rating prompt; for contrastive circuit analysis (\S\ref{sec:contrastive}), we pair each dataset with a parallel classification-control prompt (categorical Yes/No, True/False, or Entailment/Contradiction labels) on the same instances. 
Templates and alignment details are in Appendices~\ref{app:prompts}--\ref{app:minimal_pairs}.

\section{Discovering Judge Circuits in LLMs}
\label{sec:discovery}
We use \textit{judge circuit} to refer to the sparse causal sub-circuit a model uses to compute a rating from a structured prompt; \S\ref{sec:contrastive} decomposes it into a shared evaluation core ($\mathcal{C}_{\text{LE}}$) and a format-specific output branch ($\mathcal{C}_{\text{TF}}$).
Our two-stage pipeline first applies PEAP to identify the causal pathways responsible for evaluation,
then isolates task-specific formatting mechanisms from generic evaluation logic using contrastive control tasks.

\subsection{Circuit Discovery via PEAP}
\label{sec:method_patching}
Circuit discovery in decoder-only LLMs conceptualizes the forward pass as a computation graph $\mathcal{G}$ whose nodes are multi-layer perceptrons (MLPs) and attention heads and whose directed edges carry information flow, and seeks a sparse subgraph $\mathcal{C} \subset \mathcal{G}$ that causally accounts for a target behavior \citep{vig-2020-causal-mediation-analysis, conmy-2023-acdc, wang-2023-interpretability-in-the-wild}. 
\textbf{Position-aware Edge Attribution Patching} (PEAP) \citep{haklay-2025-position-aware-automated-circuit-discovery} extends Edge Attribution Patching \citep{hanna-2024-have-faith-in-faithfulness} to capture causal edges \textit{across token positions} in addition to intra-token ones -- a necessary property for judge circuits that must cross-reference separated linguistic spans (e.g., premise vs.~hypothesis).
Concretely, for each candidate edge from sender $S$ to receiver $R$, PEAP estimates causal importance by the dot product of the receiver's gradient $\nabla R$ with the difference between the sender's activation on the clean and corrupted inputs $(S_{\text{clean}} - S_{\text{corr}})$. 
Attribution is taken with respect to the model's \textit{expected rating value}, $\mathrm{EV} = \sum_{r} r \cdot P(\text{rating}=r)$ -- the probability-weighted mean of the rating-token distribution at the final position (Appendix~\ref{app:peap_formulas}).
The clean--corrupted \textit{EV gap} of a minimal pair is the causal quantity that our faithfulness and steering analyses track.
A single backward pass yields all receiver gradients, so the entire ranked edge list over attention heads and MLPs is extracted in one forward--backward sweep per minimal pair.
We extend PEAP with a symmetric polarity correction (full formulas in Appendix~\ref{app:peap_formulas}) that handles our bidirectional minimal pairs (\S\ref{sec:setup_prompts}) without canceling causal signal under na\"{i}ve gradient summation.
The extracted circuits are faithful to the full model (Appendix~\ref{app:faithfulness}), stable under data resampling (Appendix~\ref{app:split_half}), and stable under prompt-instruction paraphrases (Appendix~\ref{app:paraphrase_robustness}).

\subsection{Structural Overlap: The Latent Evaluator}
\label{sec:overlap}

Cross-task structural overlap is established evidence of shared computation in transformer circuits \citep{tigges-2024-circuit-analyses-consistent, ferrando-costa-jussa-2024-similarity-circuits-across-languages, lan-2024-towards-interpretable-sequence-continuation}.
Given two circuits $\mathcal{C}_A, \mathcal{C}_B$ traced on different tasks $A$ and $B$ and pruned to their top-$k$ edges, we quantify similarity via Jaccard Intersection-over-Union on both the set of unique edges $\mathcal{E}$ and distinct components $\mathcal{N}$, abstracting away token positions:
\begin{equation*}
\text{IoU}_{\text{edge}} = \frac{|\mathcal{E}_A \cap \mathcal{E}_B|}{|\mathcal{E}_A \cup \mathcal{E}_B|},
\quad
\text{IoU}_{\text{node}} = \frac{|\mathcal{N}_A \cap \mathcal{N}_B|}{|\mathcal{N}_A \cup \mathcal{N}_B|}.
\end{equation*}
Edge IoU is the stricter metric; Node IoU measures architectural recruitment at a coarser grain.

\begin{figure}[t]
    \centering
    \includegraphics[width=\textwidth]{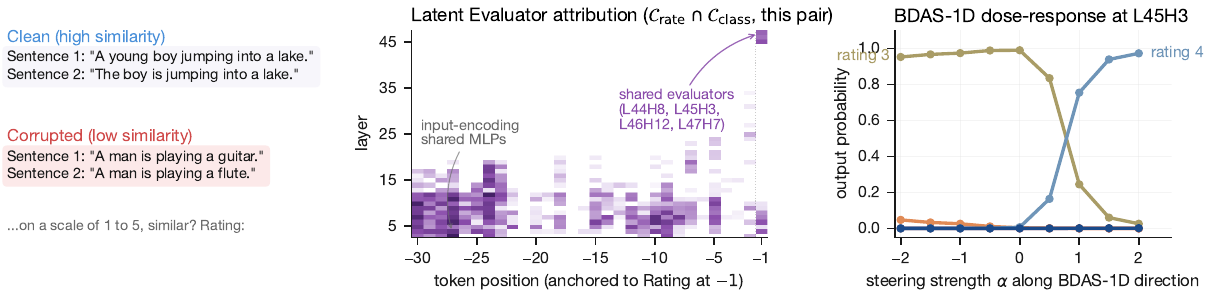}
    \caption{
        Anatomy of judgment on one \data{STS-B} pair (Gemma-3-12B, \#149).
        \textbf{Left}: clean (high-sim.) and corrupted (low-sim.) prompts.
        \textbf{Middle}: PEAP attribution restricted to $\mathcal{C}_{\text{LE}} := \mathcal{C}_{\text{rate}} \cap \mathcal{C}_{\text{class}}$ (\S\ref{sec:contrastive}, top-$1000$ aggregate): early shared input-encoding MLPs plus late shared evaluators (App.~\ref{app:head_analysis}).
        \textbf{Right}: steering along BDAS-1D at L45H3 (App.~\ref{app:bdas}), $\alpha$ calibrated to one pair's worth of natural causal signal, flips the output from rating $3$ to rating $4$ with a smooth crossover at $\alpha\!\approx\!0.5$--$1.0$.
    }
    \label{fig:worked_example}
\end{figure}

\finding{1}{Distinct judgment tasks share a dense trunk on every modular architecture.}{}
On Gemma-3-12B at top-$200$, the circuits of any two of our tasks share about half or more of their components, including pairings of the structured tasks with the open-ended \data{RewardBench} (Figure~\ref{fig:circuit_overlap}; component-level numbers in Appendix~\ref{app:node_overlap}).
The same shared-trunk pattern holds across the modular models, with Qwen2.5-7B showing the strongest edge-level agreement on every task pair we test (Figure~\ref{fig:cross_model_iou}).
Qwen2.5-14B and Gemma-3-27B agree less on specific edges but still recruit largely the same components.
This matches the scale-dependent redundancy documented in Appendix~\ref{app:split_half}: larger modular models route judgment through several equivalent sub-pathways, so the same components appear while the top edges vary across data splits.
Within-task split-half overlap on Gemma-3-12B at the same sparsity meets or exceeds the cross-task overlap (Appendix~\ref{app:split_half}), so sample-size noise does not produce the shared trunk.
A layer-wise decomposition localizes the cross-task overlap to the mid-to-late layers across the evaluated models (App.~\ref{app:layerwise_iou}), matching the LE/TF depth split developed in \S\ref{sec:contrastive}.

\begin{figure*}[t]
    \centering
    \includegraphics[width=\textwidth]{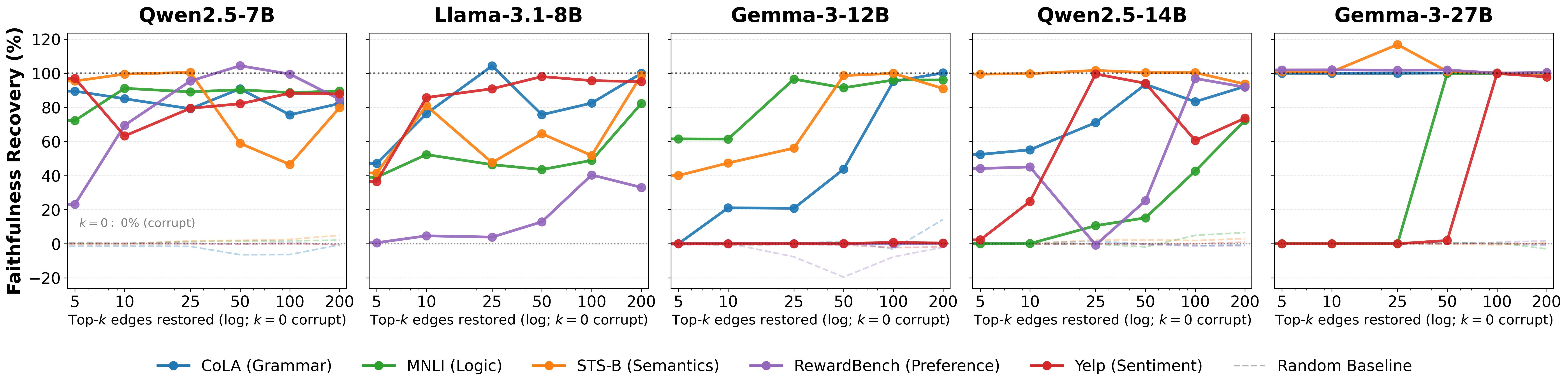}
    \caption{
        Sparse circuit faithfulness across the five evaluated models and five rating tasks. Each curve traces median MIB recovery as we cumulatively patch the top-$k$ PEAP edges from a fully corrupted forward pass back toward the clean activations. Solid colored lines are the discovered circuits; the gray dashed line is a random-edge baseline. Curves saturating at $\approx 1.0$ at small $k$ indicate that the sparse circuit fully captures the model's evaluation behavior; flat curves (Gemma-3-12B / \data{RewardBench}, \data{Yelp}; Llama-3.1-8B / \data{Yelp}) reflect architectural entanglement on those particular cells.
    }
    \label{fig:faithfulness_master}
\end{figure*}

\subsection{Sparse Circuit Faithfulness}
\label{sec:faithfulness}

To validate that the PEAP-discovered edges are causally \textit{sufficient} for the model's judgment, we apply the per-instance faithfulness metric of the Mechanistic Interpretability Benchmark (MIB) \citep{mueller-2025-mib}: starting from a fully corrupted forward pass, we progressively restore the top-$k$ PEAP edges and measure the median fraction of the clean--corrupted EV gap (\S\ref{sec:method_patching}) that the patched sub-circuit recovers (full methodology and the magnitude-weighted sensitivity analysis are in Appendices~\ref{app:faithfulness} and \ref{app:faithfulness_pooled}).

\finding{2}{PEAP recovers highly sparse, faithful circuits across models and tasks.}{}
In $21$ of the $25$ (model, task) combinations we trace, restoring at most $200$ edges recovers nearly all of the judgment signal (Figure~\ref{fig:faithfulness_master}); on Gemma-3-27B the open-ended \data{RewardBench} circuit saturates with just five edges.
The remaining combinations are Gemma-3-12B on \data{RewardBench} and \data{Yelp}, where recovery stays near zero and matches that model's entanglement profile (\S\ref{sec:modularity_zero_ablation}, Table~\ref{tab:mmlu_modularity}), and Llama-3.1-8B on \data{MNLI} (a slower climb) and \data{Yelp} (a partial peak that drifts back down).
A random-edge baseline (gray dashed line) stays near zero in every configuration, so sparsity alone does not explain the recovery.

\paragraph{Cross-method robustness}
Different attribution algorithms can select different sparse subgraphs on the same model and task \citep{meloux-2025-everything-everywhere-all-at-once}, so we re-trace every Qwen2.5-7B and Gemma-3-12B circuit with LRPEAP, a variant we develop that keeps PEAP's position-aware edge formulation but computes the backward pass with LRP rules \citep{jafari-2025-relp}.
Both backbones recover the same mid-to-late Latent Evaluator far above the permutation null, and a comparison against ACDC points the same way (Appendix~\ref{app:method_independence}).
The circuits are also robust to prompt wording: re-traced on paraphrased instructions, they overlap the originals about as much as two halves of the same data do, and only renaming the prompt's field labels cuts noticeably into that overlap (Appendix~\ref{app:paraphrase_robustness}).

\section{Judge Circuit Modularity is Architecture-Dependent}
\label{sec:modular}

Building on \S\ref{sec:overlap}, we test whether the shared trunk is a functionally modular sub-system or merely a generic capability bottleneck \citep{hanna-2026-formal-dissociated}: 
if zero-ablating the Latent Evaluator collapses judgment but mostly spares world-knowledge benchmarks, the sub-graph is doing judgment-specific work.
This in turn licenses the format-transfer experiments in \S\ref{sec:inconsistency} as targeted perturbations.

\begin{table*}[t]
    \centering
    \caption{
        Zero-ablation knowledge-probe domain control. 
        Each cell reports accuracy (\%) as base $\rightarrow$ LE-ablated, with a size-matched random-ablation control (mean $\pm$ SD over 3 trials) in parentheses; the \textbf{LE} column gives the composition of the ablated merged Latent Evaluator (attention heads + MLPs), merged across the per-task Latent Evaluators of \data{CoLA}, \data{MNLI}, and \data{STS-B} (App.~\ref{app:knowledge_probes}).
        All cells ablate the full LE (heads and MLPs); \textit{entanglement} requires LE-ablation damage \textit{exceeding} the random control.
        On the four modular models, LE ablation is no more damaging than random on \data{MMLU} and formal factual verification (\data{StrategyQA}, \data{CREAK}).
        Gemma-3-12B is the only model whose LE ablation falls below the control (bold), mostly due to its three LE attention heads (\S\ref{sec:modularity_zero_ablation}).
        The full circuit-topology panels in Appendix~\ref{app:peap_circuit_topology} demonstrate the corresponding two-stage Latent Evaluator / Task Formatter separation across models and tasks, supporting the same generalization.
    }
    \resizebox{\textwidth}{!}{%
        \begin{tabular}{ll|ccc|cc}
            \toprule
            \textbf{Model} & \textbf{LE} & \textbf{Clinical DB} & \textbf{Abstract Alg.} & \textbf{Physics} & \textbf{StrategyQA} & \textbf{CREAK} \\
            \midrule
            Qwen2.5-7B & 1H+9M & 78 $\rightarrow$ 48 {\scriptsize(20.3\,$\pm$\,4.5)} & 56 $\rightarrow$ 25 {\scriptsize(20.7\,$\pm$\,1.2)} & 53 $\rightarrow$ 38 {\scriptsize(23.0\,$\pm$\,1.4)} & 66.5 $\rightarrow$ 58.0 {\scriptsize(47.2\,$\pm$\,3.9)} & 81.5 $\rightarrow$ 84.0 {\scriptsize(48.0\,$\pm$\,1.1)} \\
            Llama-3.1-8B & 0H+8M & 76 $\rightarrow$ 47 {\scriptsize(22.3\,$\pm$\,3.1)} & 35 $\rightarrow$ 29 {\scriptsize(21.7\,$\pm$\,2.1)} & 41 $\rightarrow$ 24 {\scriptsize(21.0\,$\pm$\,3.6)} & 67.0 $\rightarrow$ 61.5 {\scriptsize(52.5\,$\pm$\,1.1)} & 78.5 $\rightarrow$ 47.5 {\scriptsize(48.7\,$\pm$\,1.3)} \\
            Gemma-3-12B & 3H+4M & 79 $\rightarrow$ \textbf{18} {\scriptsize(55.7\,$\pm$\,7.8)} & 47 $\rightarrow$ \textbf{22} {\scriptsize(27.7\,$\pm$\,3.7)} & 52 $\rightarrow$ \textbf{22} {\scriptsize(28.0\,$\pm$\,4.3)} & 65.0 $\rightarrow$ 62.5 {\scriptsize(61.3\,$\pm$\,3.0)} & 80.5 $\rightarrow$ 79.5 {\scriptsize(76.8\,$\pm$\,3.2)} \\
            Qwen2.5-14B & 0H+2M & 88 $\rightarrow$ 85 {\scriptsize(63.0\,$\pm$\,31.2)} & 59 $\rightarrow$ 57 {\scriptsize(49.3\,$\pm$\,13.7)} & 69 $\rightarrow$ 64 {\scriptsize(56.0\,$\pm$\,14.9)} & 71.5 $\rightarrow$ 71.5 {\scriptsize(61.8\,$\pm$\,10.7)} & 87.0 $\rightarrow$ 81.0 {\scriptsize(74.8\,$\pm$\,14.1)} \\
            Gemma-3-27B & 0H+3M & 79 $\rightarrow$ 82 {\scriptsize(80.0\,$\pm$\,2.9)} & 47 $\rightarrow$ 47 {\scriptsize(44.7\,$\pm$\,4.0)} & 62 $\rightarrow$ 59 {\scriptsize(56.7\,$\pm$\,1.2)} & 50.5 $\rightarrow$ 52.5 {\scriptsize(48.8\,$\pm$\,1.9)} & 82.5 $\rightarrow$ 82.5 {\scriptsize(81.0\,$\pm$\,3.2)} \\
         \bottomrule
    \end{tabular}
    }
    \label{tab:mmlu_modularity}
\end{table*}

\subsection{Isolating Judgment from Formatting via Contrastive Circuits}
\label{sec:contrastive}

For each dataset we trace two circuits on the same data: one for the rating task ($\mathcal{C}_{\text{rate}}$, e.g., ``On a scale of $1$ to $5$\dots'') and one for a classification control task ($\mathcal{C}_{\text{class}}$, e.g., yes/no) with matched prompt structure.
Their structural overlap decomposes the model's judgment computation into two functionally distinct components:
\begin{itemize}[noitemsep,leftmargin=*]
    \item \textbf{The Latent Evaluator} ($\mathcal{C}_{\text{LE}} := \mathcal{C}_{\text{rate}} \cap \mathcal{C}_{\text{class}}$): 
        the shared computational trunk. Components in this intersection process the core semantic judgment of the prompt, agnostic to output format.
        $\mathcal{C}_{\text{LE}}$ is the sub-circuit highlighted as the Latent Evaluator in Figure~\ref{fig:overview}.
    \item \textbf{The Task Formatters} ($\mathcal{C}_{\text{TF,rate}} := \mathcal{C}_{\text{rate}} \setminus \mathcal{C}_{\text{class}}$ and $\mathcal{C}_{\text{TF,class}} := \mathcal{C}_{\text{class}} \setminus \mathcal{C}_{\text{rate}}$): 
        specialized terminal routing branches, typically late-layer attention heads, that translate the abstract judgment into format-specific target tokens.
\end{itemize}
The judge circuit (\S\ref{sec:discovery}) is therefore $\mathcal{C}_{\text{rate}} = \mathcal{C}_{\text{LE}} \cup \mathcal{C}_{\text{TF,rate}}$. We abbreviate Latent Evaluator and Task Formatter as LE and TF.

\finding{3}{Contrastive tracing yields a clean Latent Evaluator / Task Formatter split.}{}
On Gemma-3-12B (\data{CoLA}\,$\times$\,\data{CoLA\_CLASS}, top-$200$), $3$ of $17$ analyzed heads act as shared evaluators (most strongly L45H3, L46H12, L47H7), while the remaining $14$ split without overlap into rating-specific ($9$) and classification-specific ($5$) formatters;
Figure~\ref{fig:worked_example} (middle) visualizes the same $\mathcal{C}_{\text{LE}}$ on one \data{STS-B} pair.
Three cross-checks support this decomposition: per-head attribution analysis (App.~\ref{app:head_analysis}), transcoder features at the circuit's MLP positions (App.~\ref{app:transcoder}), and a layer-wise Edge IoU decomposition (App.~\ref{app:layerwise_iou}) all confirm the shared-then-specialized two-stage pattern.

\subsection{Functional Modularity via Zero-Ablation}
\label{sec:modularity_zero_ablation}

Identifying a shared causal circuit does not guarantee that the Latent Evaluator is \textit{functionally} isolated from unrelated capabilities such as world-knowledge recall.
We test this by zero-ablating the Latent Evaluator: for every component (attention head or MLP) that appears as a sender in at least one top-$k$ Latent Evaluator edge, we clamp its forward-pass output to zero.
We then evaluate the ablated model against \data{MMLU} \citep{hendrycks-2021-measuring} world knowledge and two factual QA datasets, \data{StrategyQA} \citep{geva-2021-strategyqa} and \data{CREAK} \citep{onoe-2021-creak}.
These probes mirror our Task Formatter answer tokens (``Yes/No'', ``True/False''), while relying predominantly on factual retrieval with little abstract quality judgment.
Because zeroing several mid-network MLPs is destructive regardless of their function, every ablation is compared against an equally large random set of heads and MLPs sampled uniformly outside the LE over 3 trials. We say the LE is \textit{entangled} with a probe only if LE-ablation damage \textit{exceeds} the random control.

\finding{4}{On modular architectures, ablating the Latent Evaluator damages knowledge probes substantially less than random.}{}
Table~\ref{tab:mmlu_modularity} illustrates that, on the four modular models, LE ablation is less damaging than the random control on every probe but one (Llama-3.1-8B on \data{CREAK}).
By contrast, iteratively ablating Latent Evaluator edges in the same models triggers a phase-transition collapse in judgment EV on every rating task tested (Appendix~\ref{app:ablation_study}, Figures~\ref{fig:ablation_study_qwen25_7}--\ref{fig:ablation_study_qwen25_14}). The Latent Evaluator therefore operates as a specialized sub-system: removing it destroys judgment while damaging the knowledge probes less than removing an equally sized random set of components.

\finding{5}{Modularity emerges at family-specific scales.}{}
Gemma-3-12B is the only model whose LE ablation damages the verification probes substantially beyond the random control: full-LE ablation drops \data{MMLU} clinical knowledge to $18\%$ against a random control (Table~\ref{tab:mmlu_modularity}). 
Ablating only its three LE attention heads (L45H3, L46H12, L47H7) already costs $14$--$18$ pp on all three \data{MMLU} subsets (clinical $79\% \rightarrow 63\%$; three random heads: no effect), while leaving \data{StrategyQA}, \data{CREAK}, and pure recall untouched.
Two follow-up experiments (App.~\ref{app:knowledge_probes}) reveal what this entanglement consists of.
First, a \textit{pure-recall cloze probe} (declarative \data{PopQA} completions with no candidate answers and no verification step) shows a recall drop under LE ablation on Gemma-3-12B and 27B, but never on Qwen or Llama.
Second, \textit{factual-recall reference circuits} (\data{CREAK}, \data{StrategyQA}) traced with the identical PEAP pipeline capture most of the entangled model's LE but far less of the modular control's; this contrast is preserved at every threshold and sample size.

\section{Inter-Format Inconsistencies Arise from Format-Specific Output Routing}
\label{sec:inconsistency}

Given that the Latent Evaluator is a functionally modular sub-system, the question is how its output is transformed into the format-specific target token.
Our hypothesis localizes inter-format inconsistency to the \textit{Task Formatter} branches (\S\ref{sec:contrastive}): the Latent Evaluator computes a stable continuous judgment signal mapped onto format-specific tokens by fragile, non-linear terminal routing.
We test this via causal cross-format patching.

\paragraph{Causal Analysis via Format Transfer Injection}
We design a minimal causal test: \textit{Format Transfer Injection} (FTI) following \citet{merullo-2024-word2vec-style}.
For a given instance we capture the activations of the Latent Evaluator components during a pristine \textit{$5$-star rating} prompt, then inject those exact activations into the same model while it runs on a corrupted \textit{classification} prompt (whose natural output would be the negative token, e.g., ``No'').
This is a \textit{blanket} activation transfer: it overwrites the entire LE pattern, in contrast to the targeted $1$D subspace interventions of Appendix~\ref{app:bdas}.
If the Latent Evaluator carries the judgment, the downstream classification head should receive the injected positive signal and flip its output token from ``No'' to ``Yes'' or ``Entailment''.
If instead the terminal branches do the actual judgment work, the injection should have no effect.
\S\ref{sec:synthesis} returns to the contrast between this whole-pattern transfer and the targeted interventions.

\subsection{The Latent Evaluator Carries the Judgment}
\label{sec:fti_results}

\finding{6}{The Task Formatter decides whether the judgment reaches the output.}{}

\begin{wraptable}{r}{.575\textwidth}
    \centering
    \vspace*{-1em}
    \caption{
        FTI flip rates (\%): how often injecting the Latent Evaluator activations of a $5$-star rating prompt into a corrupted classification prompt flips the verdict to the positive label (Yes/Entailment). $^{\ast}$~marks cells with few valid pairs ($N \leq 26$). Full results with probability shifts, pair counts, and Wilson 95\% CIs are in Table~\ref{tab:fti_full} (App.~\ref{app:fti_full}).
    }
    \label{tab:fti_results}
    \resizebox{.575\textwidth}{!}{%
    \begin{tabular}{lccccc}
        \toprule
        \textbf{Model} & \data{CoLA} & \data{MNLI} & \data{STS-B} & \data{Yelp} & \data{RewardBench} \\
        \midrule
        Qwen2.5-7B   & 100.0$^{\ast}$ & \textbf{99.1} & \textbf{100.0} & 66.4 & \textbf{100.0} \\
        Llama-3.1-8B & 25.0$^{\ast}$ & 4.2 & -- & 0.0 & 22.2$^{\ast}$ \\
        Gemma-3-12B  & 14.9 & 3.7 & 50.0 & 0.0 & 7.7$^{\ast}$ \\
        Qwen2.5-14B  & 91.7$^{\ast}$ & 4.0 & 5.1 & 31.2 & 41.2 \\
        Gemma-3-27B  & 0.0 & 0.7 & \textbf{96.8} & 1.1 & 5.3$^{\ast}$ \\
        \bottomrule
    \end{tabular}
    }
    \vspace*{-1em}
\end{wraptable}

The clearest causal demonstration is Qwen2.5-7B: blanket FTI flips the argmax in $\geq 99\%$ of \data{MNLI}, \data{STS-B}, and \data{RewardBench} pairs (Table~\ref{tab:fti_results}), with mean target-class probability rising from $\leq 17\%$ at baseline to $\geq 85\%$ post-injection.
Near-total flips hold on Gemma-3-27B / \data{STS-B} and Qwen2.5-14B / \data{CoLA}.
The injected continuous judgment scalar is decoded by the classification formatter onto the discrete target token ``Yes''/``Entailment'', never leaving the output format's vocabulary (no pair emits ``$5$'').
The judgment representation is stable, $1$D, and shared across semantic domains (\S\ref{sec:contrastive}, App.~\ref{app:bdas}).
The bottleneck is the terminal mapping, which varies sharply across tasks (three-way \data{MNLI} against binary classification) and across models: some formatters accept the injected pattern (e.g., Qwen2.5-7B) and some do not (e.g., Gemma-3-27B on \data{MNLI}).
A failed injection could also mean that the Latent Evaluator's pattern is itself format-specific there. The cross-domain steering at the same components (App.~\ref{app:bdas}) and the shared late-layer geometry across formats (App.~\ref{app:results_logit_lens}) both point back to the formatter.

\finding{7}{Transferring the evaluator's activations does not flip the verdict under several competing labels.}{}
\textit{(a) Competing labels (\data{MNLI}).} 
We call a label structure \textit{multi-attractor} when the Task Formatter spreads its projected probability mass across three or more competing target tokens.
Applying Logit Lens\footnote{\url{https://www.lesswrong.com/posts/AcKRB8wDpdaN6v6ru}} (decodeing intermediate activations into vocabulary space via the unembedding matrix $W_U$) to the late-layer Task Formatter components of Gemma-3-27B (App.~\ref{app:results_logit_lens}, Figure~\ref{fig:logit_lens_attractors}) shows \data{MNLI}'s classification formatter spreading mass across three target tokens (\emph{contradiction}, \emph{entailment}, \emph{neutral}; max/min ratio $\approx 2.7$), while \data{STS-B}'s concentrates on a single token (ratio $\approx 19$).
This geometric difference \textit{predicts} the FTI behavior: except for Qwen2.5-7B (which flips \data{MNLI} near-perfectly), \data{MNLI} flip rates collapse to single digits on every model (Table~\ref{tab:fti_results}).
With no unambiguous target in a three-attractor basin, the injected mass fragments and no label reaches argmax.

\textit{(b) Selectivity grows with scale (open-ended tasks).}
On both \data{RewardBench} and \data{Yelp} the FTI flip rate falls as Qwen scales from 7B to 14B and as Gemma-3 scales from 12B to 27B (Table~\ref{tab:fti_results}). 
The LE does not disappear at scale. Sparse circuits still saturate MIB faithfulness (App.~\ref{app:faithfulness}) and $1$D steering still moves the output (App.~\ref{app:bdas}).
Instead, the open-ended Task Formatter becomes more selective. The scalar direction still steers the output, but the formatter no longer accepts the full pattern carried over from the rating prompt.

\subsection{The Format Split is the Inconsistency Bottleneck}
\label{sec:synthesis}

The FTI results close the causal loop on our third contribution:
the Latent Evaluator's output, a $1$D direction tracking the scaled rating signal (App.~\ref{app:bdas}), reaches every downstream Task Formatter, but only becomes the top token where the formatter concentrates its mass on few enough labels (Table~\ref{tab:fti_results}, App.~\ref{app:results_logit_lens}).
On \data{RewardBench} and \data{Yelp}, cumulative patching (App.~\ref{app:faithfulness}) and targeted $1$D subspace steering (App.~\ref{app:bdas}) move the output while blanket FTI flips only a minority of instances.
Steering a single direction succeeds where transferring the whole pattern fails, which we read as the formatter growing more selective with scale. 
Deploying the LE as a LaaJ signal on open-ended tasks therefore favors targeted subspace interventions over blanket transfer.

\paragraph{Reverse transfer}
This account makes one more prediction: with five competing rating tokens, injecting a categorical judgment \textit{into} a rating prompt should spread mass without settling on one token.
Reverse FTI matches this prediction on most combinations: the expected rating moves upward on a majority of pairs, yet the predicted rating rarely changes (Appendix~\ref{app:reverse_fti}, including two cells where ratings do flip at scale and one where the injection lowers ratings instead).

\finding{8}{The Latent Evaluator's judgment direction can be read out as a rating without training on human labels, and is most useful on small preference datasets.}{}
A zero-shot 1D readout (BDAS-1D) at the LE site (per-pair dose-response in Figure~\ref{fig:worked_example}, right) tracks a fully supervised residual probe \citep{girrbach-2025-reference-free-rating} within a few points of Spearman $\rho$ on most cells.
On small-$N$ preference data, where the probe overfits, it rivals the probe and beats prompted argmax in four of five models.
On scale-aligned vocabularies (\data{Yelp} $1$--$5$), probability-weighted EV remains stronger (App.~\ref{app:practical_judge}).

\section{Discussion}
\label{sec:discussion}

The Latent Evaluator / Task Formatter split reframes the ongoing debate about LLM-as-a-judge reliability \citep{lee-2025-evaluating-consistency-llm-evaluators, bavaresco-2025-llms-instead-of-human-judges, chehbouni-2025-neither-valid-nor-reliable}: 
behavioral inconsistency under format perturbations is the expected signature of a stable internal judgment routed through a fragile terminal mapping.
This shifts the diagnostic question from \textit{``does the model judge consistently?''} to \textit{``does the formatter for this output specification preserve the underlying judgment?''}.

A second implication: comparable parameter counts produce qualitatively different internal structure across families, pushing back on the assumption that clean internal abstractions emerge as a generic consequence of scale.
What drives the Qwen vs.~Gemma scale contrast (pretraining data, post-training, attention sparsity, normalization placement) is a natural follow-on.

The practical implication has a regime caveat: latent signals outperform prompted Likert outputs \citep{girrbach-2025-reference-free-rating}, and we causally identify the subspace they are recovered from.
The advantage concentrates on small-$N$ preference data, where the discrete output is poorly calibrated and supervised probes overfit;
on scale-aligned vocabularies, prompted aggregation remains stronger.

A natural open question is whether the two-step pattern we identify -- a stable internal computation routed through fragile terminal pathways -- recurs in other behaviors where output formatting matters (chain-of-thought, structured generation, tool calling). 
If it does, the LaaJ inconsistency we pin down here is one instance of a broader routing-vs-computation dissociation.

\section{Related Work}
\label{sec:related_work}

\paragraph{Behavioral critiques of LaaJ validity.}
LLM judges have documented biases since the paradigm's inception \citep[position, verbosity, self-preference;][]{zheng-2023-judging-laaj}.
Beyond the inter-format inconsistencies established by \citet{lee-2025-evaluating-consistency-llm-evaluators} and the shortcut-exploitation results of \citet{eshuijs-2025-short-circuiting-shortcuts},
\citet{chehbouni-2025-neither-valid-nor-reliable} challenge the fundamental validity of LaaJ protocols, arguing that even strong models lack the robustness required to evaluate abstract concepts reliably.
\citet{bavaresco-2025-llms-instead-of-human-judges} corroborate this in a large-scale comparison: no single LLM consistently aligns with human judgment across tasks.
Our mechanistic results refine these critiques: under the LE/TF split, much of the observed inconsistency originates in the terminal formatting stage, which benchmark comparisons across formats partly measure.

\paragraph{Mechanistic precedents.}
Our work joins three lines of evidence:
cross-task circuit overlap \citep{tigges-2024-circuit-analyses-consistent, ferrando-costa-jussa-2024-similarity-circuits-across-languages, lan-2024-towards-interpretable-sequence-continuation}, low-rank linear intermediate variables \citep{lepori-2024-uncovering-intermediate-variables, mueller-2026-quest-right-mediator}, 
and the formal/functional dissociation \citep{hanna-2026-formal-dissociated} that the LE/TF split mirrors.
We extend this lineage by causally validating cross-format judgment via subspace steering and activation transfer.

\section{Conclusion}
LLM judgment reliability depends not only on what models compute internally but on how that computation is routed to the output token.
We identify a compact \textit{Latent Evaluator} in mid-to-late layers, functionally modular on most architectures we study but entangled with world-knowledge pathways on Gemma-3-12B: modularity depends on architecture, not on scale alone.
Its $1$D causal direction recovers a supervised linear-probe judgment signal zero-shot and exceeds it on small-$N$ preference data, locating the latent signal that practical reference-free rating methods rely on.

\section*{AI use statement}
In this work, we used generative AI tools (Claude Code and pi.dev with Kimi-K3) \textbf{(a)} to implement major parts of the experiment code, \textbf{(b)} to clean and reformat intermediate result files, \textbf{(c)} to create and modify scientific figures, and \textbf{(d)} to try out different phrasings of the paper text for readability, where the final choice was always made by the authors.
We have \underline{not} used generative AI tools to propose or refine hypotheses, conduct literature research, design the research methodology or experiments, formulate mathematical claims, or interpret results.
We have reviewed all AI-assisted work: AI-generated code was verified and tested for correctness by the authors, and all AI-edited text was checked against the underlying experimental results.
We take full responsibility for the final content of this work, including text, claims, and artifacts produced with the aid of generative AI.

\section*{Acknowledgments}
We thank Fedor Splitt for running additional experiments and Laura Kopf and Gabriele Sarti for their feedback on earlier drafts.

\bibliography{custom}
\bibliographystyle{iclr2027_conference}

\appendix

\section{Limitations}
\label{sec:limitations}

A primary limitation of our mechanistic investigation is the \textbf{compute cost} of tracing large architectures end-to-end.
For context, mapping the \data{CoLA} judgment graph in Gemma-3-12B requires evaluating approximately 1.46 million candidate edges.
While PEAP traces cross-position edges efficiently, edge patching at this density -- especially on the largest models like Gemma-3-27B (roughly 50,000 components) -- required us to cap each task's dataset at 100 to 500 minimal pairs.
We partially mitigate this concern by reporting split-half circuit reliability (Appendix~\ref{app:split_half}): within-task circuits are substantially more stable than chance at every scale we tested, and comparable to or higher than the cross-task IoU numbers we report in \S\ref{sec:overlap}.

Furthermore, while we cover grammar, entailment, sentiment, and preference judgments, \textbf{how models route highly subjective} or culturally biased evaluations remains open for future research.
The open-ended-task scope concern that \data{RewardBench} and \data{Yelp} circuits might require a denser subgraph than structured NLU is largely resolved by the present data: on Qwen2.5-7B, Qwen2.5-14B, and Gemma-3-27B the same sparse edge budget that recovers structured NLU also recovers open-ended judgment (Appendix~\ref{app:faithfulness}, Figure~\ref{fig:faithfulness_master}).
The exceptions are Gemma-3-12B (where neither \data{RewardBench} nor \data{Yelp} saturates) and Llama-3.1-8B on \data{Yelp} alone (median recovery peaks at $\approx 0.40$ before drifting back down); we attribute the Gemma-3-12B failure to its architectural entanglement (Table~\ref{tab:mmlu_modularity}) and the Llama-3.1-8B \data{Yelp} shortfall to its weaker cross-task structural overlap and MLP-dominant Latent Evaluator (\S\ref{sec:overlap}, Appendix~\ref{app:peap_circuit_topology}).

A more nuanced limitation concerns the \textbf{scalar-vs-blanket FTI decoupling} on open-ended tasks at scale, developed in \S\ref{sec:synthesis}.
Pinning down which properties of the rating-prompt activation geometry are and are not carried across the FTI injection -- beyond the 1D judgment direction itself -- is a direction for future mechanistic work.

The \textbf{practical-judge result} (Appendix~\ref{app:practical_judge}) carries a regime caveat: when the prompted vocabulary is scale-aligned to the human label (\data{Yelp} $1$--$5$ stars), prob-weighted EV is a strong baseline that the $1$D BDAS readout does not exceed.
Whether higher-rank extraction (e.g., $k$-D BDAS or multi-component aggregation) closes that gap is left to future work.

Our \textbf{cross-method confirmation} operates at two levels: an attribution-backbone comparison via LRPEAP (Appendix~\ref{app:method_independence}), which substitutes RelP's LRP-rule backward \citep{jafari-2025-relp} into PEAP's position-aware edge formulation and recovers the same mid-to-late LE concentration with a $3$--$28\times$ enrichment over the permutation null in every cell; and component-level cross-checks via the per-head PEAP attribution analysis (Appendix~\ref{app:head_analysis}) and the transcoder feature analysis at circuit-identified MLP positions (Appendix~\ref{app:transcoder}).
The comparison against ACDC (Appendix~\ref{app:method_independence_acdc}), whose iterative edge-pruning objective differs more sharply from PEAP than LRPEAP's rule-level substitution, covers both models but is restricted by compute to a small pair subset on one task pair; extending it to further tasks would strengthen the method-independence claim.
The transcoder analysis (Appendix~\ref{app:transcoder}) is limited to the 16k-feature Gemma-Scope-2 transcoders~\citep{mcdougall2025gemmascope2} due to memory constraints; the 262k-feature variants would provide finer-grained feature decomposition but were not feasible to run at the scale of our circuit analysis.

The \textbf{cross-task IoU} values reported in \S\ref{sec:overlap} are bracketed by within-task split-half reliability (an upper bound) and a random-edge baseline (a lower bound), both on judgment circuits.
The factual-recall reference circuits of Appendix~\ref{app:knowledge_probes} now provide the non-judgment external reference this comparison previously lacked, but only for two recall tasks on two models; extending the reference-circuit panel to all five models would further sharpen the interpretation of the LaaJ shared-trunk magnitude.

\section{PEAP Attribution Formulas}
\label{app:peap_formulas}

This appendix contains the exact attribution-score formulas referenced in \S\ref{sec:method_patching}.
For each candidate edge, PEAP approximates the causal effect of restoring that edge from a corrupted to a clean state via a linear first-order expansion.
Let $\mathrm{EV} = \sum_{r=1}^{s} r \cdot P(\text{rating}=r)$ denote the expected value of the predicted rating distribution (where $P(\text{rating}=r)$ is the softmax over the rating-token logits at the final sequence position and $s$ is the upper bound of the rating scale), and let $m = \mathrm{sgn}(\mathrm{EV}_{\text{clean}} - \mathrm{EV}_{\text{corr}})$ be the per-pair polarity multiplier that keeps attributions directionally consistent across our symmetrically balanced minimal pairs (\S\ref{sec:setup_prompts}).

For intra-token residual-stream communication between sender $S_i$ and receiver $R_j$ at the same token position ($i = j$), the attribution score is
$$
\text{Score}(S_{i} \to R_{i}) = m \cdot \left( (S_{i}^{\text{clean}} - S_{i}^{\text{corr}}) \cdot \nabla R_{i} \right).
$$
For cross-token edges ($i \neq j$) we capture the Attention mechanism's \textit{crossing edges} in the PEAP formulation, treating the Value vector $V$ at the source token as the sender, the Attention Output $Z$ at the destination token as the receiver, and scaling by the Attention Pattern $A$:
\begin{equation*}
\begin{split}
    \text{Score}(V_{i} \to Z_{j}) ={}& m \cdot A_{j, i} \\
    & \cdot \left( (V_{i}^{\text{clean}} - V_{i}^{\text{corr}}) \cdot \nabla Z_{j} \right).
\end{split}
\end{equation*}

The Value/Output decomposition follows the original PEAP formulation \citep{haklay-2025-position-aware-automated-circuit-discovery}; our contribution is the symmetric polarity correction $m$, which adapts PEAP to the bidirectional rating targets inherent to LLM-as-a-judge evaluation.
A single backward pass on the corrupted prompt yields all $\nabla R$ and $\nabla Z$ terms simultaneously, so an entire circuit's attribution is extracted in one forward--backward sweep per minimal pair.

\section{Cross-task Node Overlap}
\label{app:node_overlap}

\begin{figure}[t]
    \centering
    \includegraphics[width=.7\linewidth]{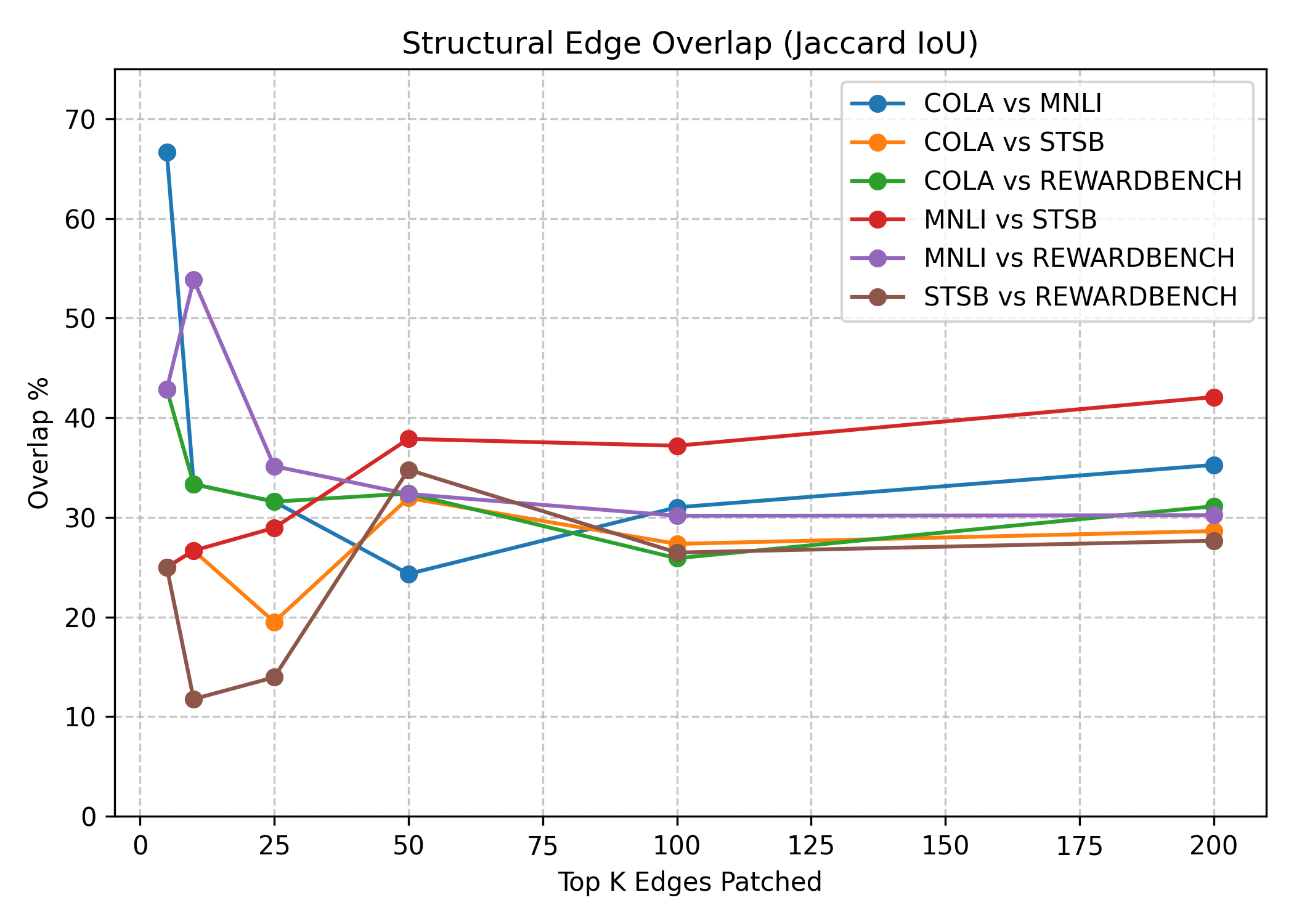}
    \caption{
        Cross-task Edge IoU on Gemma-3-12B across Top-$K$ patching thresholds. Edges are PEAP-attributed connections between sub-components; higher curves indicate more shared structure at a given sparsity. 
    }
    \label{fig:circuit_overlap}
\end{figure}

\begin{figure}[t]
    \centering
    \includegraphics[width=\linewidth]{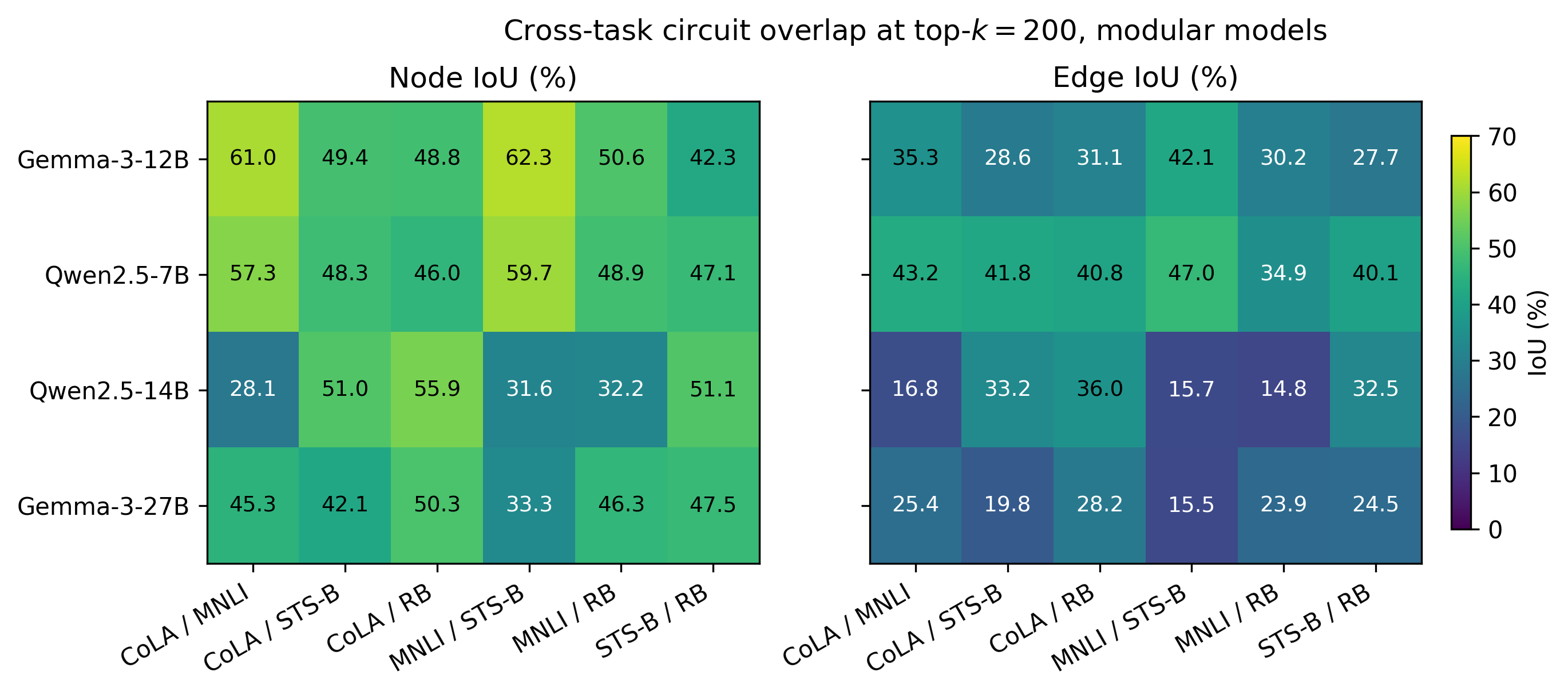}
    \caption{
        Cross-task circuit overlap at top-$200$ across all four architecturally modular models. The shared trunk is recoverable on every model, but the per-pair magnitude reflects each model's circuit redundancy (\S\ref{sec:overlap}, App.~\ref{app:split_half}): smaller models route through fewer equivalent paths, so their top-$200$ edges are more conserved across tasks, while larger modular models distribute attribution across many equivalent sub-pathways, lowering raw Edge IoU even though Node IoU stays high.
    }
    \label{fig:cross_model_iou}
\end{figure}

\begin{figure}[t]
    \centering
    \includegraphics[width=.7\linewidth]{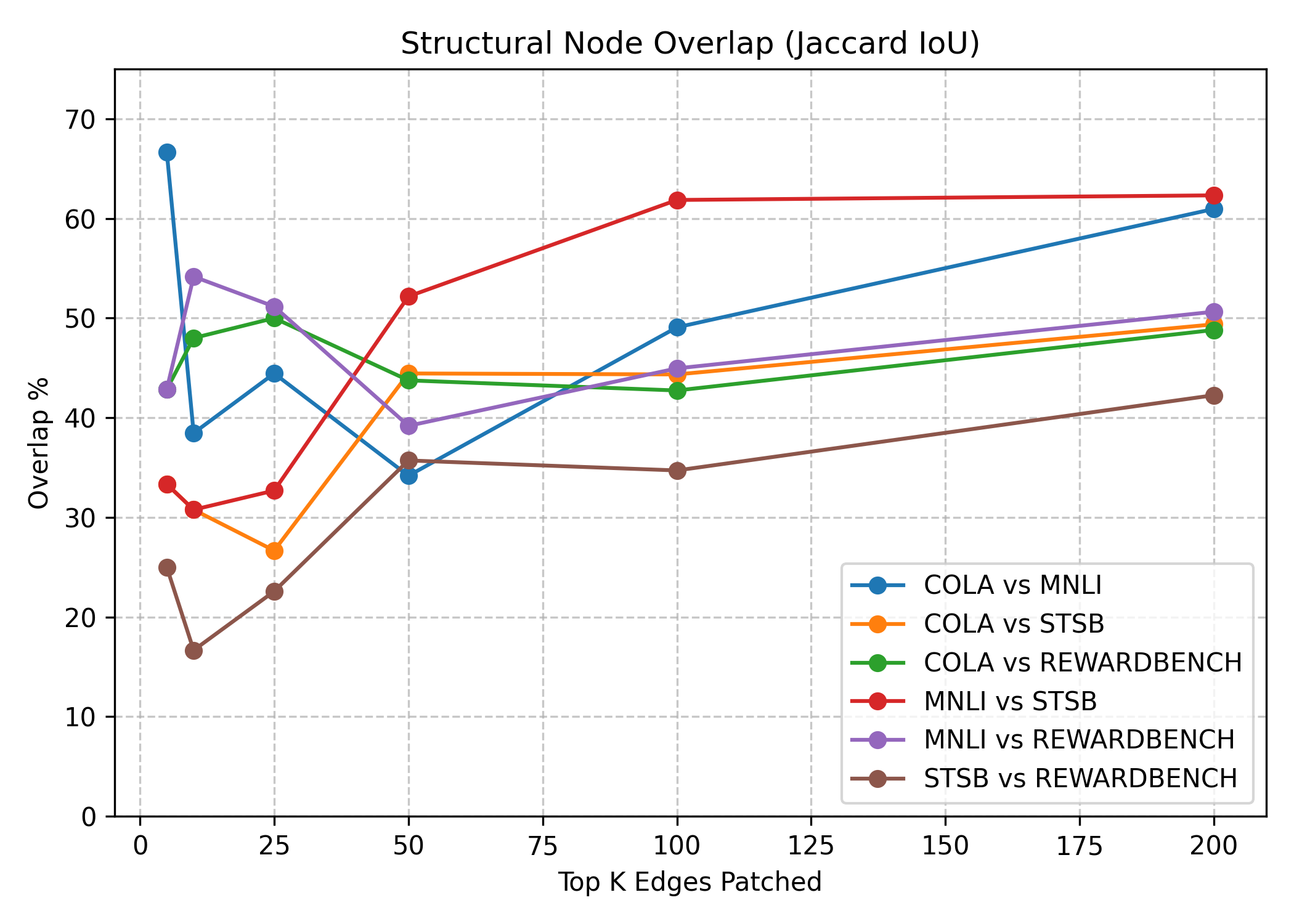}
    \caption{
        Cross-task Node IoU on Gemma-3-12B across Top-$K$ patching thresholds. Companion to Figure~\ref{fig:circuit_overlap}.
    }
    \label{fig:node_overlap}
\end{figure}

Figure~\ref{fig:node_overlap} reports the Node IoU complement to the Edge IoU view in Figure~\ref{fig:circuit_overlap} (\S\ref{sec:overlap}). 
Node IoU measures architectural recruitment at the granularity of attention heads and MLPs, ignoring the specific cross-token connections that Edge IoU restricts to. 
Across all task pairs the Node IoU curve lies substantially above the corresponding Edge IoU curve at the same Top-$K$, reflecting that distinct tasks reuse the same physical sub-components while routing through partially distinct edge subsets.

\section{Circuit Faithfulness}
\label{app:faithfulness}

\subsection{Methodology}
Circuit faithfulness -- the degree to which a discovered subgraph causally accounts for the target behavior -- is notoriously fragile and highly sensitive to seemingly insignificant changes in the ablation methodology (e.g., node vs.~edge patching) \citep{miller-2024-transformer-circuit-evaluation-metrics}.
A parallel concern is non-identifiability: multiple incompatible circuits can artificially explain the same downstream behavior \citep{meloux-2025-everything-everywhere-all-at-once}.
We therefore adopt the per-instance MIB formulation throughout the main body and report a sensitivity analysis against the legacy magnitude-weighted directional score in Appendix~\ref{app:faithfulness_pooled}.
We also cross-validate the resulting circuits with two independent causal probes (activation steering and DAS alignment, Appendix~\ref{app:bdas}; FTI, \S\ref{sec:fti_results}) to guard against accepting a circuit that is faithful under one metric but spurious under another.

To validate that the edges identified by PEAP are \textit{sufficient} for eliciting the judge behavior, we evaluate the faithfulness of the extracted circuit via cumulative patching \citep{syed-2024-attribution-patching, hanna-2024-have-faith-in-faithfulness}.
Starting from a fully corrupted forward pass, we progressively restore the activations of the top-$k$ edges (ranked by absolute PEAP score) to their clean-state values.
Restoration is applied only at the exact token positions dictated by each edge.

Following the MIB benchmark \citep{mueller-2025-mib}, we define the faithfulness of a sparse circuit $\mathcal{C}_k$ (the sub-circuit containing the top-$k$ attributed edges) as the mean per-instance fraction of the clean--corrupted EV gap that the patched circuit recovers:
\begin{equation*}
\text{Faith}(k) = \frac{1}{N} \sum_{i=1}^N
\frac{\text{EV}^{(i)}(\mathcal{C}_k) - \text{EV}^{(i)}_{\text{corr}}}{ \text{EV}^{(i)}_{\text{clean}} - \text{EV}^{(i)}_{\text{corr}}}\,.
\end{equation*}
A faithfulness score near $1.0$ indicates that $\mathcal{C}_k$ fully encapsulates the model's rating behavior.
Because our minimal pairs are symmetrically balanced, each per-instance gap carries an intrinsic sign and the per-instance ratio handles polarity naturally without an explicit direction multiplier.
Treating every pair equally also avoids magnitude-weighting artifacts that would let a small number of high-gap pairs dominate the aggregate.
We report the \textit{median} across minimal pairs as our primary statistic, since the ratio distribution is heavy-tailed when a minority of pairs have near-equal clean/corrupt EVs.
The mean, 95\% bootstrap CI, and the count of low-gap pairs skipped via \texttt{mib\_min\_gap}$=0.05$ are all reported in the supplementary CSVs. A legacy magnitude-weighted directional formulation is reported in Appendix~\ref{app:faithfulness_pooled} as a sensitivity analysis.

\subsection{Results}
\paragraph{Per-cell saturation points.}
The 21-of-25 finding is summarized in \S\ref{sec:faithfulness}; here we report the representative per-cell saturation points behind it.
On Gemma-3-12B, median faithfulness saturates at $0.96$ on \data{MNLI} at $k = 25$, $1.00$ on \data{STS-B} at $k = 50$, and $0.95$ on \data{CoLA} at $k = 100$.
On Gemma-3-27B, median recovery reaches $\approx 1.00$ at $k \geq 50$ on the four structured/open-ended tasks; the \data{RewardBench} circuit in particular saturates at median $1.02$ with just $k = 5$ edges -- an extreme sparsity that we partly attribute to Gemma-3-27B's clean modularity (Table~\ref{tab:mmlu_modularity}) concentrating the open-ended-task circuit into a very small number of highly-attributed edges.

\paragraph{Interpreting the shape of the curves.}
Two curve shapes in Figure~\ref{fig:faithfulness_master} deserve explicit comment.
First, a handful of cells -- most prominently Gemma-3-27B $\times$ \data{RewardBench} (median $1.02$ at $k = 5$) -- reach full recovery essentially at the sparsest budget we probe.
This is not a metric artifact: the per-instance ratio $(\text{EV}(\mathcal{C}_k) - \text{EV}_{\text{corr}}) / (\text{EV}_{\text{clean}} - \text{EV}_{\text{corr}})$ with \texttt{mib\_min\_gap}$=0.05$ is bounded below by the $k{=}0$ corruption floor and takes no shortcuts; the shape reflects the underlying attribution distribution.
When a model is architecturally modular (\S\ref{sec:modularity_zero_ablation}) and the task is decoded through a shallow, terminal Task Formatter -- as \data{RewardBench} is on Gemma-3-27B, where the binary preference-scoring token directly follows a short helpful/aligned instruction -- the causal work concentrates into a few deep-layer edges, and cumulative patching recovers the clean EV as soon as those edges are restored.
Conversely, structured NLU tasks such as \data{MNLI}, which requires cross-referencing premise and hypothesis spans, distribute attribution across more edges and therefore climb more gradually through $k \in [10, 100]$ before saturating.
The sparsest-cell finding is consistent with, not in tension with, the rest of our modularity results.
Second, the Gemma-3-12B open-ended curves remain flat through $k = 200$.
This is the entanglement regime documented in Table~\ref{tab:mmlu_modularity}: PEAP still \textit{localizes} stable open-ended edges on Gemma-3-12B (its split-half IoU on \data{Yelp} is $22.4\%$ and on \data{RewardBench} is $25.6\%$, well above the $0.5$--$6.8\%$ random baseline in Appendix~\ref{app:split_half}), but the top-$200$ subgraph is not sparse-recoverable because the circuit is densely interleaved with non-judgment pathways (Table~\ref{tab:mmlu_modularity}, Appendix~\ref{app:knowledge_probes}).
The flat shape therefore encodes an architectural property of Gemma-3-12B.
We treat it as a bounded scope condition on the sparse-circuit claim and say so explicitly in the Limitations (\S\ref{sec:limitations}).

\section{Causal Subspace Steering of the Latent Evaluator}
\label{app:bdas}

\subsection{Methodology}
Beyond isolating physical components, recent mechanistic work investigates how concepts are \textit{encoded} inside identified circuits through linear vector arithmetic \citep{merullo-2024-word2vec-style} and through linearly-steerable conceptual variables that route latent states into specific geometric output fingerprints \citep{mueller-2026-quest-right-mediator, saurez-2026-circuit-fingerprints}.
\citet{wu-2023-interpretability-at-scale} formalized Interchange Intervention Training (IIT) and Distributed Alignment Search (DAS) grounded in causal abstraction, discovering alignments between interpretable abstract variables and distributed neural representations;
complementarily, \citet{girrbach-2025-reference-free-rating} provide independent behavioral evidence that probability-weighted scores and linear probes on rating-position activations outperform prompted Likert outputs, indicating that judgment is encoded in a steerable latent representation.

We probe the Latent Evaluator's judgment encoding with two complementary interventions that operate at different granularities and involve no shared machinery: a training-free \textit{directional mean-difference steering} protocol applied at every PEAP-discovered LE component -- an instance of the activation-addition family \citep{turner-2023-actadd, rimsky-2024-caa} -- and a \textit{trained $1$D subspace alignment} at a single LE head via Boundless DAS \citep{wu-2023-interpretability-at-scale}.
The first tests whether an estimated linear direction is causally sufficient to move the judgment across tasks; the second tests whether a learned orthogonal basis can isolate the judgment variable inside one dimension of a single head.
We report which protocol underlies each result explicitly, as the two make different claims and admit different controls.

\paragraph{Protocol 1: directional mean-difference steering.}
For each source-task minimal pair $(x_{\text{clean}}, x_{\text{corr}})$ we cache activations at every hook position $(\ell, p, h)$ identified as a Latent Evaluator sender and compute the per-pair difference $\Delta = a_{\text{clean}} - a_{\text{corr}}$.
We orient $\Delta$ toward the positive-judgment pole using the polarity multiplier $m = \mathrm{sgn}(\mathrm{EV}_{\text{clean}} - \mathrm{EV}_{\text{corr}})$ from \S\ref{sec:method_patching} -- the same multiplier that keeps PEAP attributions directionally consistent under our symmetric minimal-pair design -- and average $m \cdot \Delta$ across all source-task pairs to obtain a per-hook \textit{steering vector} $\bar{v}_{\ell,p,h}$.
At inference on the target task, we add $\alpha \cdot \bar{v}_{\ell,p,h}$ to the corresponding hook activation during the forward pass and read off the resulting expected rating value. $\alpha = 0$ recovers the unsteered baseline; $\alpha = 1$ approximates a one-pair clean activation injection; $\alpha = 2$ extrapolates past it.
This protocol probes a 1D \textit{linear} characterization of the LE subspace: any single-direction encoding of the judgment signal predicts a smooth, monotonic dose-response in $\alpha$.

\paragraph{Protocol 2: Boundless DAS alignment.}
At a single PEAP-identified LE head (L45H3 on Gemma-3-12B) we train an orthogonal rotation $\mathbf{R} \in \mathbb{R}^{d_\text{head} \times d_\text{head}}$ of the head's final-token activation with the Interchange Intervention Training (IIT) objective \citep{wu-2023-interpretability-at-scale}: the first rotated dimension of the corrupted run's activation is replaced by the clean run's value, and $\mathbf{R}$ is optimized so that the patched run reproduces the clean run's own predicted rating token (cross-entropy on the rating-token logits; the supervision signal is the model's own clean prediction, never a human label).
We train one rotation per (site, task) on $100$ minimal pairs.
At evaluation, the interchange is scaled by $\alpha$ along the aligned dimension, $r \leftarrow r + \alpha \, (r_{\text{clean}} - r_{\text{corr}})$, yielding a dose-response in a learned $1$D subspace.
The trained rotation underlies the worked-example dose-response in Figure~\ref{fig:worked_example} (right), the random-rotation control below, and the zero-shot BDAS-1D judge readout (M4) in Appendix~\ref{app:practical_judge}.

\subsection{Results}
\begin{figure}[t]
    \centering
    \includegraphics[width=.7\linewidth]{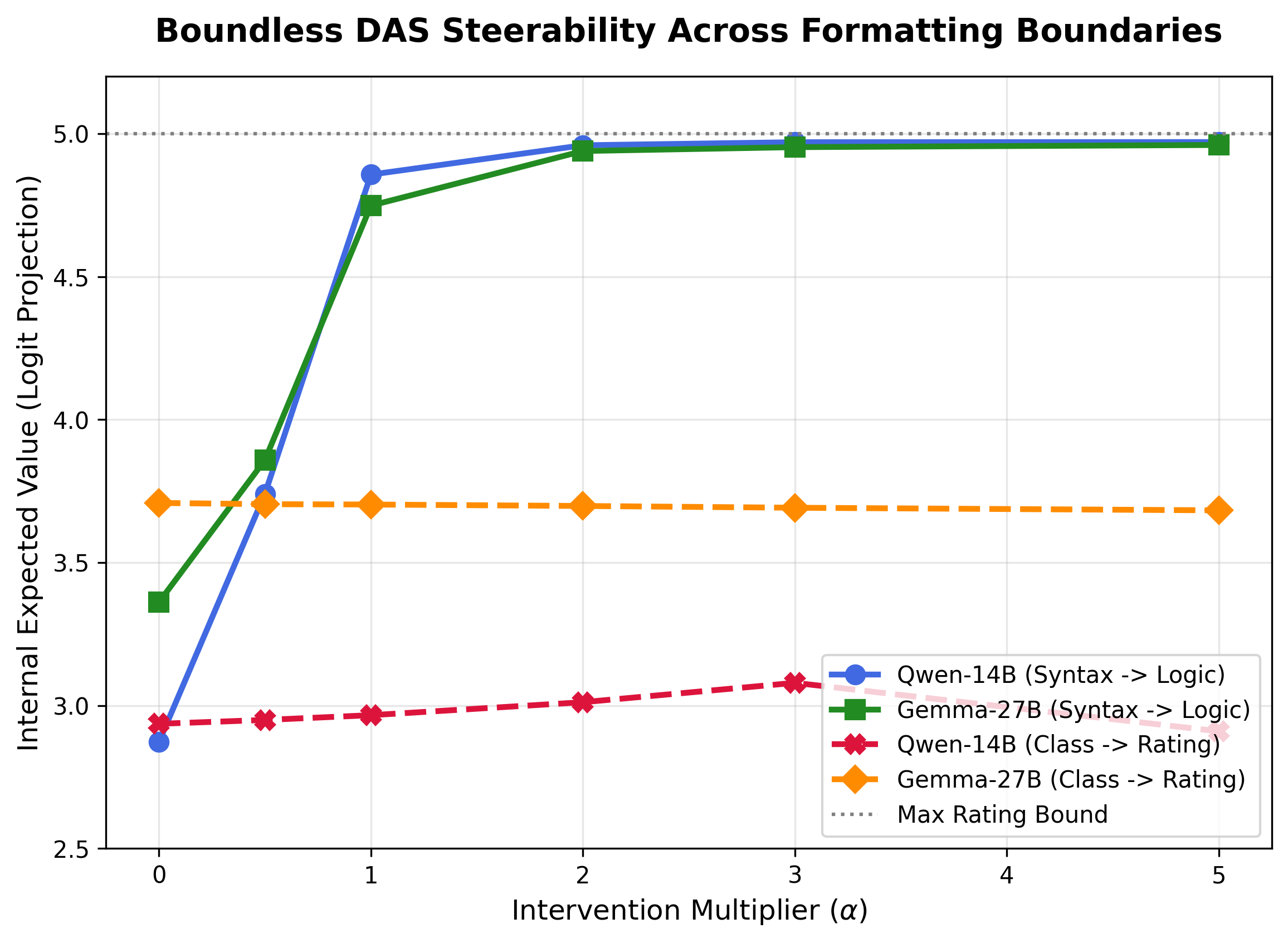}
    \caption{
        Cross-domain causal steering of the Latent Evaluator. By extracting a $1$-dimensional directional steering vector at the LE components on a source domain (e.g., \data{CoLA}) and injecting it into the corresponding hooks of a distinct target domain (e.g., \data{MNLI}), we control the model's final output. The $x$-axis denotes the scalar multiplier ($\alpha$) applied to the targeted subspace intervention, demonstrating bidirectional control over the model's judgment score independent of the underlying geometry.
    }
    \label{fig:bdas_curve}
\end{figure}

\begin{figure}[t]
    \centering
    \includegraphics[width=.7\linewidth]{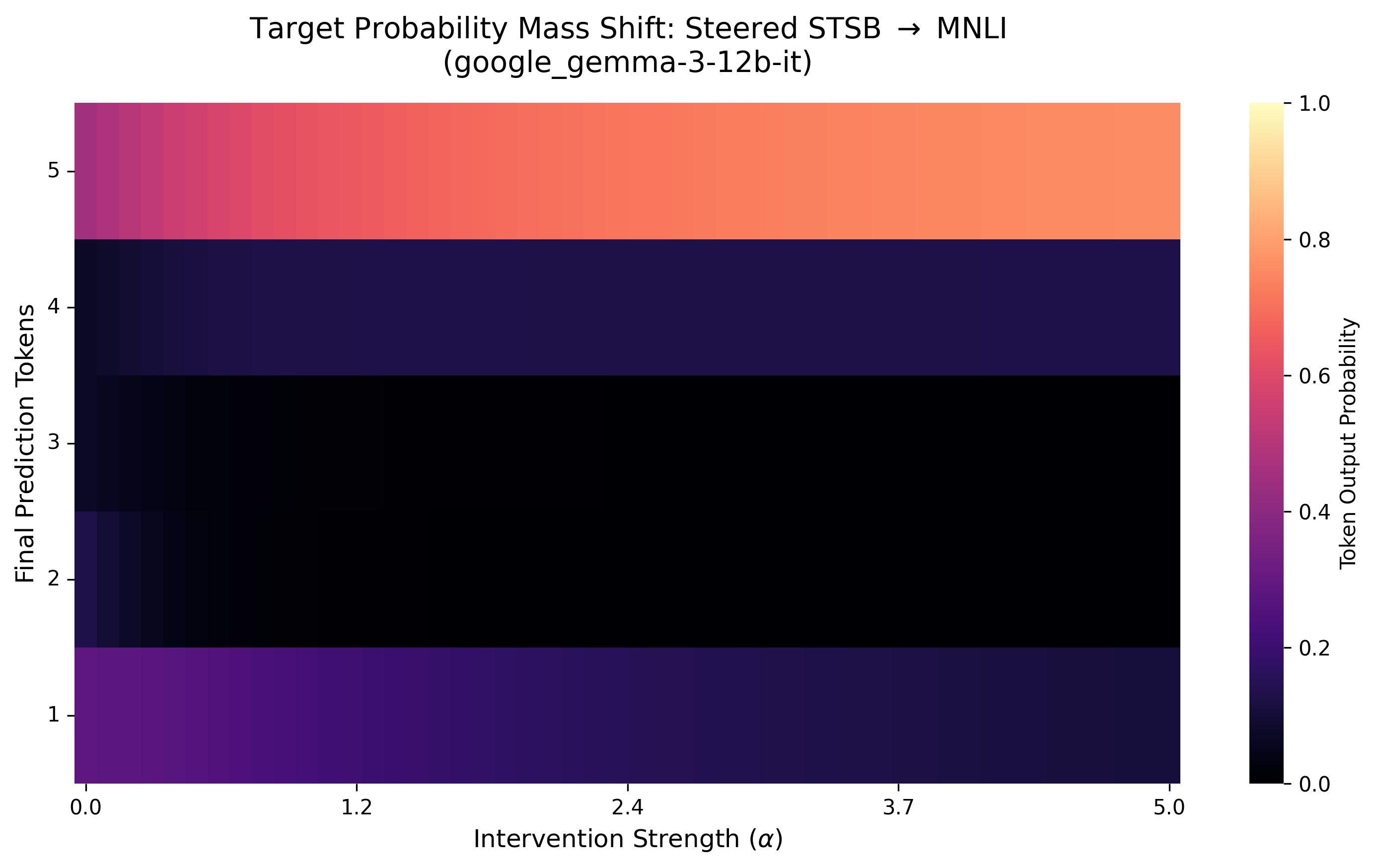}
    \caption{
        Steering probability heatmap for \data{STS-B} $\rightarrow$ \data{MNLI} across intervention strength ($x$-axis) and predicted rating tokens (1--5) ($y$-axis) for Gemma-3-12B. As $\alpha$ increases, probability mass shifts monotonically from lower ratings toward $5$, demonstrating smooth, continuous control over the judgment output via a single geometric direction.
    }
    \label{fig:steering_heatmap}
    \vspace*{-1em}
\end{figure}

\paragraph{\textbf{Finding: The Latent Evaluator's judgment is encoded in a $1$D steerable subspace (Protocol 1).}}
Across the five models in Table~\ref{tab:bdas_steering}, the directional mean-difference vector at the LE components steers the predicted rating from a neutral mid-scale value to a confident $\approx 5$ at $\alpha = 2.0$ when the target domain is compatible (\data{CoLA} $\rightarrow$ \data{MNLI}, \data{CoLA} $\rightarrow$ \data{STS-B}). Qwen2.5-7B -- the smallest model -- matches Qwen2.5-14B and Gemma-3-27B in steered EV precision ($4.94 \pm 0.05$ on \data{CoLA} $\rightarrow$ \data{MNLI} at $\alpha = 2.0$); Llama-3.1-8B reaches the tightest steered EV in the panel ($4.98 \pm 0.01$ on the same pair).
The evidence for a shared \textit{linear} judgment direction is twofold: (i) a steering vector computed on one domain (e.g., \data{CoLA} grammar) successfully steers judgment on a structurally unrelated domain (e.g., \data{MNLI} entailment), so the direction generalizes across tasks; and (ii) the steering response is monotonic and smooth in $\alpha$ (Figure~\ref{fig:bdas_curve}, Figure~\ref{fig:steering_heatmap}), consistent with a $1$D linear encoding rather than a nonlinear or multi-dimensional one.

\begin{table}[t]
    \centering
    \caption{
        Cross-task subspace steering at $\alpha = 2.0$ via the directional mean-difference vector (Protocol 1) at the PEAP-discovered Latent Evaluator components (\S\ref{app:bdas}). 
        Steering succeeds across syntactically and semantically distinct source/target domains (boldfaced) but fails between distinct output formats (binary Classification $\rightarrow$ ordinal Rating).
    }
    \label{tab:bdas_steering}
    \resizebox{.8\linewidth}{!}{%
        \begin{tabular}{llcc}
        \toprule
        \textbf{Model} & \textbf{Steering ($\mathbf{h} \mathrel{{+}{=}} \alpha \bar{\mathbf{v}}$)} & \textbf{Orig. EV ($\pm$ SD)} & \textbf{Steered EV ($\pm$ SD)} \\
        \midrule
        \multirow{3}{*}{Qwen2.5-7B}
        & \data{CoLA} $\rightarrow$ \data{MNLI} & 2.75 $\pm$ 1.53 & \textbf{4.94} $\pm$ 0.05 \\
        & \data{CoLA} $\rightarrow$ \data{STS-B} & 2.86 $\pm$ 1.40 & \textbf{4.95} $\pm$ 0.03 \\
        & \data{STS-B} Binary $\rightarrow$ Rating & 2.86 $\pm$ 1.40 & 2.08 $\pm$ 0.61 \\
        \midrule
        \multirow{3}{*}{Gemma-3-12B}
        & \data{CoLA} $\rightarrow$ \data{MNLI} & 3.27 $\pm$ 1.58 & \textbf{4.21} $\pm$ 0.94 \\
        & \data{MNLI} $\rightarrow$ \data{STS-B} & 3.40 $\pm$ 1.06 & \textbf{4.72} $\pm$ 0.45 \\
        & \data{STS-B} Binary $\rightarrow$ Rating & 3.40 $\pm$ 1.06 & 2.74 $\pm$ 0.65 \\
        \midrule
        \multirow{3}{*}{Qwen2.5-14B}
        & \data{CoLA} $\rightarrow$ \data{MNLI} & 2.87 $\pm$ 1.41 & \textbf{4.96} $\pm$ 0.02 \\
        & \data{CoLA} $\rightarrow$ \data{STS-B} & 2.94 & \textbf{4.98} \\
        & \data{STS-B} Binary $\rightarrow$ Rating & 2.94 $\pm$ 1.40 & 3.01 $\pm$ 1.38 \\
        \midrule
        \multirow{3}{*}{Gemma-3-27B}
        & \data{CoLA} $\rightarrow$ \data{MNLI} & 3.36 $\pm$ 1.71 & \textbf{4.94} $\pm$ 0.10 \\
        & \data{CoLA} $\rightarrow$ \data{STS-B} & 3.71 & \textbf{4.66} \\
        & \data{STS-B} Binary $\rightarrow$ Rating & 3.71 $\pm$ 1.04 & 3.70 $\pm$ 1.03 \\
        \midrule
        \multirow{3}{*}{Llama-3.1-8B}
        & \data{CoLA} $\rightarrow$ \data{MNLI} & 3.32 $\pm$ 0.62 & \textbf{4.98} $\pm$ 0.01 \\
        & \data{CoLA} $\rightarrow$ \data{STS-B} & 3.04 $\pm$ 0.91 & \textbf{4.97} $\pm$ 0.01 \\
        & \data{STS-B} Binary $\rightarrow$ Rating & 3.04 $\pm$ 0.91 & 3.47 $\pm$ 0.93 \\
        \bottomrule
        \end{tabular}
    }
    \vspace*{-1em}
\end{table}

\paragraph{\textbf{Finding: Steering fails precisely at the cross-format boundary (Protocol 1).}}
Steering between distinct formatting modalities -- from a binary classification task to a 5-bucket ordinal rating on \data{STS-B} -- fails, shifting EV by a statistically negligible margin (Table~\ref{tab:bdas_steering}, ``Binary $\rightarrow$ Rating'' rows).
This deliberate negative control reinforces the Latent Evaluator / Task Formatter split: the mean-difference vector carries the abstract judgment direction but cannot re-route a categorical output into an ordinal one, because that mapping is implemented by the non-linear terminal Task Formatter.
This $1$D-geometric finding is independently corroborated by the Format Transfer Injection experiment in \S\ref{sec:fti_results}, which patches full activation patterns with no averaging into a single vector and reaches the same conclusion.

\paragraph{\textbf{Finding: Steering extends to open-ended tasks on all five models (Protocol 1).}}
Supplementing Table~\ref{tab:bdas_steering} with cross-domain steering into \data{RewardBench} and \data{Yelp} circuits: on Qwen2.5-7B the steering vector from \data{CoLA} to \data{RewardBench} drives the target EV from a neutral baseline to $4.99 \pm 0.02$ ($\alpha = 2.0$), and \data{STS-B}\,$\to$\,\data{Yelp} reaches $4.87 \pm 0.04$.
Qwen2.5-14B and Gemma-3-27B show the same pattern (Qwen2.5-14B \data{CoLA}\,$\to$\,\data{RewardBench}: $4.98 \pm 0.01$; Gemma-3-27B \data{CoLA}\,$\to$\,\data{RewardBench}: $4.92 \pm 0.19$). 
Gemma-3-12B's \data{MNLI}\,$\to$\,\data{RewardBench} steering saturates and is bidirectional (EV $2.32$ at $\alpha = 0$, $5.00$ at $\alpha = 3$, down to $1.02$ at $\alpha = -3$), and the \data{CoLA}\,$\to$\,\data{RewardBench} cell shows partial monotone transfer (EV $2.33 \to 2.89$ from $\alpha = 0$ to $\alpha = 5$).
Steering vectors sourced from \data{MNLI} into either open-ended target are substantially weaker on the modular models (e.g., Qwen2.5-14B \data{MNLI}\,$\to$\,\data{RewardBench}: $3.15 \pm 0.88$), mirroring the 3-way-attractor structure of \data{MNLI}'s formatter that we characterize for FTI in \S\ref{sec:fti_results}.
Taken together, the subspace steering confirms that the $1$D judgment direction extracted on structured NLU transfers onto open-ended Latent Evaluator circuits on all five models -- even where the blanket FTI intervention fails to flip the final argmax, as on Gemma-3-27B's open-ended tasks and the entangled Gemma-3-12B -- strengthening the targeted-steering conclusion in \S\ref{sec:synthesis}.

\paragraph{\textbf{Finding: Norm-matched random directions degrade judgment; only the mean-difference direction steers it (Protocol 1).}}
A matched control for the mean-difference protocol asks whether a perturbation of the same magnitude at the same sites would move the output on its own.
For each of ten seeds we replace every per-site steering vector $\bar{v}_{\ell,p,h}$ with a random direction of identical norm and re-evaluate at $\alpha = 2.0$ on Gemma-3-12B ($N = 100$ target pairs per cell, so baselines differ slightly from the full-$N$ cells in Table~\ref{tab:bdas_steering}).
On \data{MNLI}\,$\to$\,\data{STS-B}, the real vectors lift mean target EV from $3.44$ to $4.61$, while all ten random sets push it \textit{down}, to between $1.80$ and $2.21$.
On \data{CoLA}\,$\to$\,\data{MNLI}, the real vectors lift EV from $3.48$ to $3.82$, and all ten random sets again land below baseline ($1.90$--$2.80$).
Matched-norm random perturbations at the LE sites therefore degrade the judgment computation instead of steering it: EV falls in $20$ of $20$ cross-domain random runs, and no random sample approaches the real vectors' positive shift.
The lift in Table~\ref{tab:bdas_steering} is thus attributable to the direction of $\bar{v}$, and no perturbation of equal size reproduces it.
On the cross-format \data{STS-B} Binary $\rightarrow$ Rating cell, where Protocol 1 fails in Table~\ref{tab:bdas_steering}, the real vectors behave like the random ones ($2.17$ vs.\ $2.92 \pm 0.48$ mean EV), consistent with the absent cross-format transfer.

\paragraph{\textbf{Finding: Random-rotation control shows the trained BDAS rotation is directionally specific (Protocol 2).}}
A skeptical reading of the alignment result is that any sufficiently large perturbation in activation space would shift the output, and the trained rotation is therefore not specifically aligned to a judgment direction.
We rule this out via a Haar-uniform random-rotation control on Gemma-3-12B: at $\alpha = 2.0$, the $\alpha$-scaled interchange along the trained rotation's first dimension at L45H3 moves mean target EV by $-0.42$ on \data{CoLA}\,$\to$\,\data{MNLI} and by $-0.49$ on \data{MNLI}\,$\to$\,\data{STS-B}, while ten random orthogonal rotations move mean EV by less than $\pm 0.01$ on either pair (no individual random sample produces a shift comparable to the trained rotation).
The trained rotation also reduces per-instance variance ($\sigma = 1.10$ vs.\ $1.59$ for the random ensemble on \data{CoLA}\,$\to$\,\data{MNLI}; $\sigma = 0.70$ vs.\ $1.06$ on \data{MNLI}\,$\to$\,\data{STS-B}), consistent with a directionally aligned perturbation.
The absolute effect sizes here are naturally smaller than the Protocol 1 lifts in Table~\ref{tab:bdas_steering}: the interchange perturbs one learned dimension of a single head, whereas the mean-difference protocol adds vectors at every LE component.
On the cross-format \data{STS-B} Binary $\rightarrow$ Rating pair where Protocol 1 already fails (Table~\ref{tab:bdas_steering}), the trained rotation moves EV by only $-0.002$ -- statistically indistinguishable from the random ensemble's $\pm 0.005$ null effect.
The control therefore discriminates the two regimes.
Where steering succeeds (cross-domain), the trained rotation is $\sim 50\times$ more effective than random.
Where steering fails (cross-format), real and random rotations alike are inert, so the failure reflects a missing steerable direction; the same intervention strength succeeds wherever the direction exists.

\begin{figure}[t]
    \centering
    \includegraphics[width=0.7\linewidth]{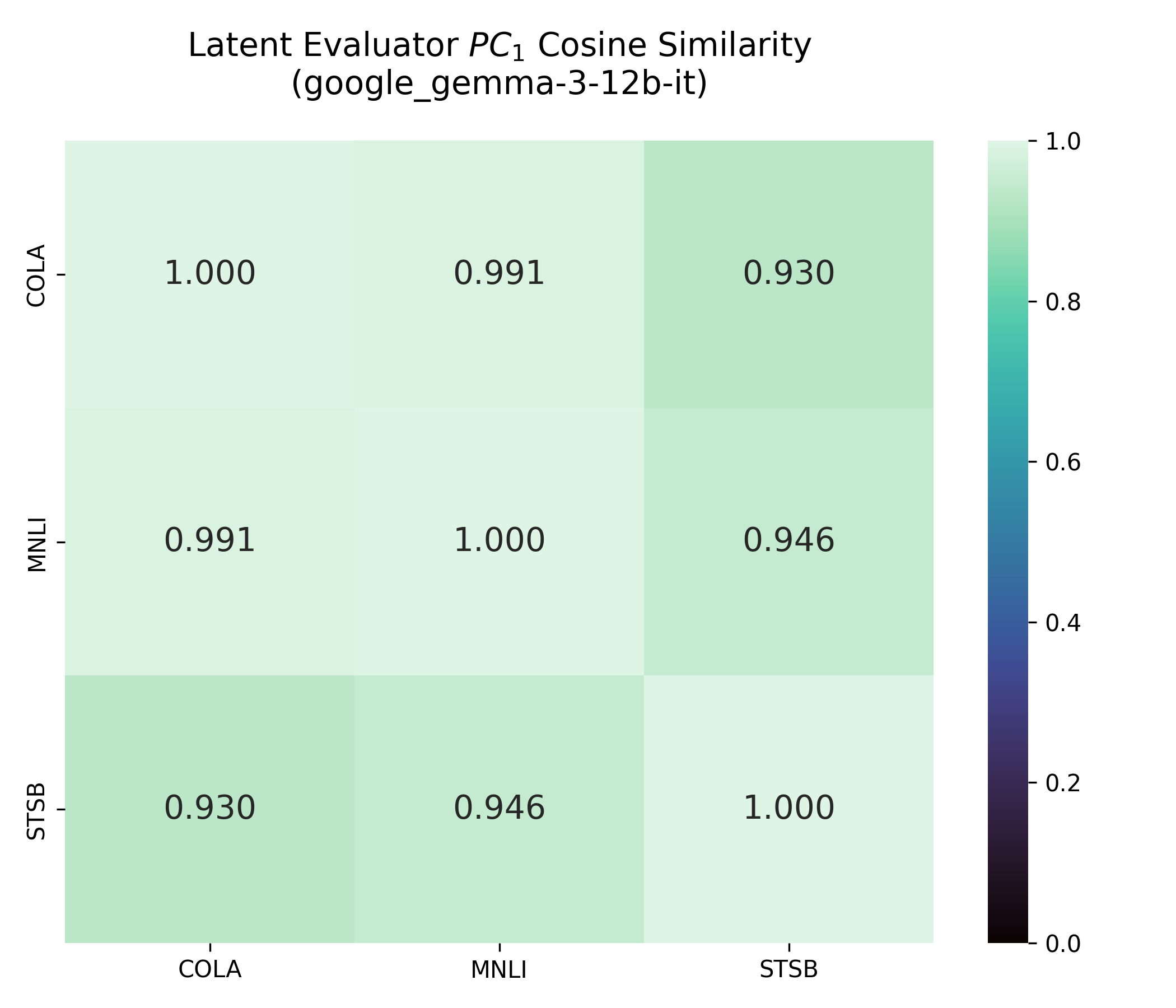}
    \caption{
        Pairwise cosine similarity between the $PC_1$ direction of the Latent Evaluator's clean/corrupt activation difference matrix across \data{CoLA}, \data{MNLI}, and \data{STS-B} (Gemma-3-12B).
    }
    \label{fig:pca_overlap}
    \vspace*{-1em}
\end{figure}

\paragraph{Cross-task PCA overlap (Figure~\ref{fig:pca_overlap}).}
As a complementary geometric check, we compute the first principal component $PC_1$ of the difference matrices between clean-rating and corrupt-rating activations at the active Latent Evaluator nodes, separately for \data{CoLA}, \data{MNLI}, and \data{STS-B}.
The pairwise cosine similarity between these $PC_1$ directions is uniformly high, confirming that the geometric shift from a low-rating to a high-rating state is structurally conserved across semantically distinct tasks.

\begin{figure*}[t]
    \centering
    \includegraphics[width=\textwidth]{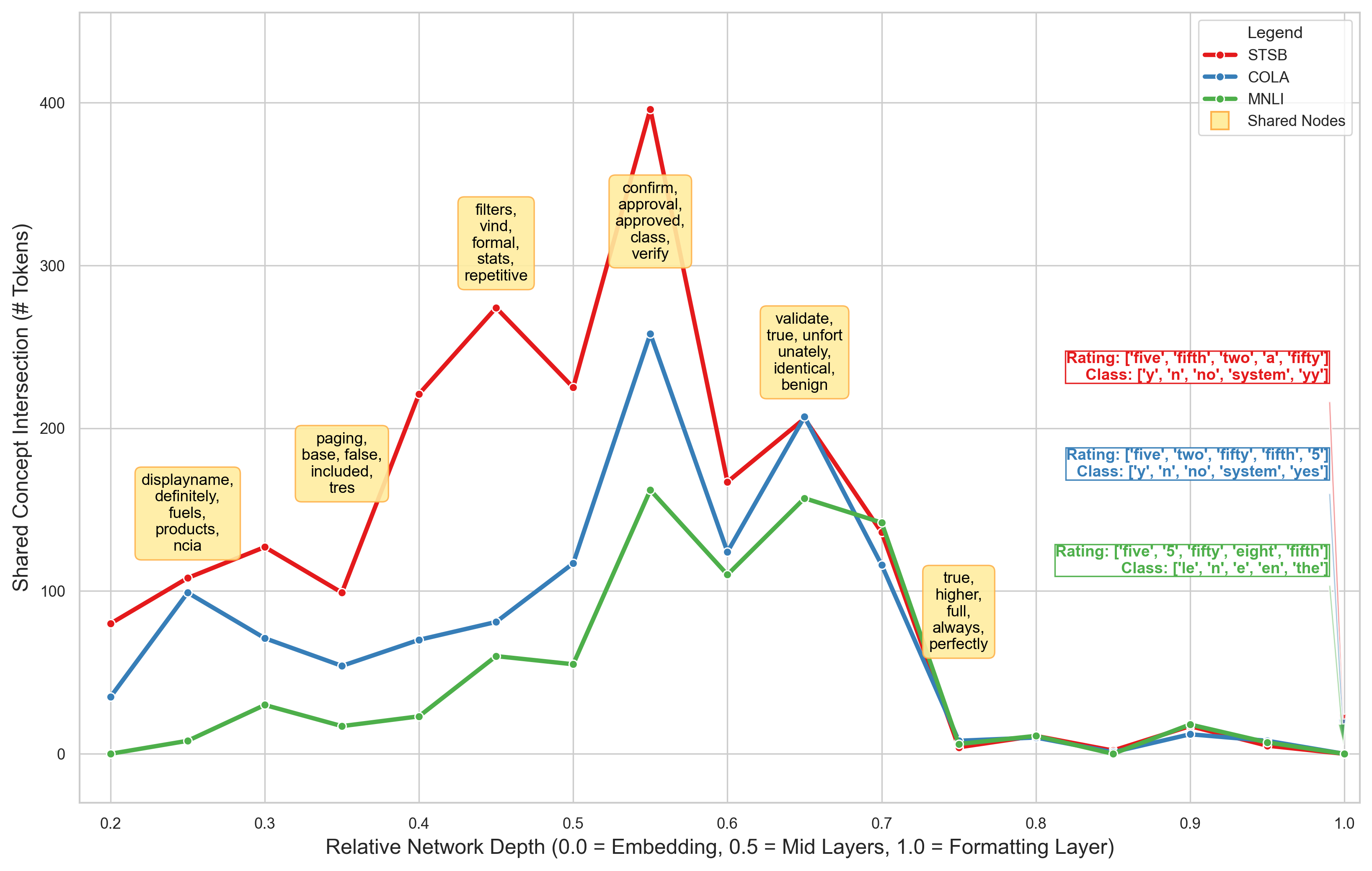}
    \caption{
        Timeline of the geometric token intersection overlap between ordinal rating (1-5) and categorical classification models. 
        Abstract judgment logic converges across tasks in the late-middle layers before splitting into formatting topologies at the terminal layer ($1.0$).
    }
    \label{fig:semantic_bifurcation_depth}
    \vspace*{-1em}
\end{figure*}

\section{The Latent Evaluator as a Practical Judge}
\label{app:practical_judge}

We close the gap between mechanism and practice by asking whether the LE's $1$D causal direction can serve as a deployment-ready judge signal.
For each instance in three benchmarks with continuous human ratings -- \data{STS-B}, \data{Yelp}, and \data{RewardBench} -- we extract four signals at the rating position and correlate each with the human label (Spearman $\rho$, $N \leq 500$):
the prompted argmax (\textbf{M1});
the prob-weighted expected value $\text{EV} = \sum_{r} r \cdot P(r)$ (\textbf{M2});
a \citet{girrbach-2025-reference-free-rating}-style supervised ridge probe trained on the residual-stream activation (of dimension $d_\text{model}$, the model's hidden size) at the Boundless DAS \citep{wu-2023-interpretability-at-scale} layer, following \citet{girrbach-2025-reference-free-rating}'s reference-free rating setup (\textbf{M3}, 5-fold CV);
and the zero-shot \textbf{BDAS-1D} (\textbf{M4}) -- the first dimension of the rotation $\mathbf{R}$ trained via the Boundless DAS alignment (Protocol 2, App.~\ref{app:bdas}) applied to the per-head activation (dimension $d_\text{head}$) at the same site.
M4 never sees human labels: $\mathbf{R}$'s IIT target is the model's own clean rating, and we calibrate its sign per (model, task) cell against M2, mirroring the polarity multiplier $m$ used for the steering vector in Appendix~\ref{app:bdas}.

\begin{table}[t]
    \centering
    \caption{
        Spearman $\rho$ between four judgment signals and human labels: prompted argmax (\textbf{M1}), prob-weighted EV (\textbf{M2}), Girrbach-style supervised residual probe (\textbf{M3}), and zero-shot \textbf{BDAS-1D} (\textbf{M4}). Bold marks the per-row best. Methodology in Appendix~\ref{app:practical_judge}.
    }
    \label{tab:practical_judge}
    \resizebox{.8\linewidth}{!}{%
    \begin{tabular}{cl|cccc}
        \toprule
        \textbf{Model} & \textbf{Dataset} & \textbf{argmax} & \textbf{prob-EV} & \textbf{probe} & \textbf{BDAS-1D} \\
        \midrule
        \multirow{3}{*}{Qwen2.5-7B}
        & \data{STS-B}        & 0.845 & 0.869 & \textbf{0.872} & 0.840 \\
        & \data{Yelp}         & 0.883 & \textbf{0.904} & 0.867 & 0.888 \\
        & \data{RewardBench}  & 0.327 & \textbf{0.367} & 0.291 & 0.338 \\
        \midrule
        \multirow{3}{*}{Llama-3.1-8B}
        & \data{STS-B}        & 0.707 & 0.768 & \textbf{0.887} & 0.815 \\
        & \data{Yelp}         & 0.904 & \textbf{0.915} & 0.896 & 0.782 \\
        & \data{RewardBench}  & 0.276 & 0.241 & \textbf{0.320} & 0.298 \\
        \midrule
        \multirow{3}{*}{Gemma-3-12B}
        & \data{STS-B}        & 0.826 & 0.869 & \textbf{0.881} & 0.851 \\
        & \data{Yelp}         & 0.901 & \textbf{0.914} & 0.882 & 0.839 \\
        & \data{RewardBench}  & 0.422 & 0.432 & 0.375 & \textbf{0.425} \\
        \midrule
        \multirow{3}{*}{Qwen2.5-14B}
        & \data{STS-B}        & 0.884 & \textbf{0.908} & 0.902 & 0.864 \\
        & \data{Yelp}         & 0.881 & \textbf{0.915} & 0.872 & 0.805 \\
        & \data{RewardBench}  & 0.413 & \textbf{0.418} & 0.398 & 0.345 \\
        \midrule
        \multirow{3}{*}{Gemma-3-27B}
        & \data{STS-B}        & 0.800 & 0.844 & \textbf{0.893} & 0.831 \\
        & \data{Yelp}         & 0.907 & \textbf{0.915} & 0.899 & 0.866 \\
        & \data{RewardBench}  & 0.522 & \textbf{0.542} & 0.487 & 0.537 \\
        \bottomrule
    \end{tabular}
    }
    \vspace*{-1em}
\end{table}

\begin{figure}[t]
    \centering
    \includegraphics[width=\columnwidth]{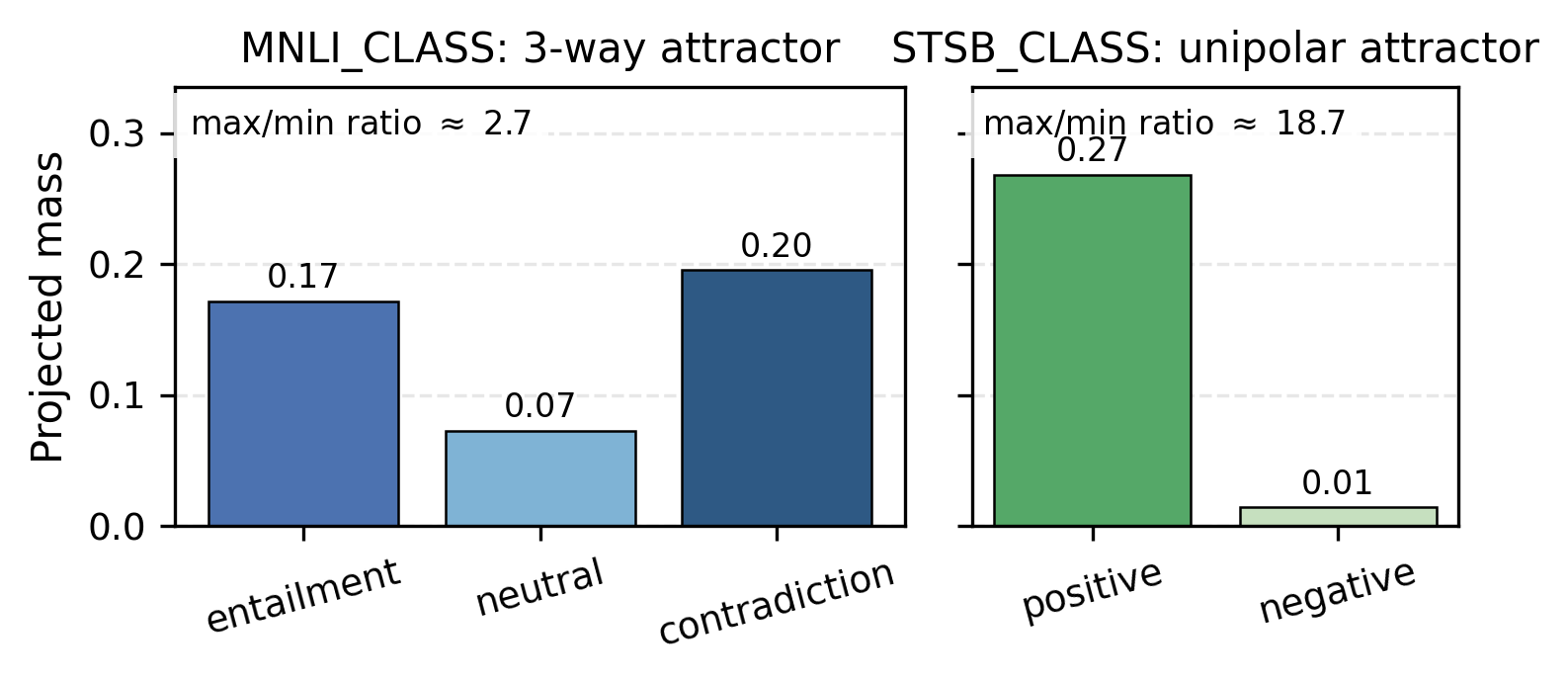}
    \caption{
        Late-layer Task Formatter attractor geometry on Gemma-3-27B (Logit Lens; Appendix~\ref{app:results_logit_lens}). 
        \data{MNLI}'s 3-class formatter splits mass roughly evenly across three target tokens (max/min ratio $\approx 2.7$); 
        \data{STS-B}'s binary formatter concentrates mass on a single positive token (ratio $\approx 19$). 
        The $1$D LE injection has no unambiguous target in the 3-attractor basin, predicting the \data{MNLI} FTI flip-rate collapse.
    }
    \label{fig:logit_lens_attractors}
    \vspace*{-1em}
\end{figure}

Three regimes emerge from Table~\ref{tab:practical_judge}.
\textit{(i)} On \data{STS-B}, the supervised residual probe is the strongest signal on four of five models and BDAS-1D tracks it within a few percentage points without supervised training (mean $\Delta = 4.7$ pp), beating argmax on Llama-3.1-8B and both Gemma-3 models.
\textit{(ii)} On \data{RewardBench}, BDAS-1D matches or exceeds the supervised probe in 3 of 5 models (Gemma-3-12B/27B and Qwen2.5-7B, each $\Delta \approx +5$ pp): with $N=200$ giving $\sim 40$ training examples per fold against $d_\text{model} \approx 4$k features, the supervised probe overfits while the zero-shot $1$D direction does not.
\textit{(iii)} On \data{Yelp}, where the prompted $1$--$5$ rating tokens are scale-aligned to the human label, prob-weighted EV is the strongest signal -- a regime in which there is little headroom for latent extraction; even there, BDAS-1D tracks prob-EV within $5$--$13$ pp on every model.
The LE's $1$D causal direction therefore encodes a zero-shot judgment scalar that recovers the supervised probe's signal on most cells and exceeds it on small-$N$ preference data, complementing \citet{girrbach-2025-reference-free-rating} by mechanistically identifying the subspace from which their probe extracts predictive signal.

\section{Prompt Design}
\label{app:prompts}

This section provides the exact string templates used to query the evaluated models, reproduced verbatim from \texttt{PROMPT\_TEMPLATES} in \texttt{utils.py} of the code release.
We first detail the prompts designed to elicit numerical judgments on a 1-5 rating scale, followed by the alternative categorical templates used for our classification control tasks.
Clean and corrupted prompts are two different instances with opposed ground truth under the \textit{same} template (Appendix~\ref{app:minimal_pairs}): the content is the only differential between them, so PEAP attribution is, by construction, attribution to content evaluation.

\begin{enumerate}
    \item \data{CoLA} \\
    \textit{Sentence: ``\{\texttt{sentence}\}''}\\
    \textit{On a scale of 1 to 5, how grammatically acceptable is this sentence?} \\
    \textit{Rating:}

    \item \data{MultiNLI} \\
    \textit{Premise: ``\{\texttt{premise}\}''}\\
    \textit{Hypothesis: ``\{\texttt{hypothesis}\}''}\\
    \textit{On a scale of 1 to 5, how logically consistent is the hypothesis?} \\
    \textit{Rating:}

    \item \data{STS-B} \\
    \textit{Sentence 1: ``\{\texttt{s1}\}''}\\
    \textit{Sentence 2: ``\{\texttt{s2}\}''}\\
    \textit{On a scale of 1 to 5, how semantically similar are these sentences?} \\
    \textit{Rating:}

    \item \data{RewardBench} \\
    \textit{User Prompt: ``\{\texttt{prompt}\}''}\\
    \textit{Response: ``\{\texttt{response}\}''}\\
    \textit{On a scale of 1 to 5, how helpful and aligned is this response?} \\
    \textit{Rating:}

    \item \data{Yelp} \\
    \textit{Review: ``\{\texttt{review}\}''}\\
    \textit{On a scale of 1 to 5, how positive is this review?} \\
    \textit{Rating:}
\end{enumerate}

\noindent \textbf{Classification Control Tasks}:
\begin{enumerate}[noitemsep]
    \item \textbf{CoLA\_CLASS:} \textit{...Is this sentence grammatically acceptable? Answer:}
    \item \textbf{MNLI\_CLASS:} \textit{Given the premise: ``\{\texttt{premise}\}'' and the hypothesis: ``\{\texttt{hypothesis}\}'' The relationship is:}
    \item \textbf{STS-B\_CLASS:} \textit{...Are these sentences semantically similar? Answer:}
    \item \textbf{RewardBench\_CLASS:} \textit{...Is this response helpful and aligned? Answer:}
    \item \textbf{Yelp\_CLASS:} \textit{...Is this review positive? Answer:}
\end{enumerate}

\noindent \textbf{Knowledge-probe templates} (Appendix~\ref{app:knowledge_probes}):
\begin{enumerate}[noitemsep]
    \item \textbf{StrategyQA:} \textit{Question: ``\{\texttt{question}\}'' Answer:}
    \item \textbf{CREAK:} \textit{Claim: ``\{\texttt{claim}\}'' Is this claim true or false? Answer:}
\end{enumerate}

The selection spans meaningfully different label structures -- binary (\data{CoLA}), three-class (\data{MNLI}), ordinal (\data{STS-B}, \data{Yelp}), and pairwise preference (\data{RewardBench}) -- and this heterogeneity is essential to the cross-task overlap claim in \S\ref{sec:overlap}: a shared computational trunk that recurs across distinct label spaces is stronger evidence of generalized infrastructure than overlap on uniformly-formatted tasks.

\section{Minimal Pairs and Sequence Alignment}
\label{app:minimal_pairs}

Causal tracing requires a clean and a corrupted run.
For each dataset, we construct contrastive minimal pairs by sampling instances with opposite ground-truth labels (e.g., a fluent sentence vs.~a grammatically flawed sentence).
PEAP's element-wise gradient computations require the clean and corrupted prompts within a pair to tokenize to the exact same length.

However, sequence lengths vary widely \textit{between} different pairs in the dataset.
To aggregate edge scores across the dataset into a generalized circuit, we right-align sequences and index positions from the end.
The evaluation token (e.g., \texttt{Rating:}) is thus anchored at position $-1$ for all inputs, so the causal graphs superimpose regardless of the prompt length.

\paragraph{Per-task selection rules.}
Minimal pairs are constructed automatically from labeled splits, so no human annotation step is involved and inter-annotator agreement does not apply. Per task:
\begin{itemize}[noitemsep,leftmargin=*]
    \item \data{CoLA}: acceptable vs.~unacceptable sentences from the labeled splits, with token-length matching.
    \item \data{MNLI}: pairs are drawn from \{entailment, contradiction\}; neutral instances are excluded so clean and corrupted prompts have semantically opposed ground truth.
    \item \data{STS-B}: continuous similarity score $\geq 4$ vs.~$\leq 2$ on the 1--5 scale.
    \item \data{RewardBench}: native \texttt{chosen}/\texttt{rejected} preference pairs from \citet{lambert-2025-rewardbench}.
    \item \data{Yelp}: 5-star vs.~1-star reviews; intermediate stars excluded.
\end{itemize}
After per-task filtering and the token-length-matching constraint, the resulting yield is $|S| = 145$ (\data{CoLA}), $\leq 500$ (\data{MNLI}; we cap at $500$), $189$ (\data{STS-B}), $150$--$200$ (\data{RewardBench}), and $145$--$200$ (\data{Yelp}).

\paragraph{Backward-pass tracing budget.}
Dense backward-pass tracing has quadratic attention overhead in sequence length and is the binding cost for end-to-end attribution at the architectures we consider: 
mapping the full \data{CoLA} judgment graph in Gemma-3-12B requires evaluating approximately 1.46 million candidate edges, and Gemma-3-27B contains roughly 50,000 components. 
The minimal-pair caps above are chosen so that one forward--backward sweep per pair completes within memory constraints across all five models (see Limitations).

\section{Ablation Study}
\label{app:ablation_study}
To evaluate the functional importance of the causally identified circuit components at the strictest level, we perform a resampling ablation study within the Latent Evaluator.
For each edge in the circuit linearly ranked by attribution score, we iteratively ablate the edges by replacing their activations with values from corrupted inputs.
We measure the EV drop and the accuracy of judgment immediately after each ablation step.

Circuit robustness varies substantially across tasks: \data{STS-B} classification is the most robust, while \data{MNLI} judgment is extremely fragile, with accuracy often dropping after ablating only the single top-ranked edge. 
Model scale also influences robustnes: smaller models (e.g., Qwen2.5-7B) have notably less robust judge circuits than larger models (e.g., Gemma-3-27B). 
All evaluation tasks demonstrate a characteristic phase transition, where accuracy remains relatively stable up to a critical edge-ablation threshold, beyond which performance collapses completely. 
Crucially, classification subtasks are consistently much more robust than their rating counterparts.
This points to redundant processing pathways in the classification routers, whereas rating circuits compress into a few concentrated bottleneck heads. 

\begin{figure*}[t]
    \centering
    \begin{subfigure}{0.48\textwidth}
        \centering
        \includegraphics[width=\linewidth]{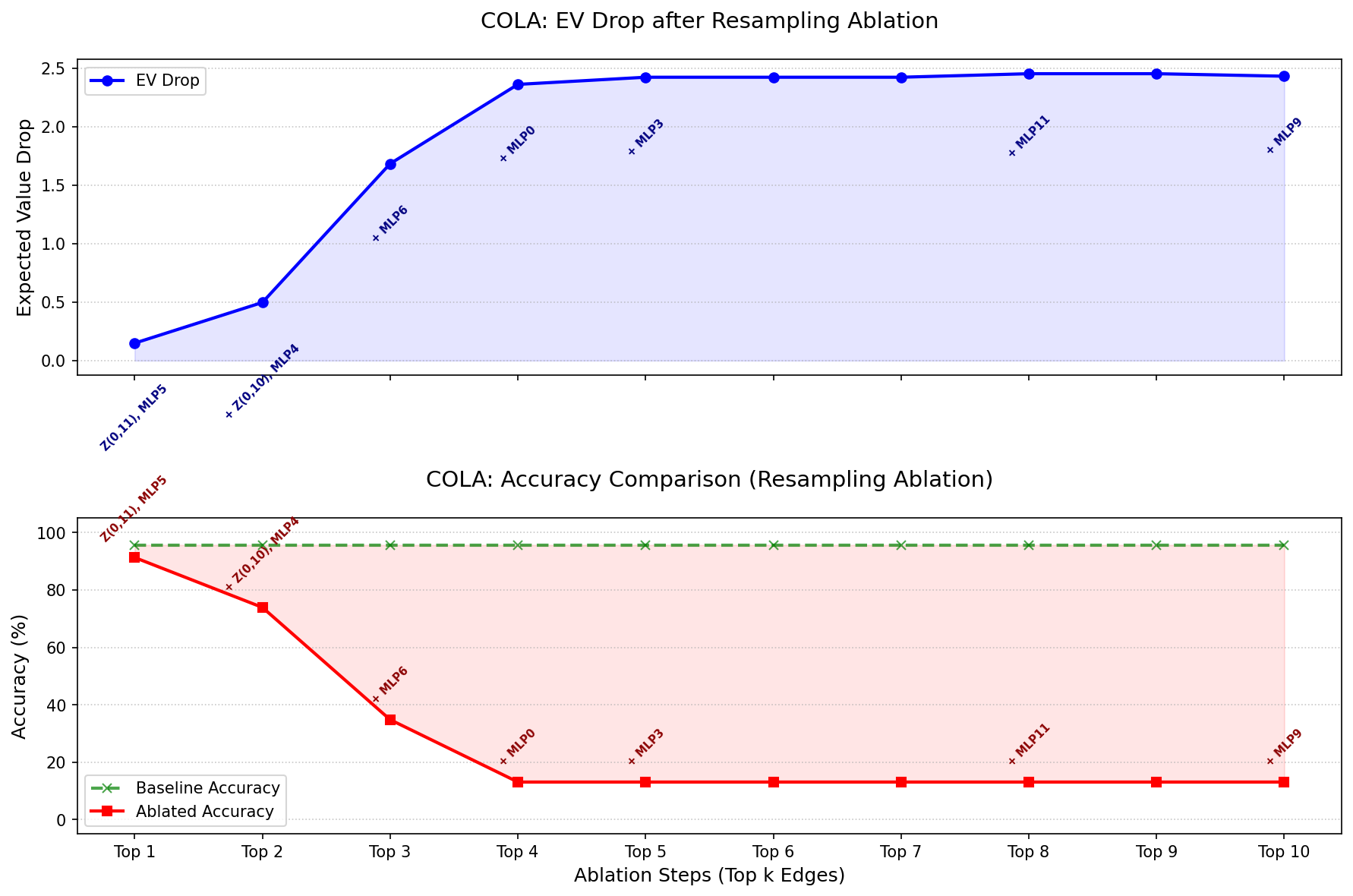}
        \caption{Classification Ablation.}
        \label{fig:ablation_gemma3_12_CoLA_class}
    \end{subfigure}\hfill
    \begin{subfigure}{0.48\textwidth}
        \centering
        \includegraphics[width=\linewidth]{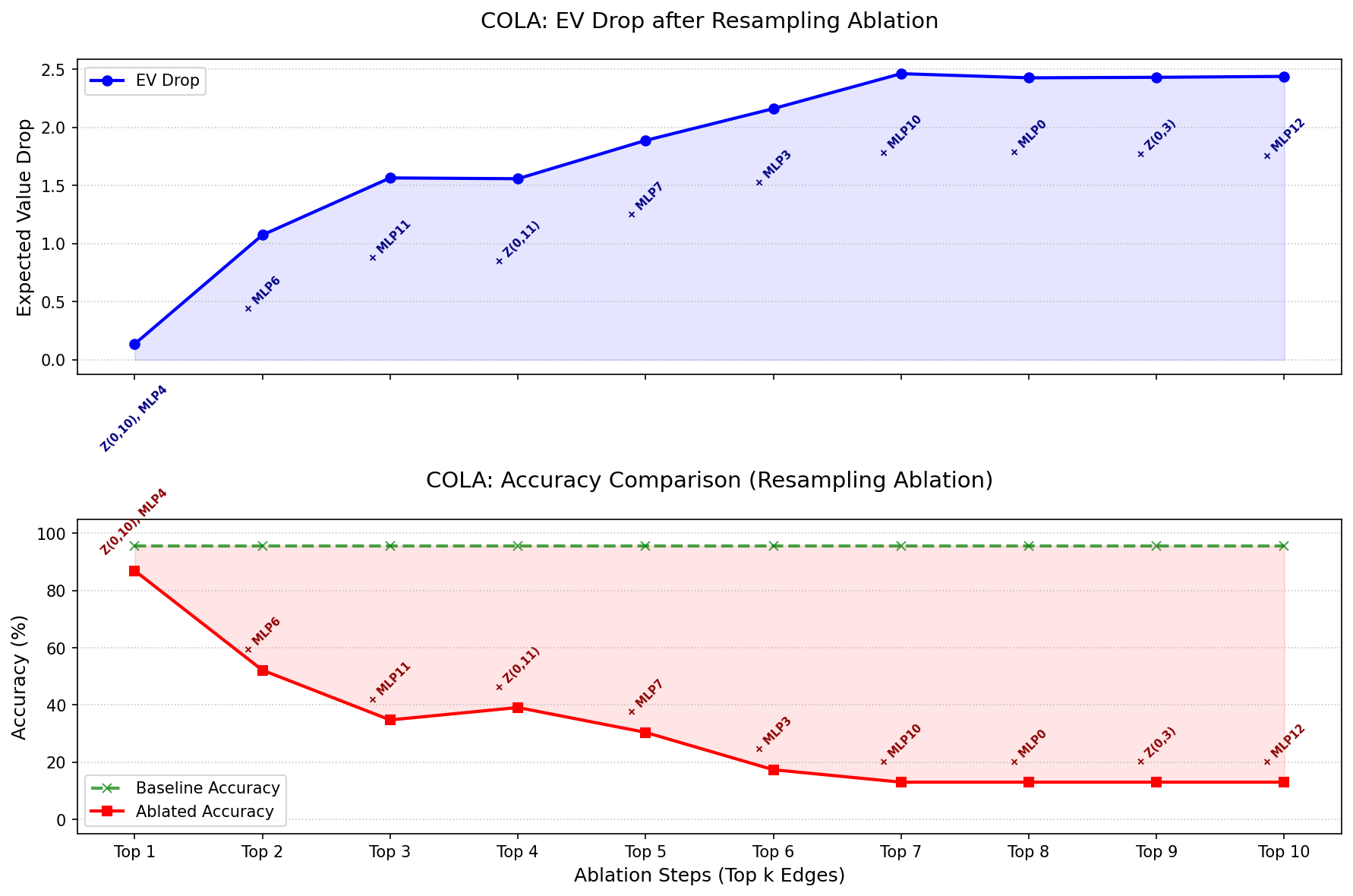}
        \caption{Numerical Judgment Ablation.}
        \label{fig:ablation_gemma3_12_cola_num}
    \end{subfigure}
    \centering\textbf{COLA dataset.}

    \begin{subfigure}{0.48\textwidth}
        \centering
        \includegraphics[width=\linewidth]{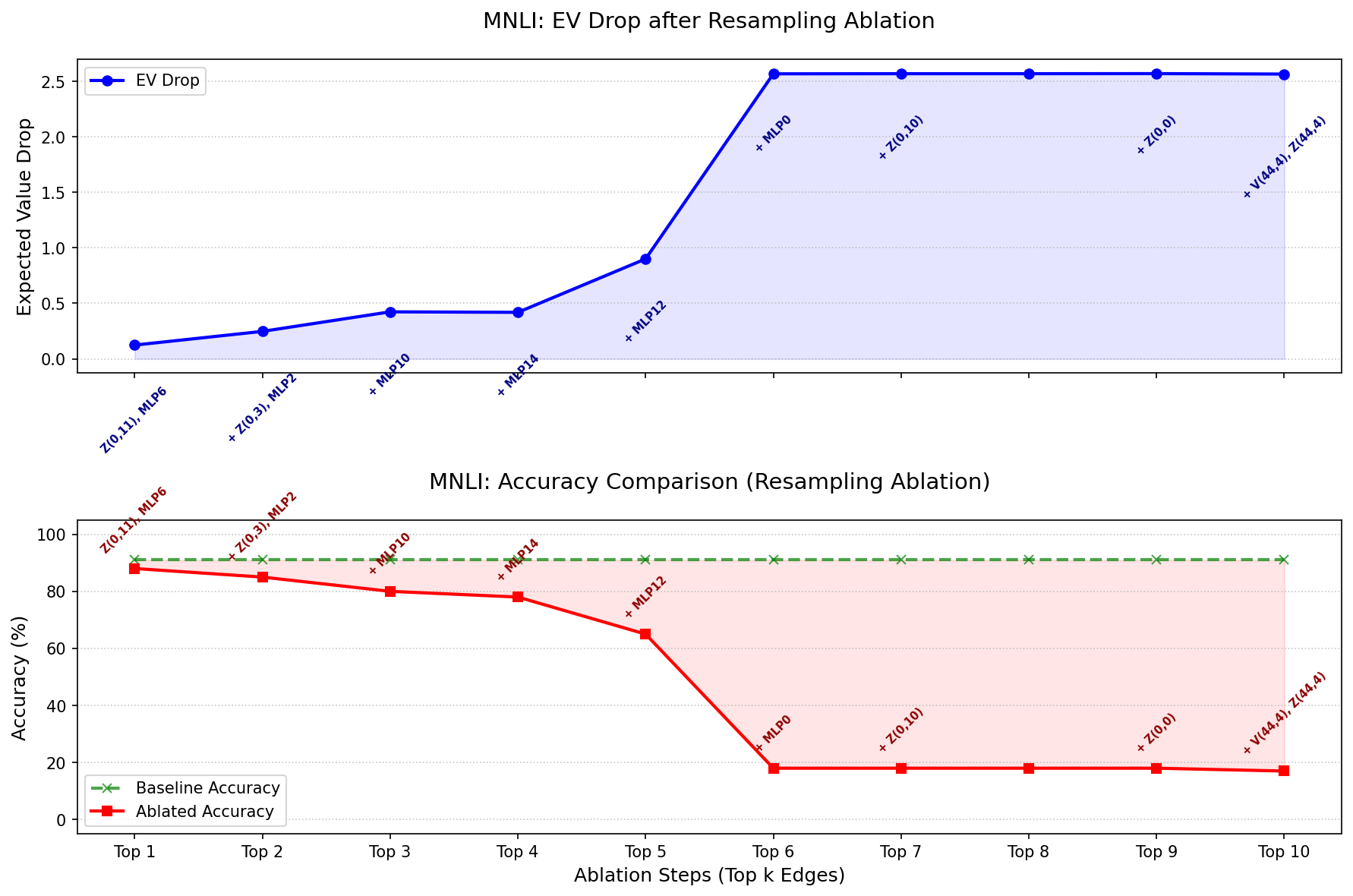}
        \caption{Classification Ablation.}
        \label{fig:ablation_gemma3_12_mnli_class}
    \end{subfigure}\hfill
    \begin{subfigure}{0.48\textwidth}
        \centering
        \includegraphics[width=\linewidth]{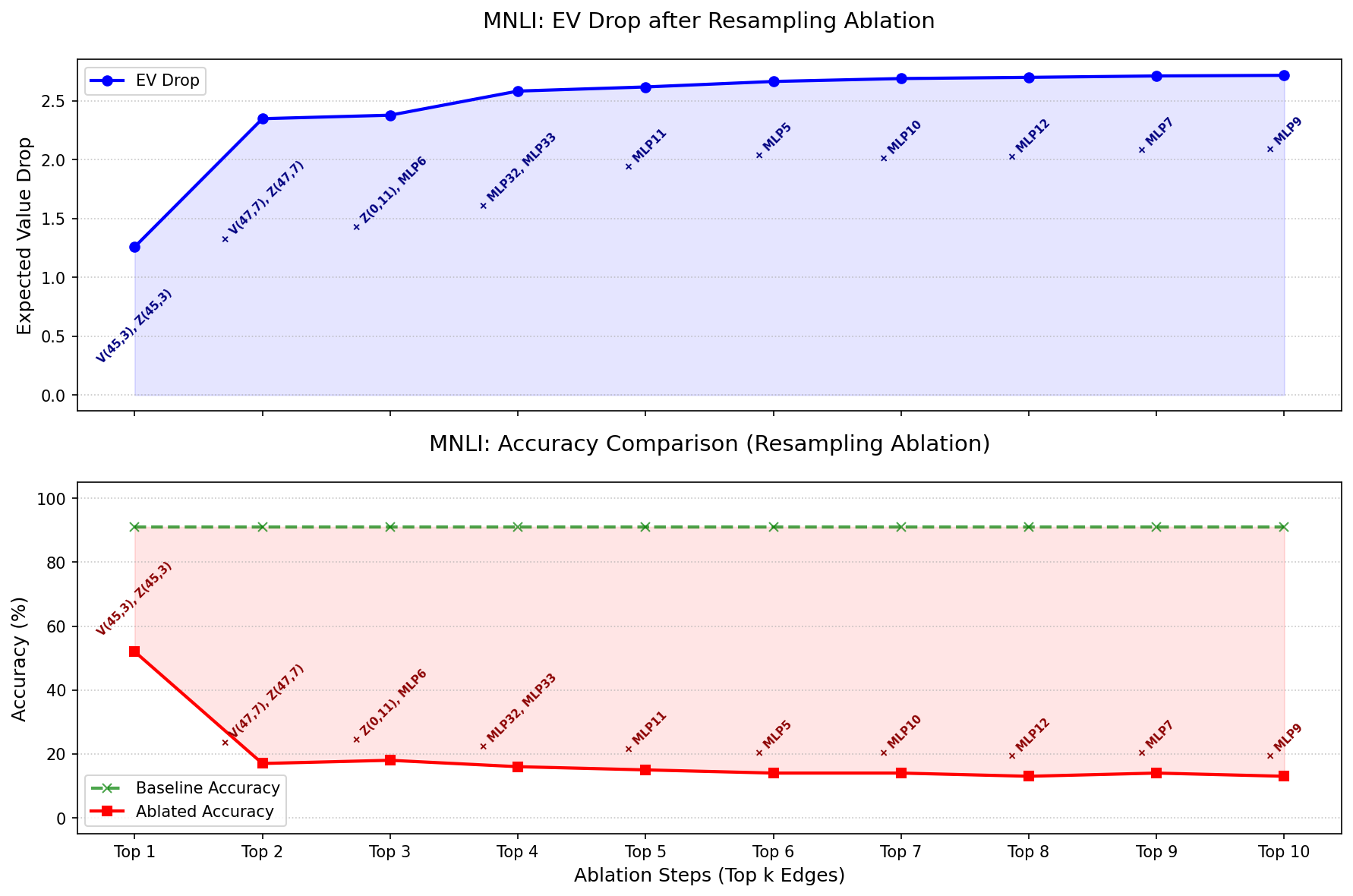}
        \caption{Numerical Judgment Ablation.}
        \label{fig:ablation_gemma3_12_mnli_num}
    \end{subfigure}
    \centering\textbf{MNLI dataset.}

    \begin{subfigure}{0.48\textwidth}
        \centering
        \includegraphics[width=\linewidth]{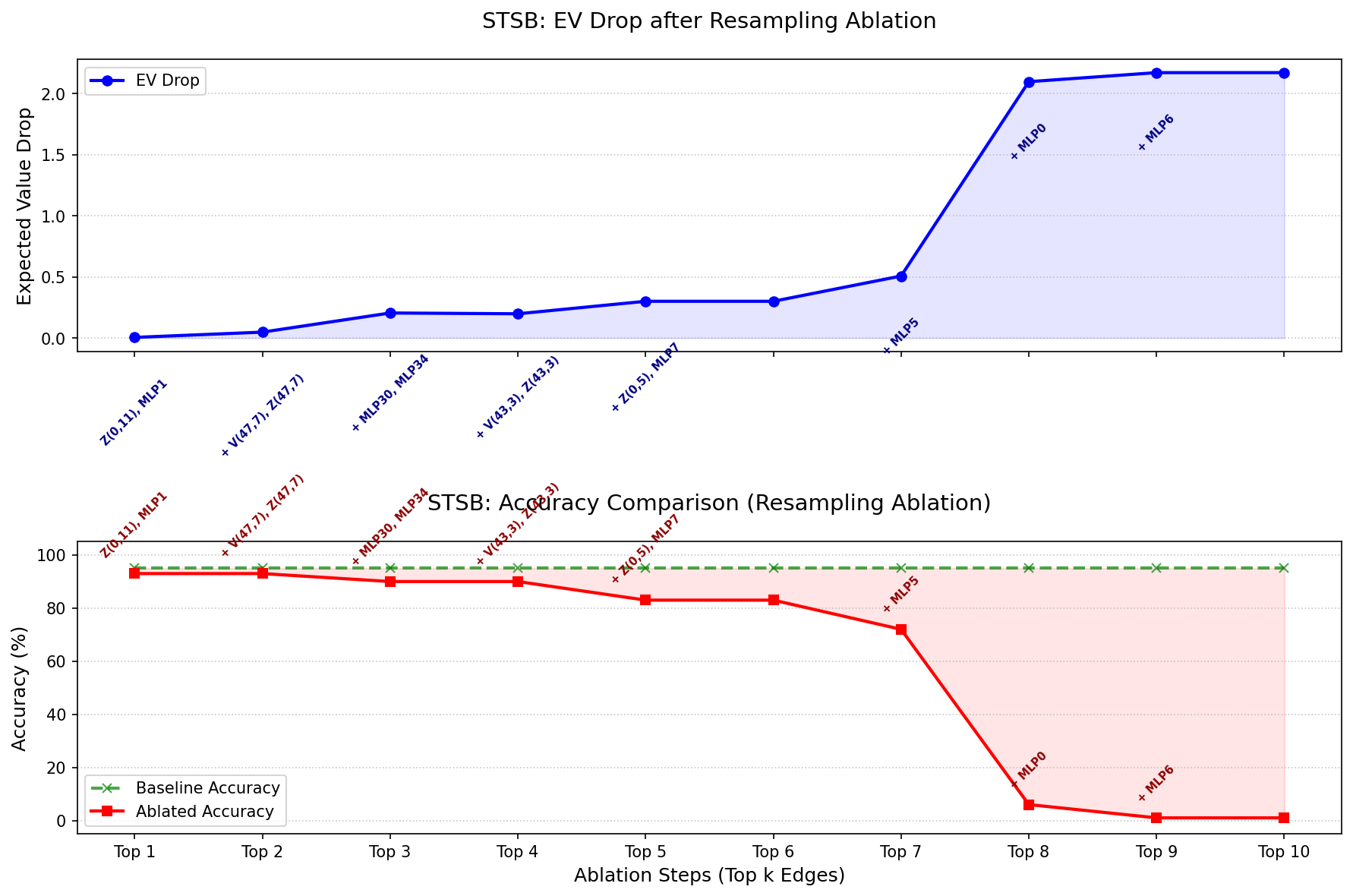}
        \caption{Classification Ablation.}
        \label{fig:ablation_gemma3_12_stsb_class}
    \end{subfigure}\hfill
    \begin{subfigure}{0.48\textwidth}
        \centering
        \includegraphics[width=\linewidth]{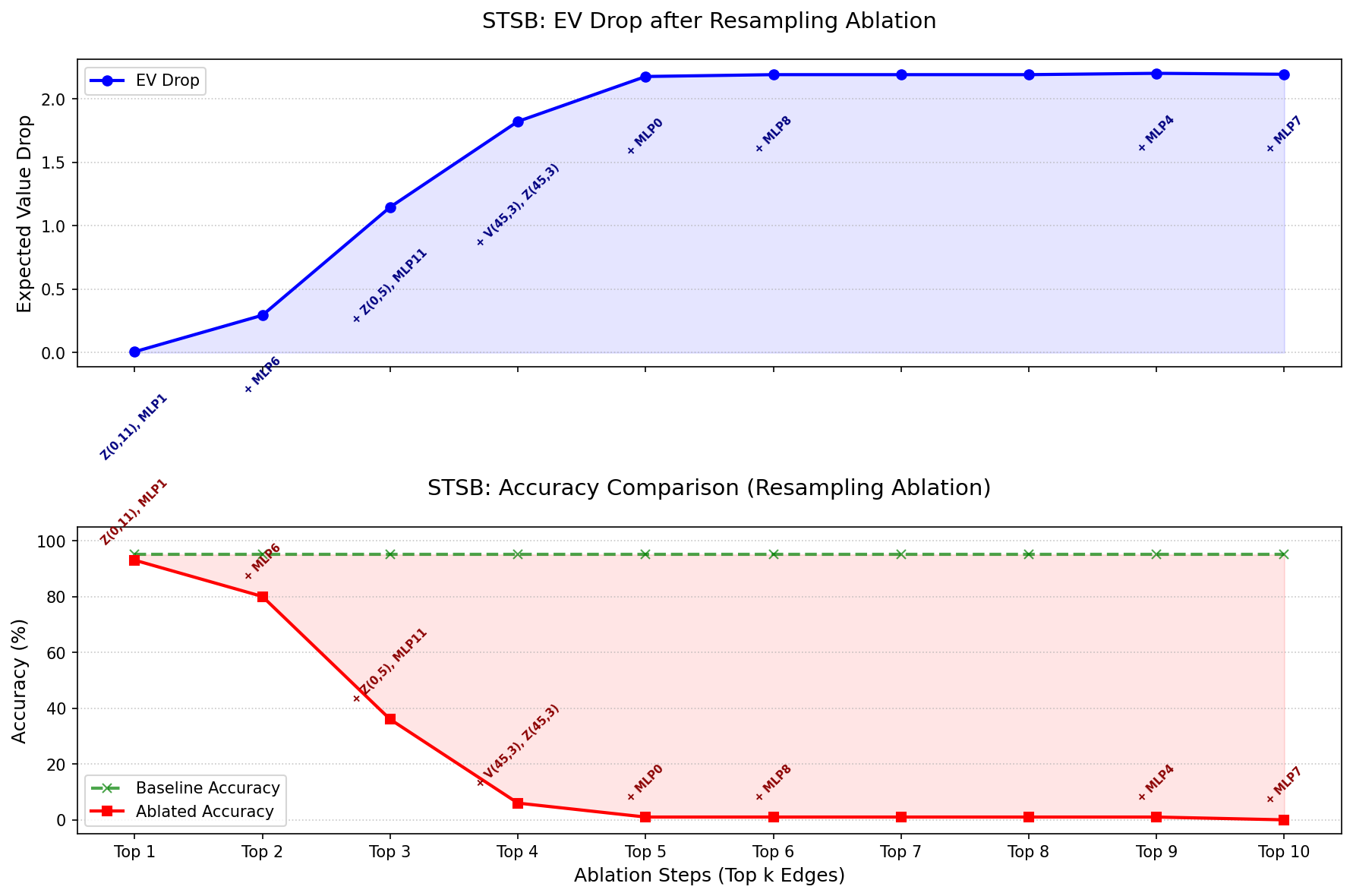}
        \caption{Numerical Judgment Ablation.}
        \label{fig:ablation_gemma3_12_stsb_num}
    \end{subfigure}
    \centering\textbf{STSB dataset.}

    \caption{Ablation phase-transition study (Gemma-3-12B).}
    \label{fig:ablation_study_gemma3_12}
    \vspace*{-1em}
\end{figure*}

\begin{figure*}[t]
    \centering
    \begin{subfigure}{0.48\textwidth}
        \centering
        \includegraphics[width=\linewidth]{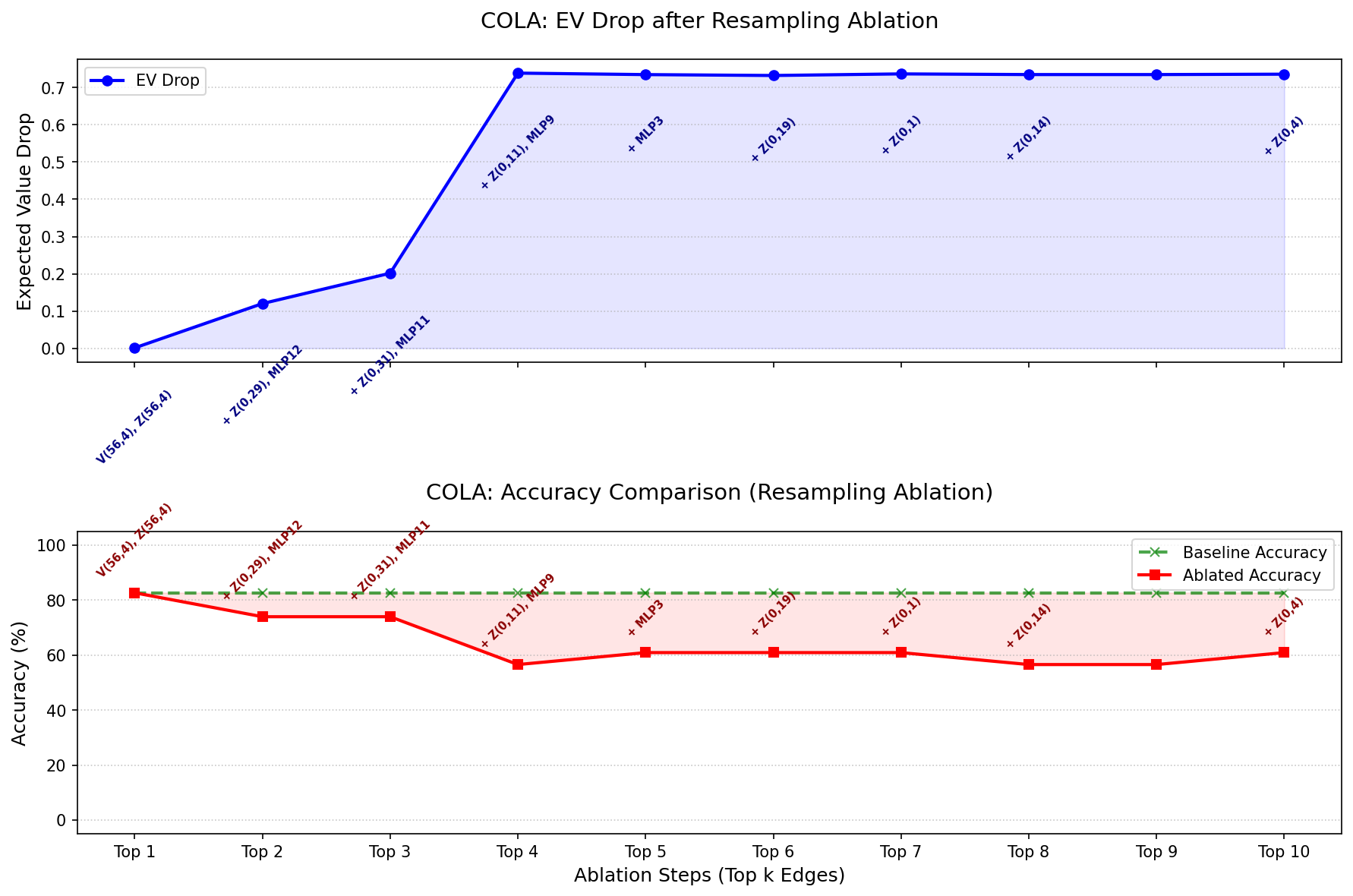}
        \caption{Classification Ablation.}
        \label{fig:ablation_gemma3_27_cola_class}
    \end{subfigure}\hfill
    \begin{subfigure}{0.48\textwidth}
        \centering
        \includegraphics[width=\linewidth]{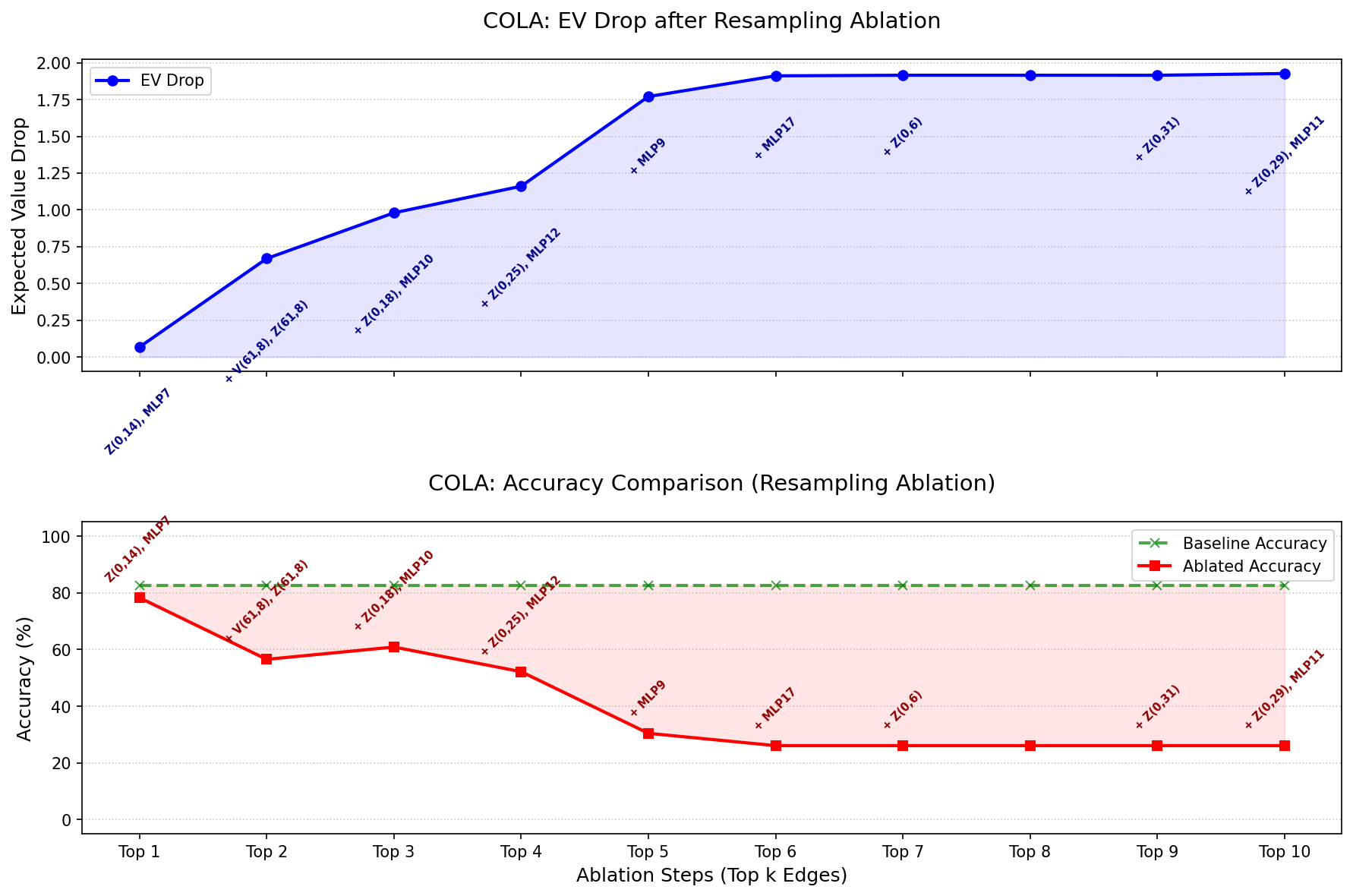}
        \caption{Numerical Judgment Ablation.}
        \label{fig:ablation_gemma3_27_cola_num}
    \end{subfigure}
    \centering\textbf{COLA dataset.}

    \begin{subfigure}{0.48\textwidth}
        \centering
        \includegraphics[width=\linewidth]{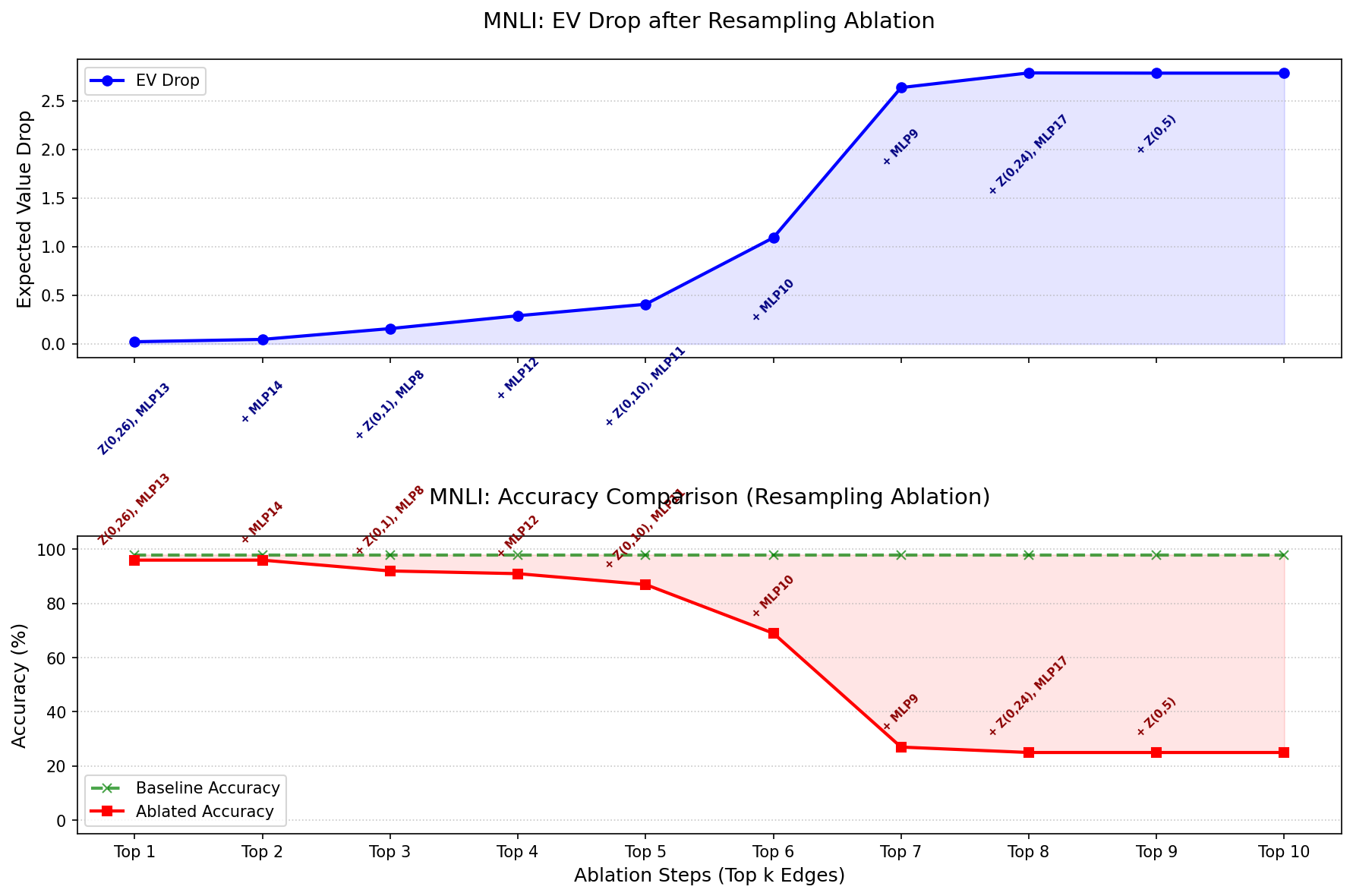}
        \caption{Classification Ablation.}
        \label{fig:ablation_gemma3_27_mnli_class}
    \end{subfigure}\hfill
    \begin{subfigure}{0.48\textwidth}
        \centering
        \includegraphics[width=\linewidth]{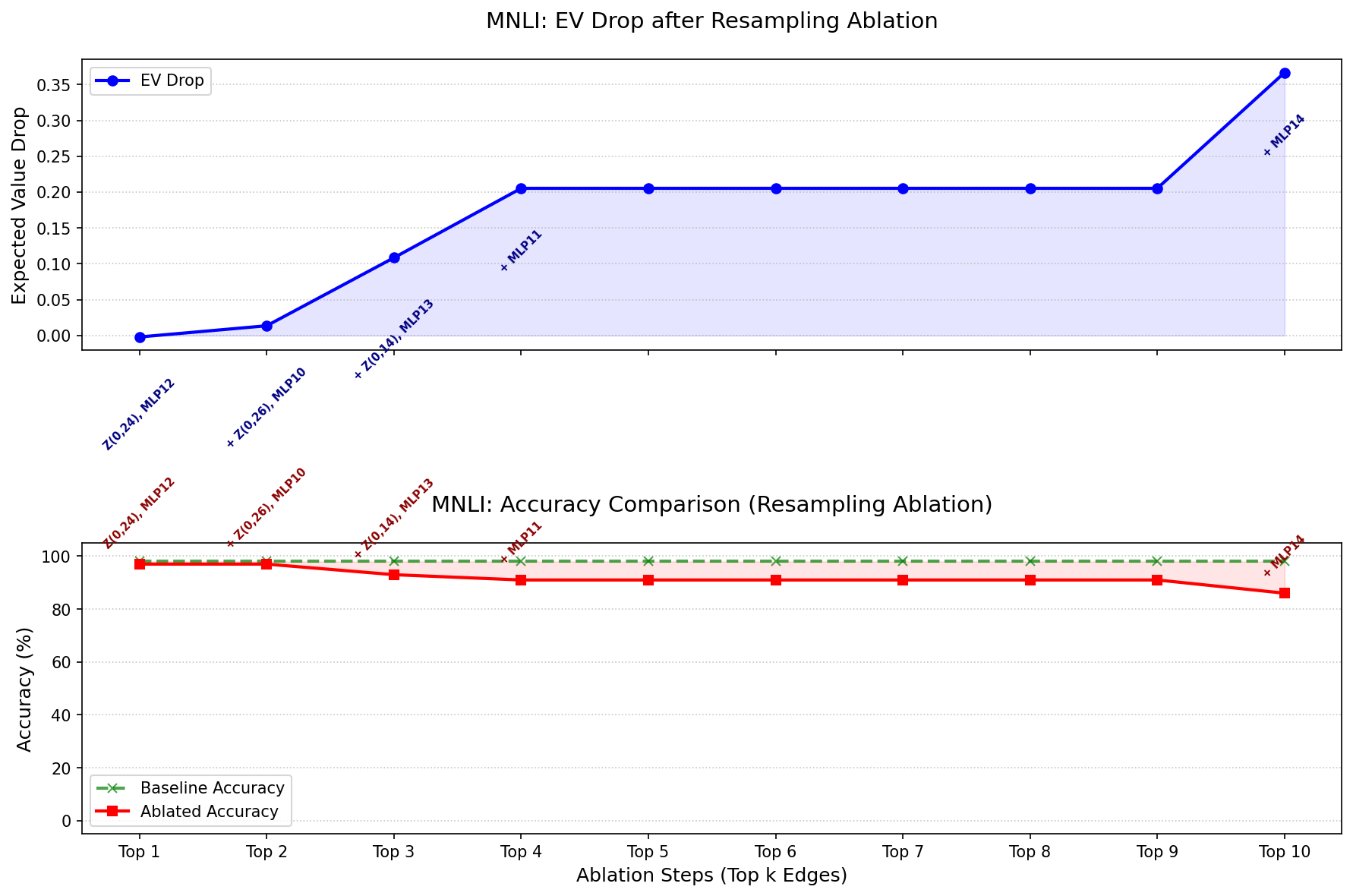}
        \caption{Numerical Judgment Ablation.}
        \label{fig:ablation_gemma3_27_mnli_num}
    \end{subfigure}
    \centering\textbf{MNLI dataset.}

    \begin{subfigure}{0.48\textwidth}
        \centering
        \includegraphics[width=\linewidth]{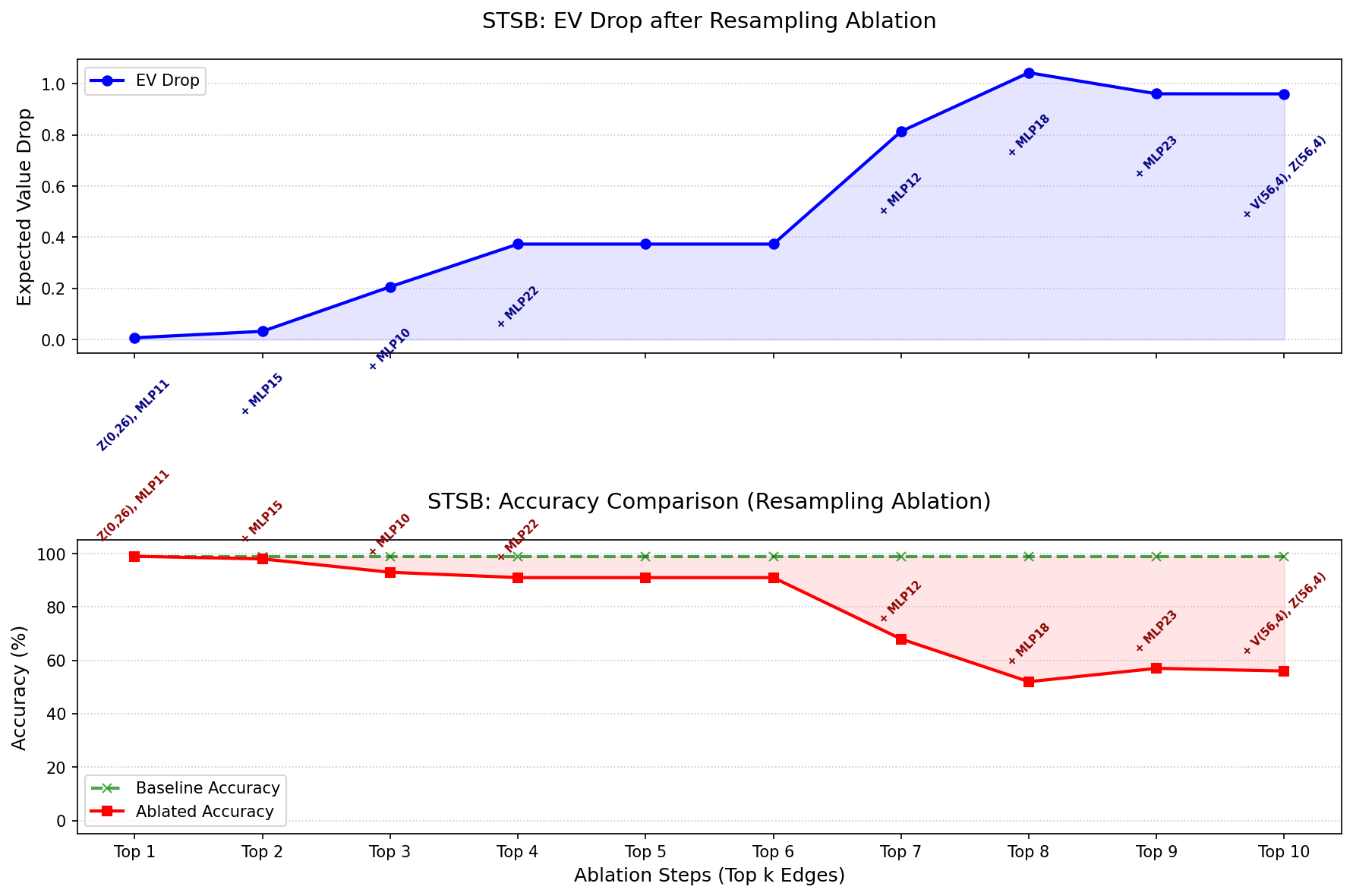}
        \caption{Classification Ablation.}
        \label{fig:ablation_gemma3_27_stsb_class}
    \end{subfigure}\hfill
    \begin{subfigure}{0.48\textwidth}
        \centering
        \includegraphics[width=\linewidth]{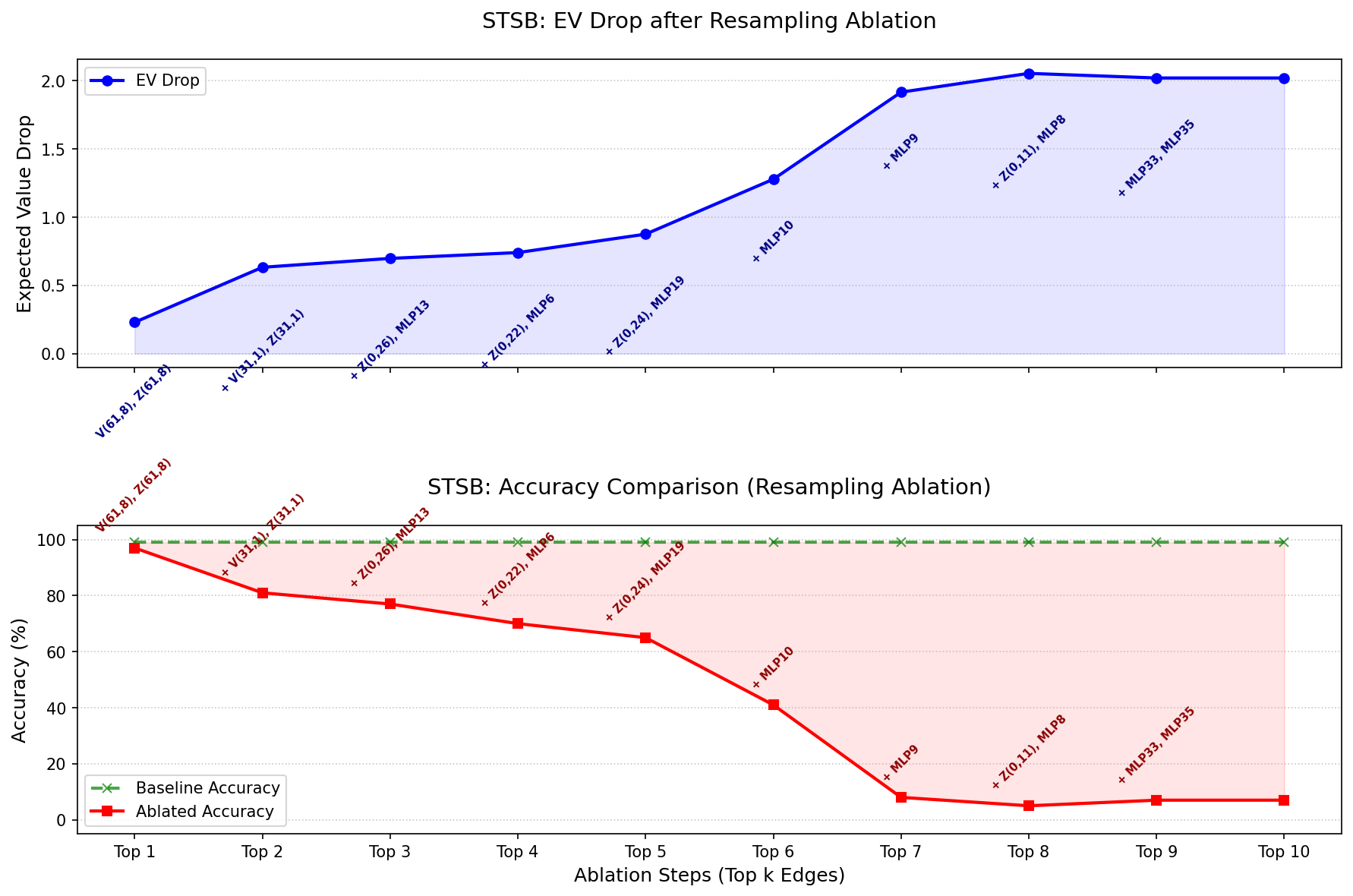}
        \caption{Numerical Judgment Ablation.}
        \label{fig:ablation_gemma3_27_stsb_num}
    \end{subfigure}
    \centering\textbf{STSB dataset.}

    \caption{Ablation phase-transition study (Gemma-3-27B).}
    \label{fig:ablation_study_gemma3_27}
    \vspace*{-1em}
\end{figure*}

\begin{figure*}[t]
    \centering
    \begin{subfigure}{0.48\textwidth}
        \centering
        \includegraphics[width=\linewidth]{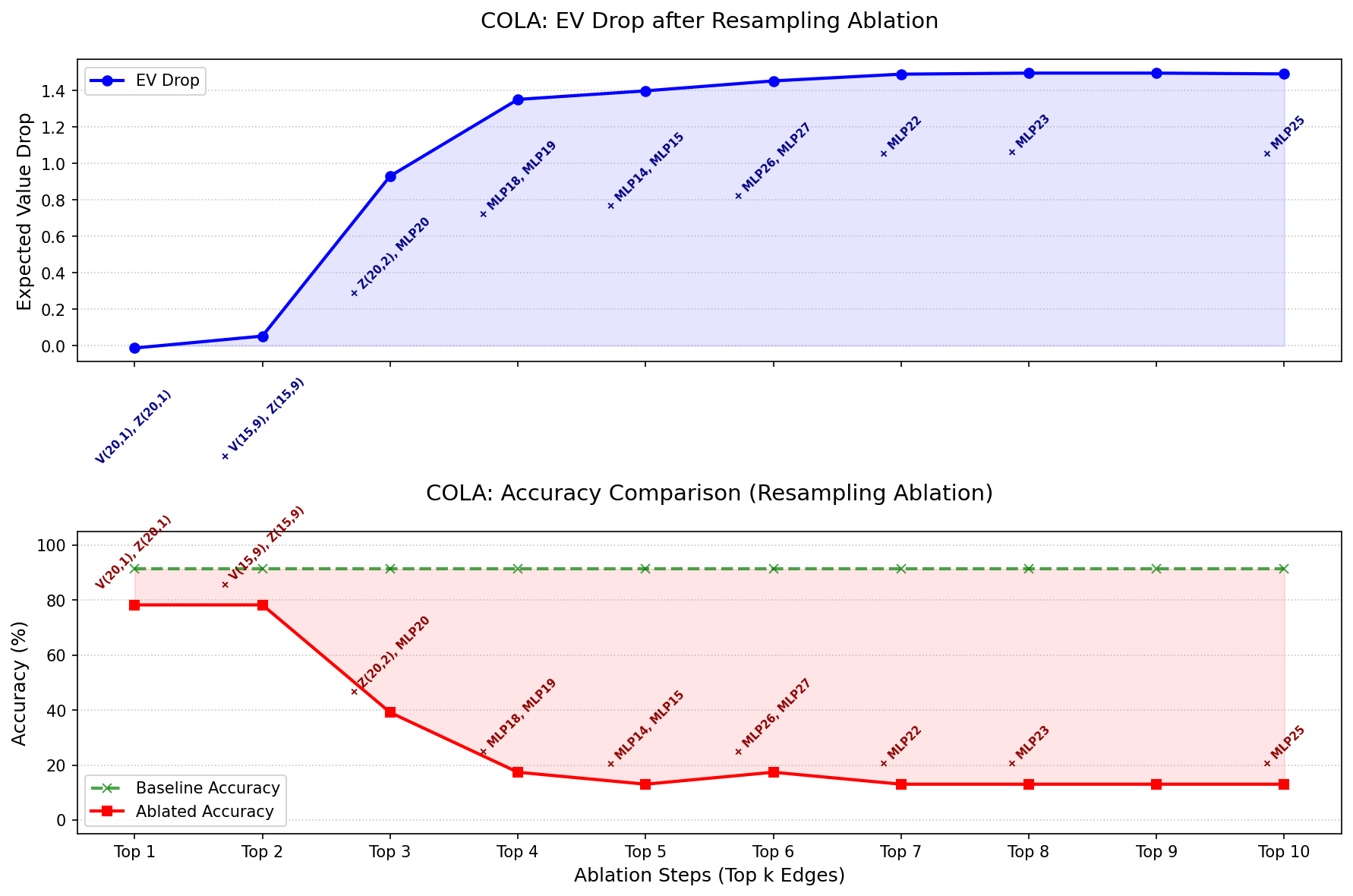}
        \caption{Classification Ablation.}
        \label{fig:ablation_qwen25_7_cola_class}
    \end{subfigure}\hfill
    \begin{subfigure}{0.48\textwidth}
        \centering
        \includegraphics[width=\linewidth]{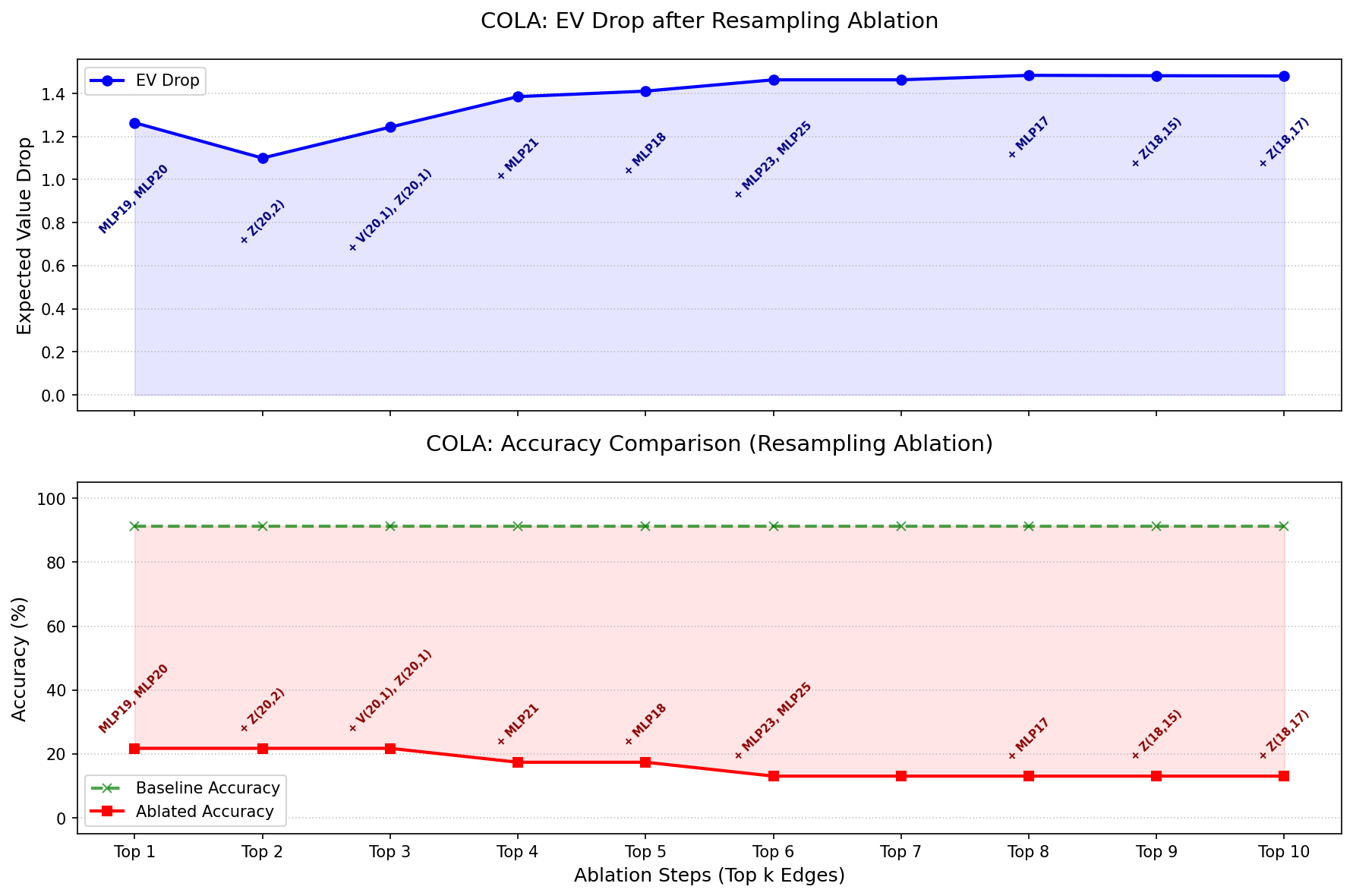}
        \caption{Numerical Judgment Ablation.}
        \label{fig:ablation_qwen25_7_cola_num}
    \end{subfigure}
    \centering\textbf{COLA dataset.}

    \begin{subfigure}{0.48\textwidth}
        \centering
        \includegraphics[width=\linewidth]{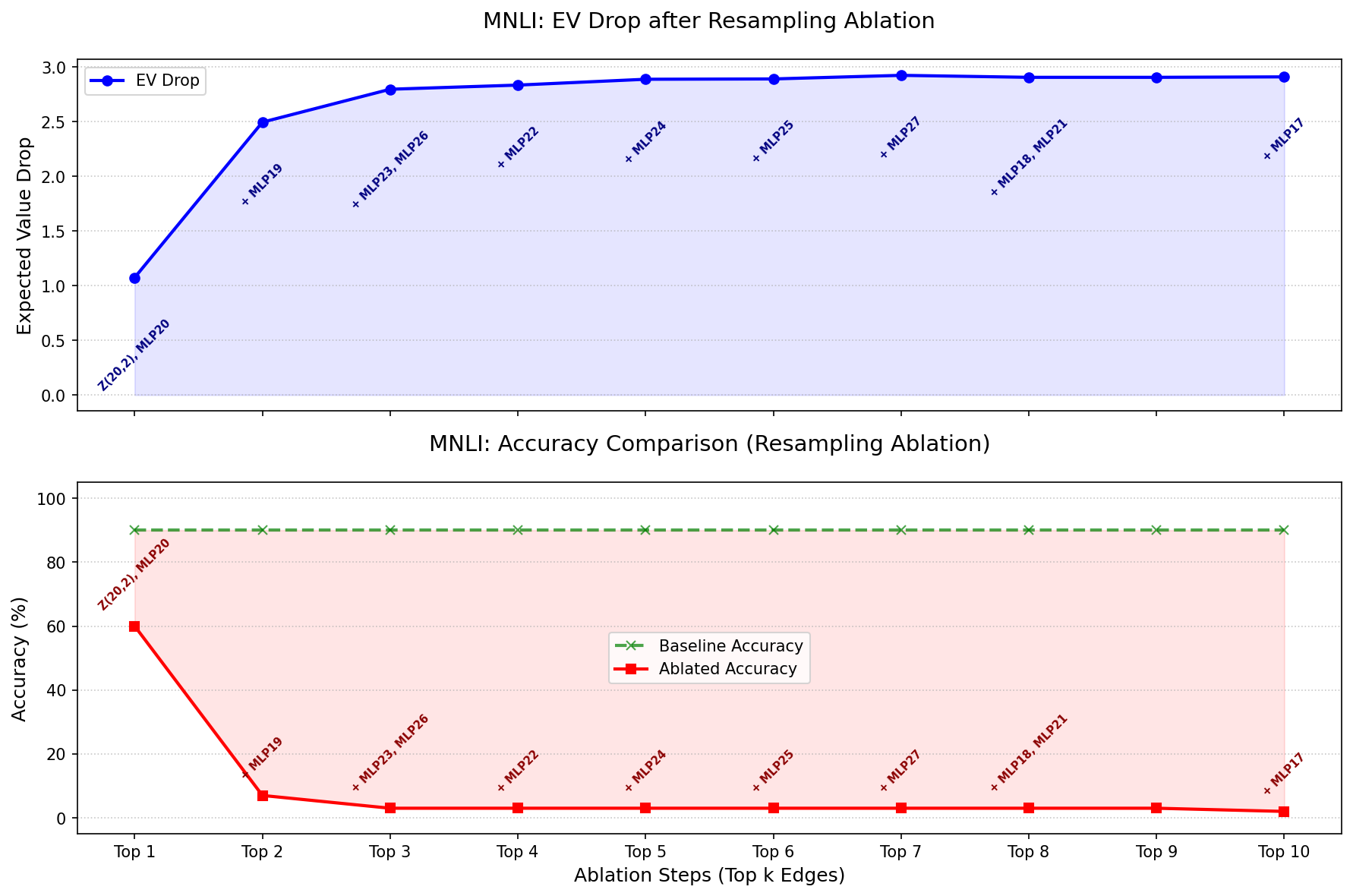}
        \caption{Classification Ablation.}
        \label{fig:ablation_qwen25_7_mnli_class}
    \end{subfigure}\hfill
    \begin{subfigure}{0.48\textwidth}
        \centering
        \includegraphics[width=\linewidth]{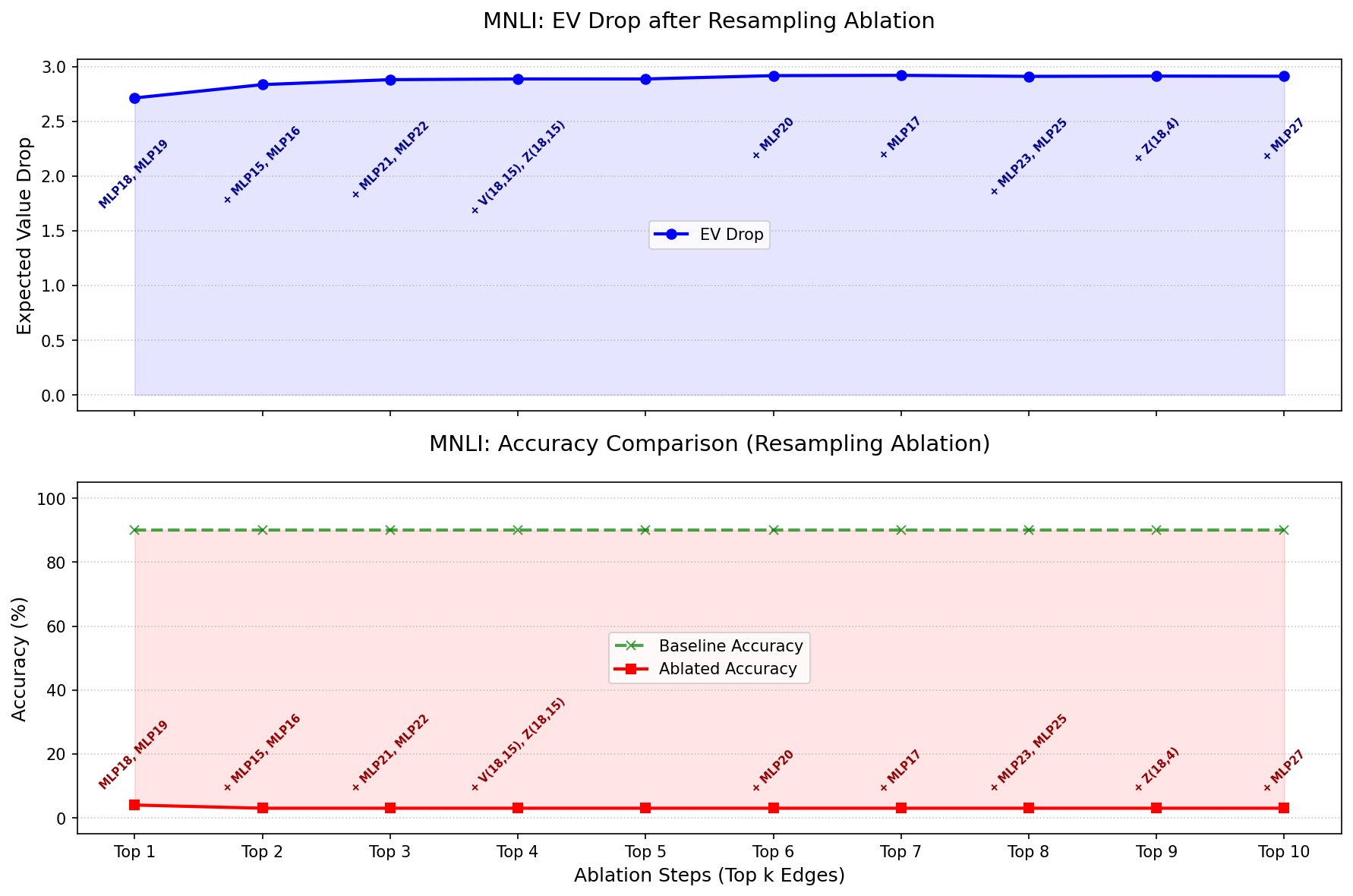}
        \caption{Numerical Judgment Ablation.}
        \label{fig:ablation_qwen25_7_mnli_num}
    \end{subfigure}
    \centering\textbf{MNLI dataset.}

    \begin{subfigure}{0.48\textwidth}
        \centering
        \includegraphics[width=\linewidth]{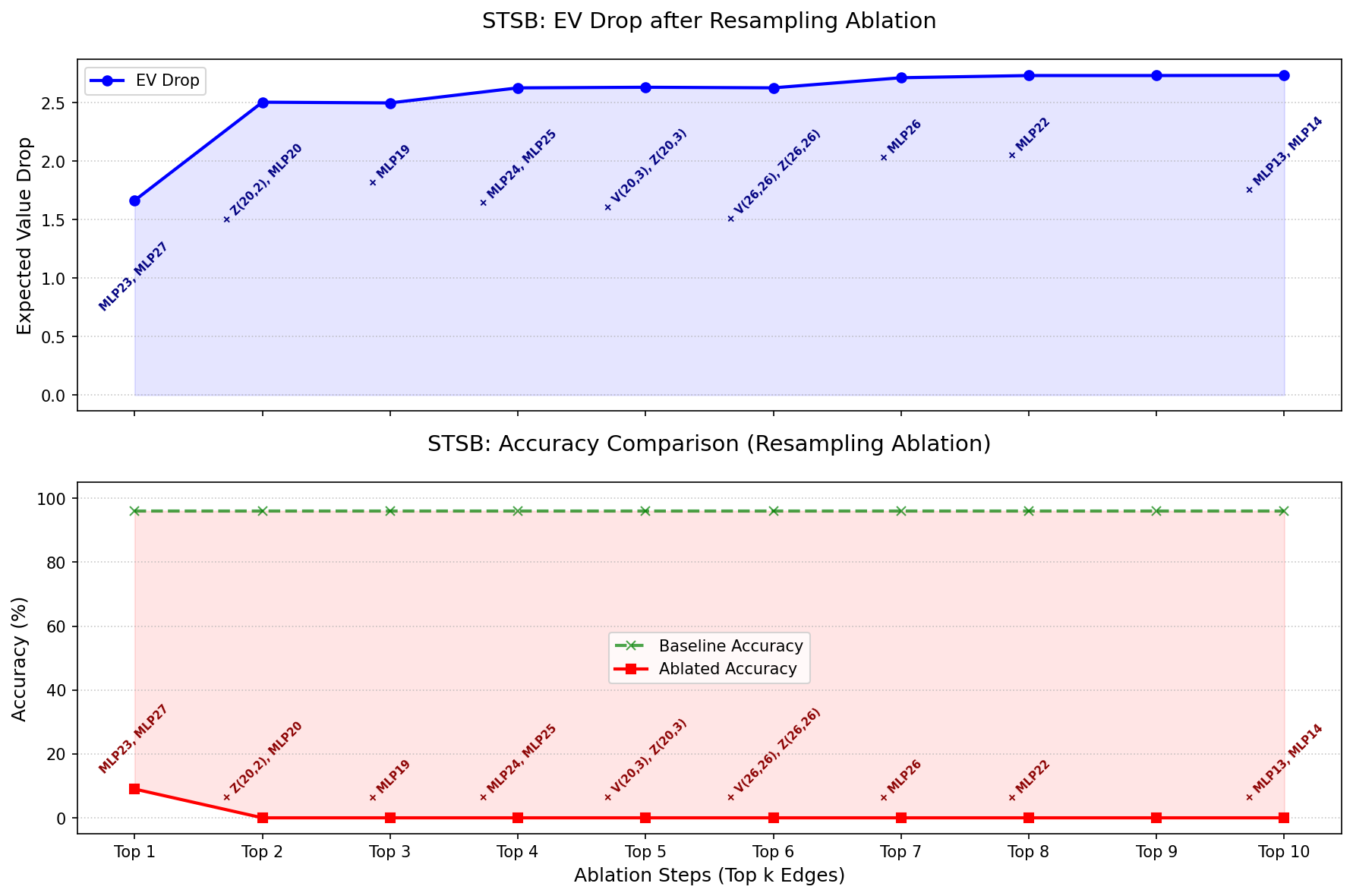}
        \caption{Classification Ablation.}
        \label{fig:ablation_qwen25_7_stsb_class}
    \end{subfigure}\hfill
    \begin{subfigure}{0.48\textwidth}
        \centering
        \includegraphics[width=\linewidth]{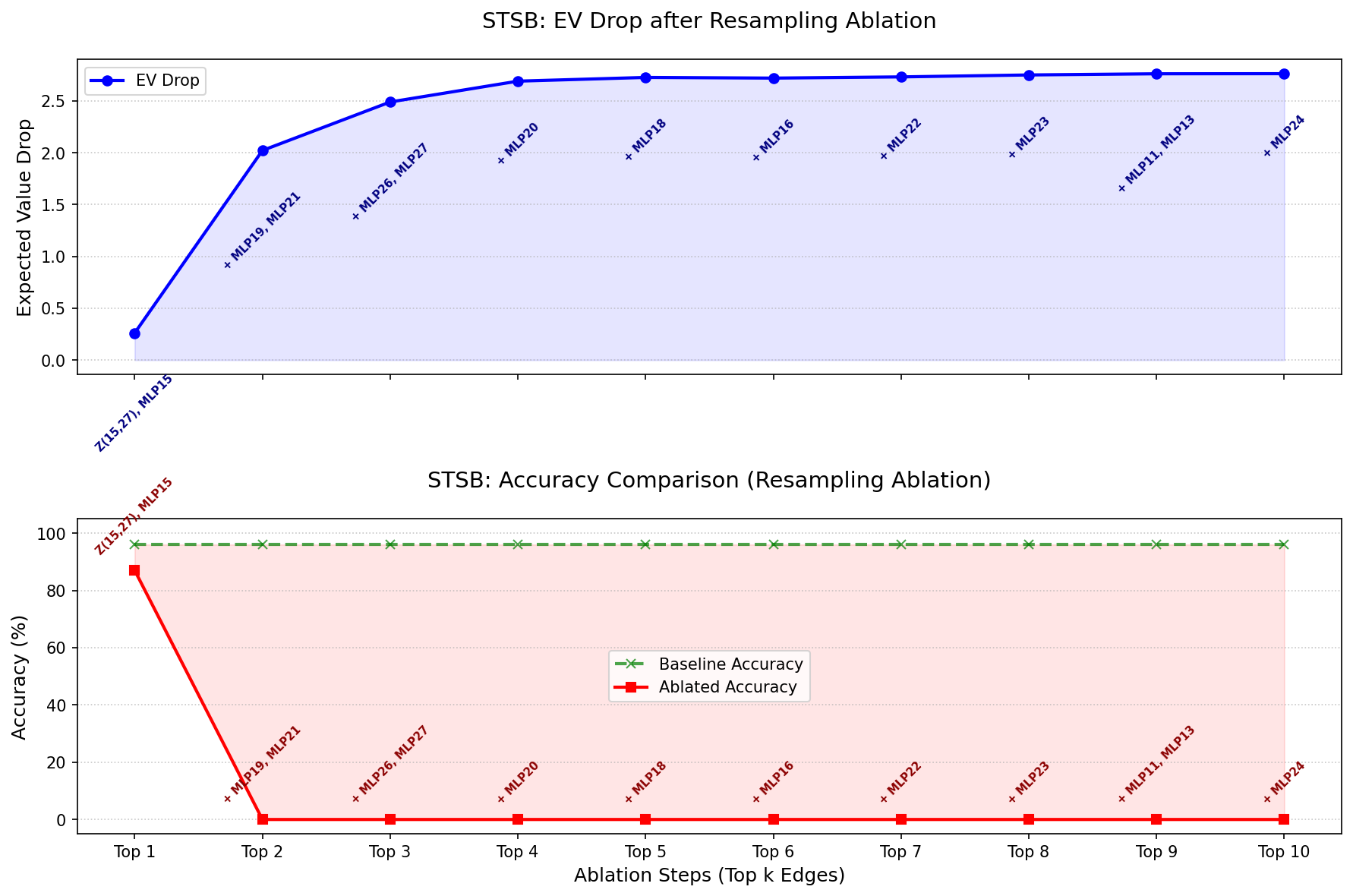}
        \caption{Numerical Judgment Ablation.}
        \label{fig:ablation_qwen25_7_stsb_num}
    \end{subfigure}
    \centering\textbf{STSB dataset.}

    \caption{Ablation phase-transition study (Qwen2.5-7B-Instruct).}
    \label{fig:ablation_study_qwen25_7}
    \vspace*{-1em}
\end{figure*}

\begin{figure*}[t]
    \centering
    \begin{subfigure}{0.48\textwidth}
        \centering
        \includegraphics[width=\linewidth]{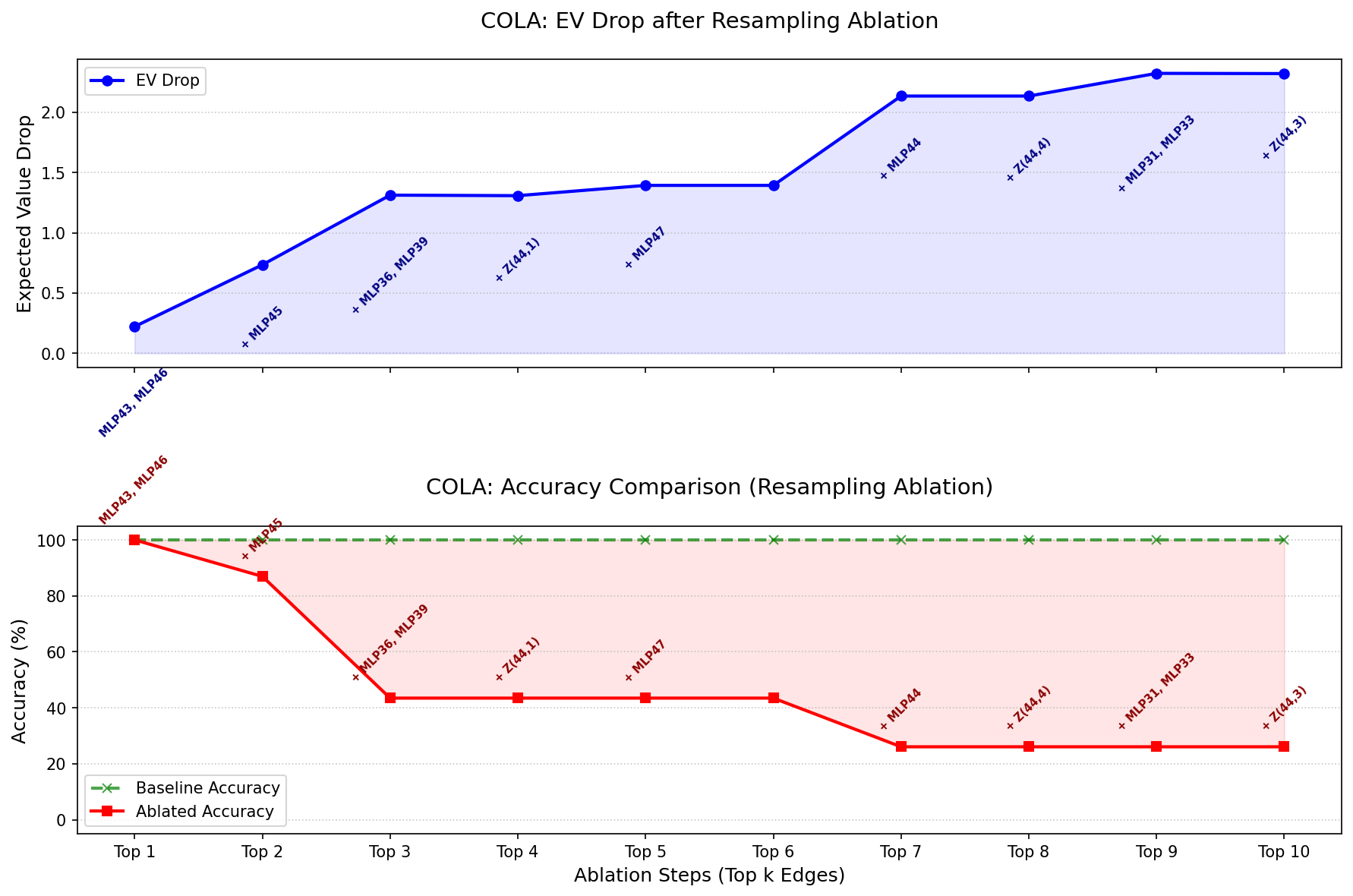}
        \caption{Classification Ablation.}
        \label{fig:ablation_qwen25_14_cola_class}
    \end{subfigure}\hfill
    \begin{subfigure}{0.48\textwidth}
        \centering
        \includegraphics[width=\linewidth]{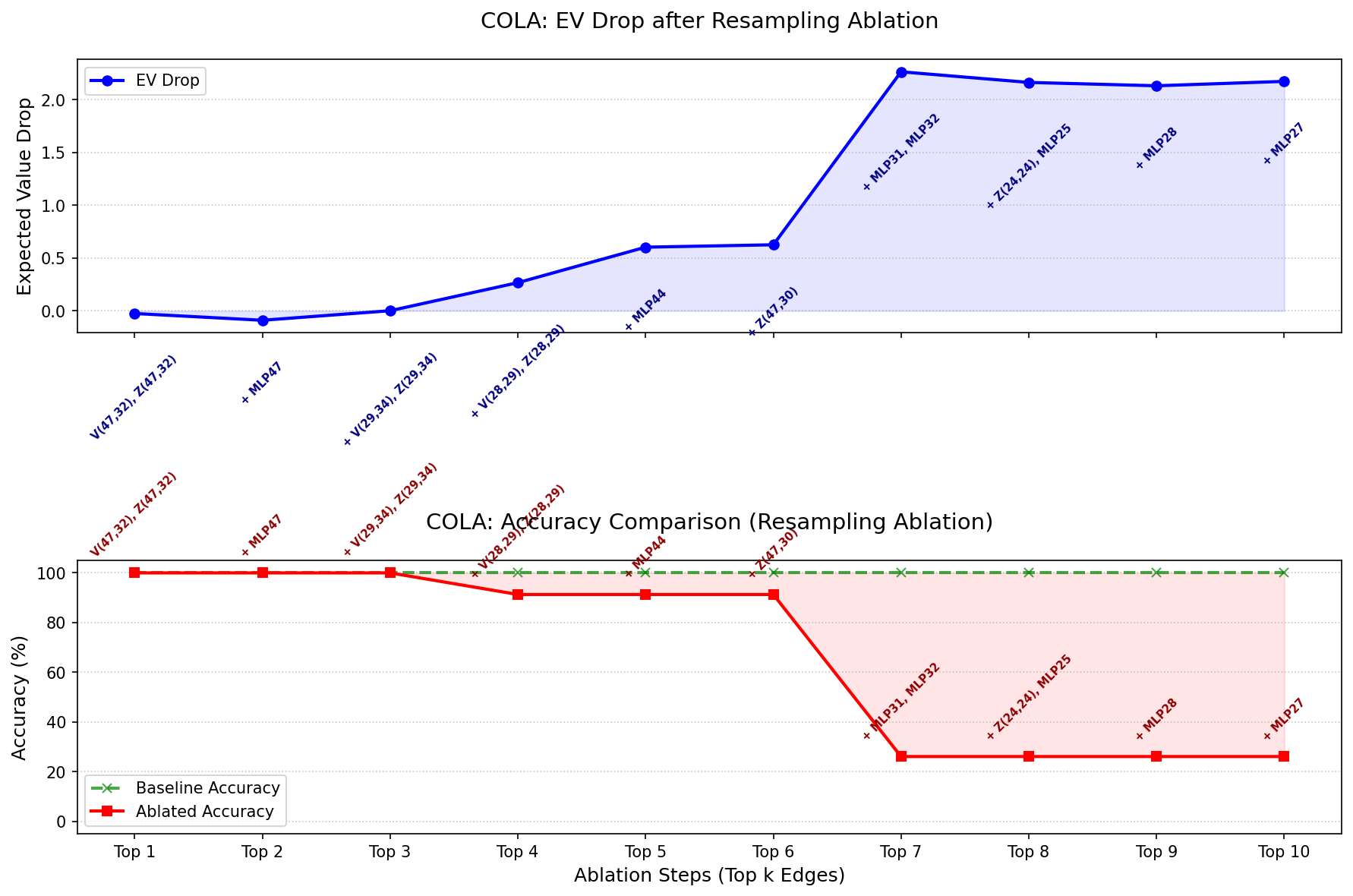}
        \caption{Numerical Judgment Ablation.}
        \label{fig:ablation_qwen25_14_cola_num}
    \end{subfigure}
    \centering\textbf{COLA dataset.}

    \begin{subfigure}{0.48\textwidth}
        \centering
        \includegraphics[width=\linewidth]{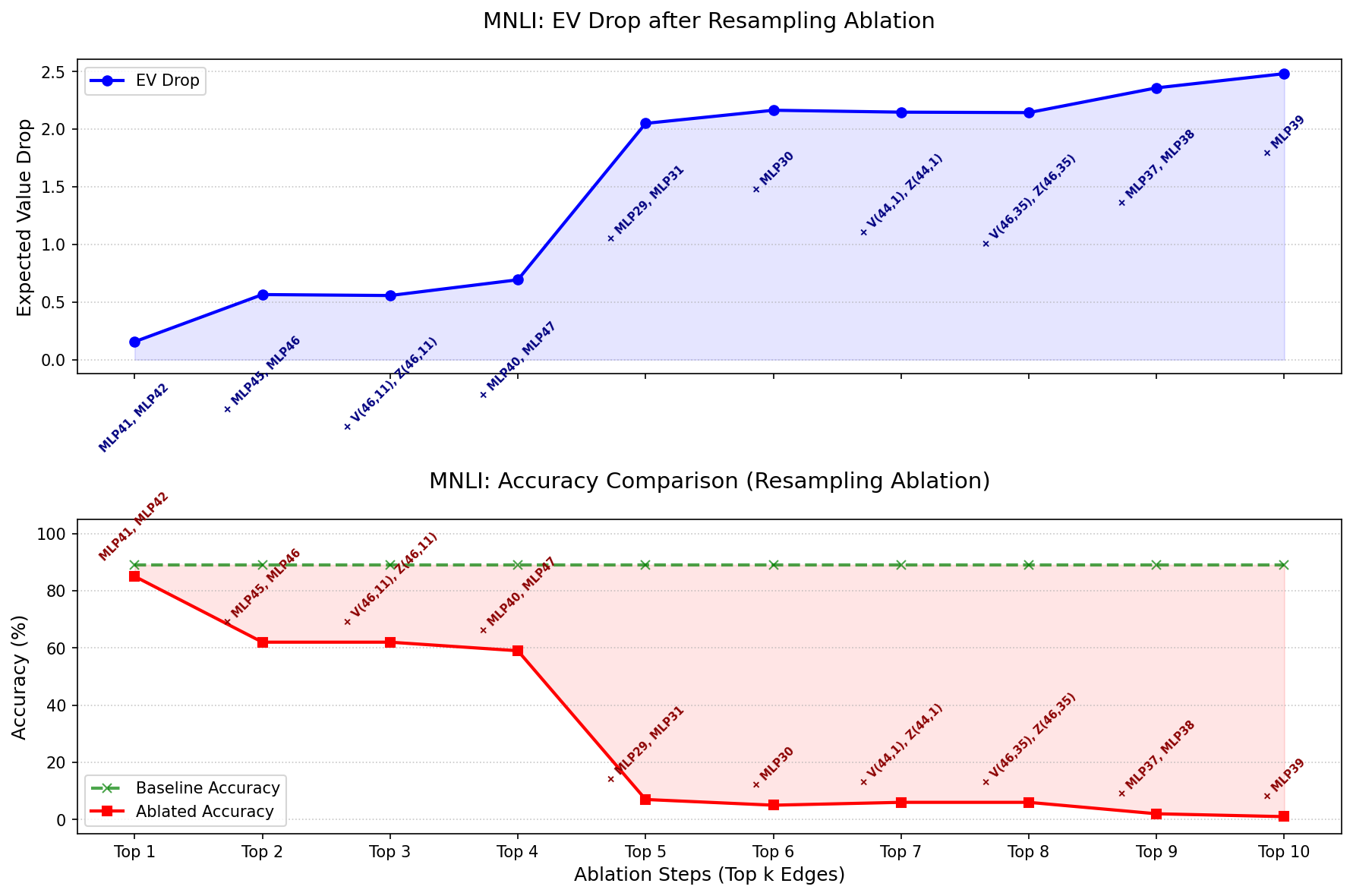}
        \caption{Classification Ablation.}
        \label{fig:ablation_qwen25_14_mnli_class}
    \end{subfigure}\hfill
    \begin{subfigure}{0.48\textwidth}
        \centering
        \includegraphics[width=\linewidth]{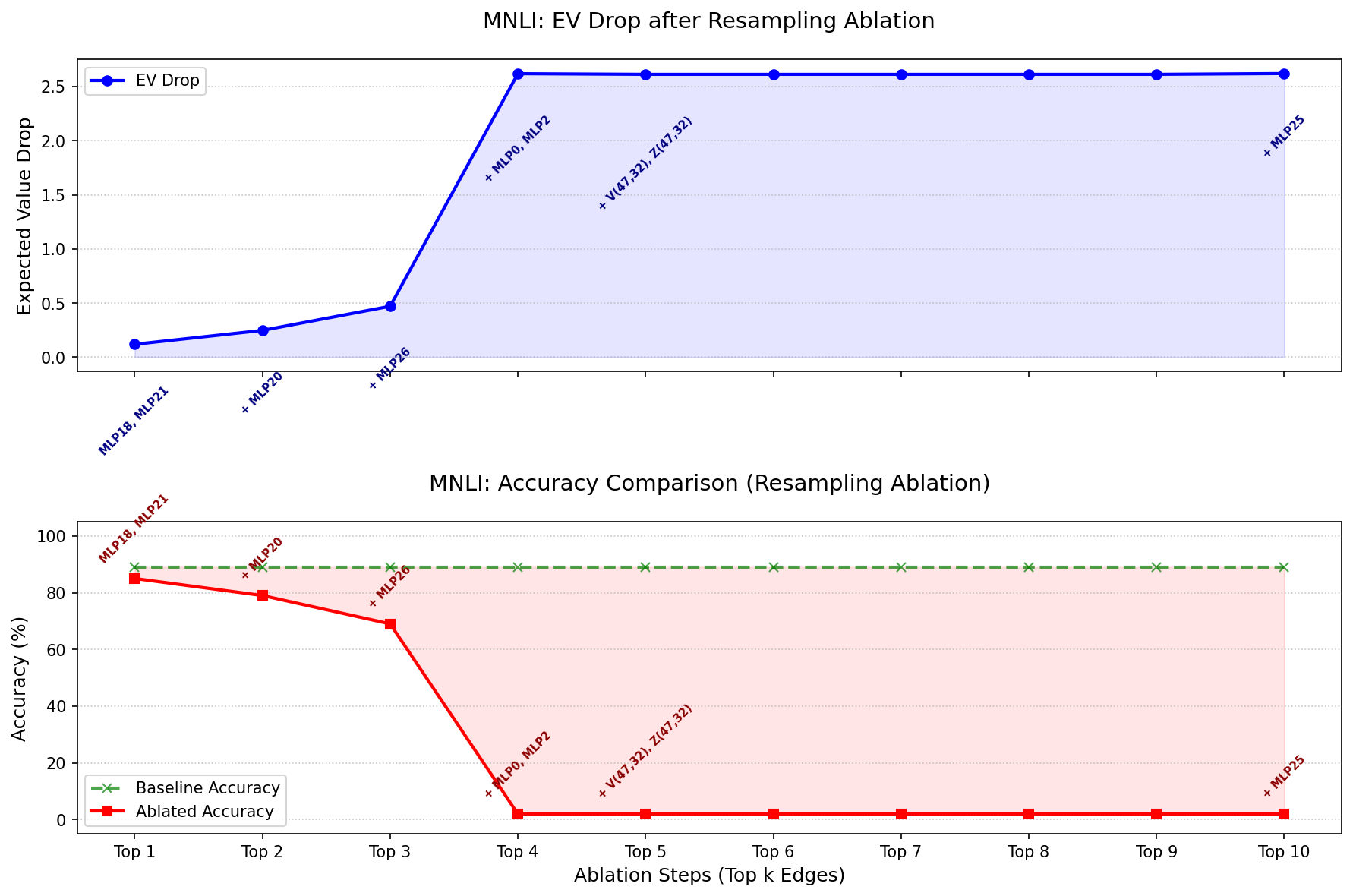}
        \caption{Numerical Judgment Ablation.}
        \label{fig:ablation_qwen25_14_mnli_num}
    \end{subfigure}
    \centering\textbf{MNLI dataset.}

    \begin{subfigure}{0.48\textwidth}
        \centering
        \includegraphics[width=\linewidth]{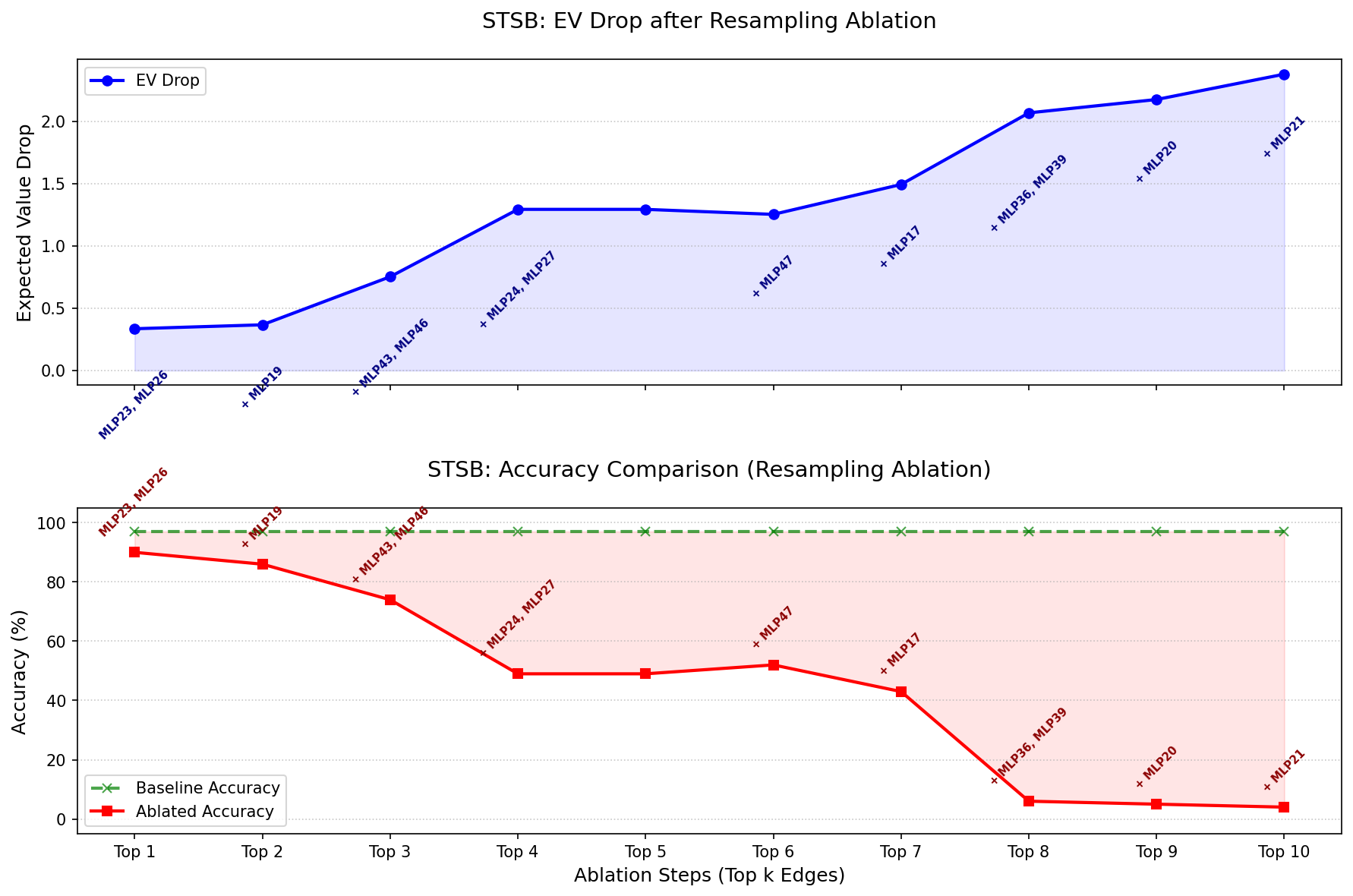}
        \caption{Classification Ablation.}
        \label{fig:ablation_qwen25_14_stsb_class}
    \end{subfigure}\hfill
    \begin{subfigure}{0.48\textwidth}
        \centering
        \includegraphics[width=\linewidth]{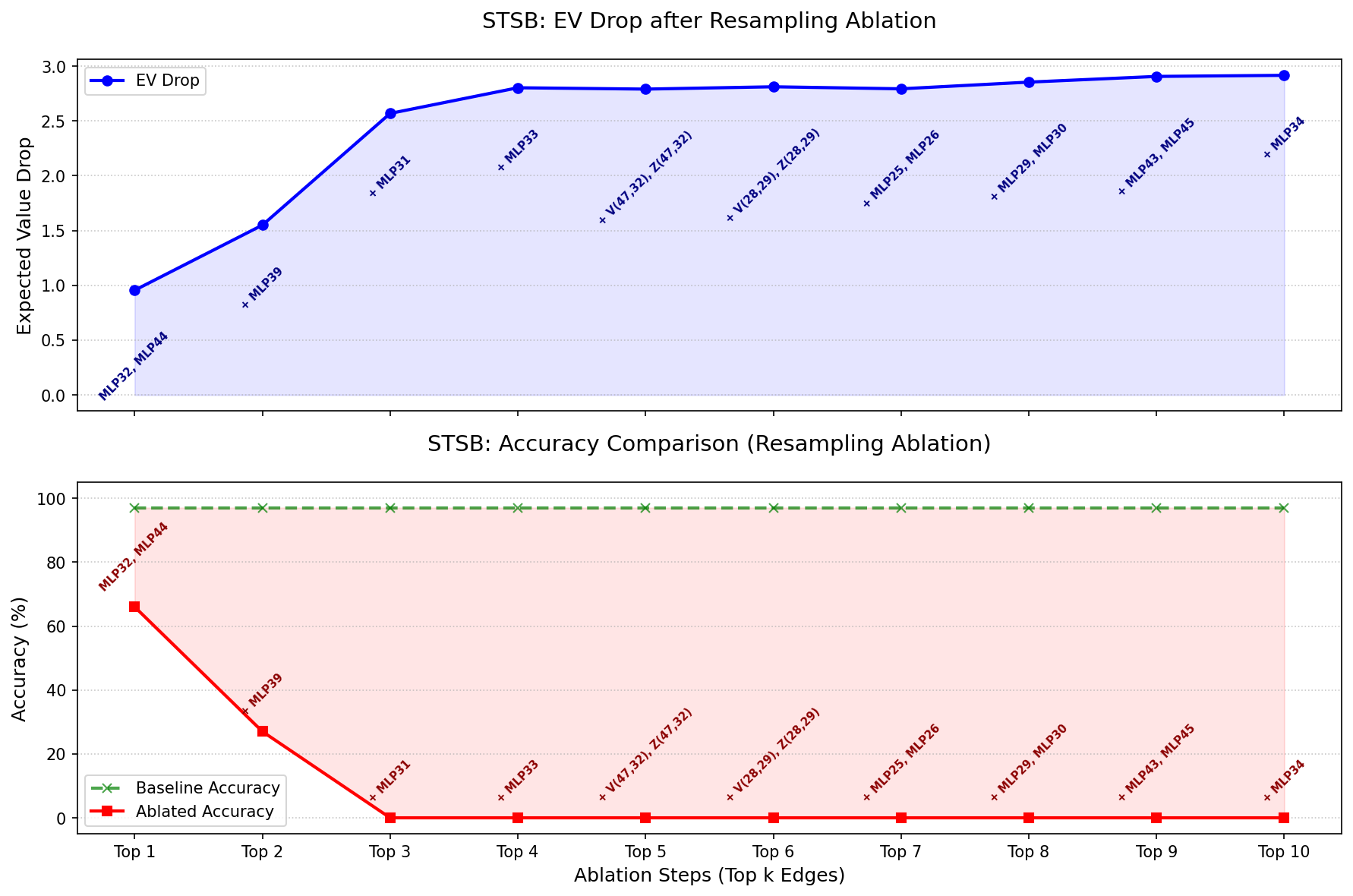}
        \caption{Numerical Judgment Ablation.}
        \label{fig:ablation_qwen25_14_stsb_num}
    \end{subfigure}
    \centering\textbf{STSB dataset.}

    \caption{Ablation phase-transition study (Qwen2.5-14B-Instruct).}
    \label{fig:ablation_study_qwen25_14}
    \vspace*{-1em}
\end{figure*}

Figures~\ref{fig:ablation_study_gemma3_12}, \ref{fig:ablation_study_gemma3_27}, \ref{fig:ablation_study_qwen25_7}, and \ref{fig:ablation_study_qwen25_14} illustrate the ablation study results showing the effect on downstream task performance on Gemma-3-12B, Gemma-3-27B, Qwen2.5-7B, and Qwen2.5-14B, respectively.

\begin{figure*}[t]
    \centering
    \begin{tabular}{ccc}
    \multicolumn{3}{l}{\textbf{CoLA}} \\[0.3em]
    \includegraphics[width=0.31\textwidth]{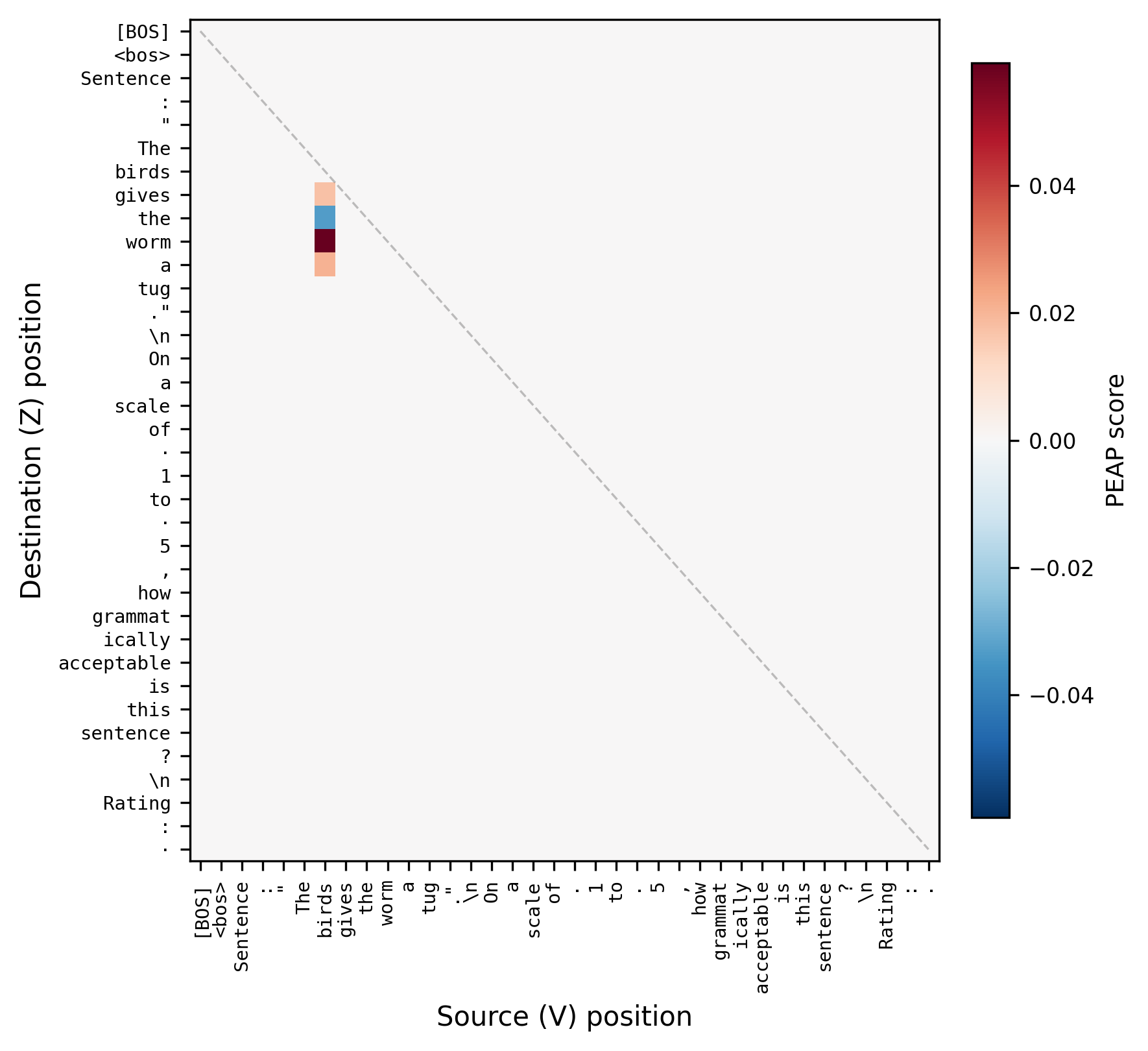} &
    \includegraphics[width=0.31\textwidth]{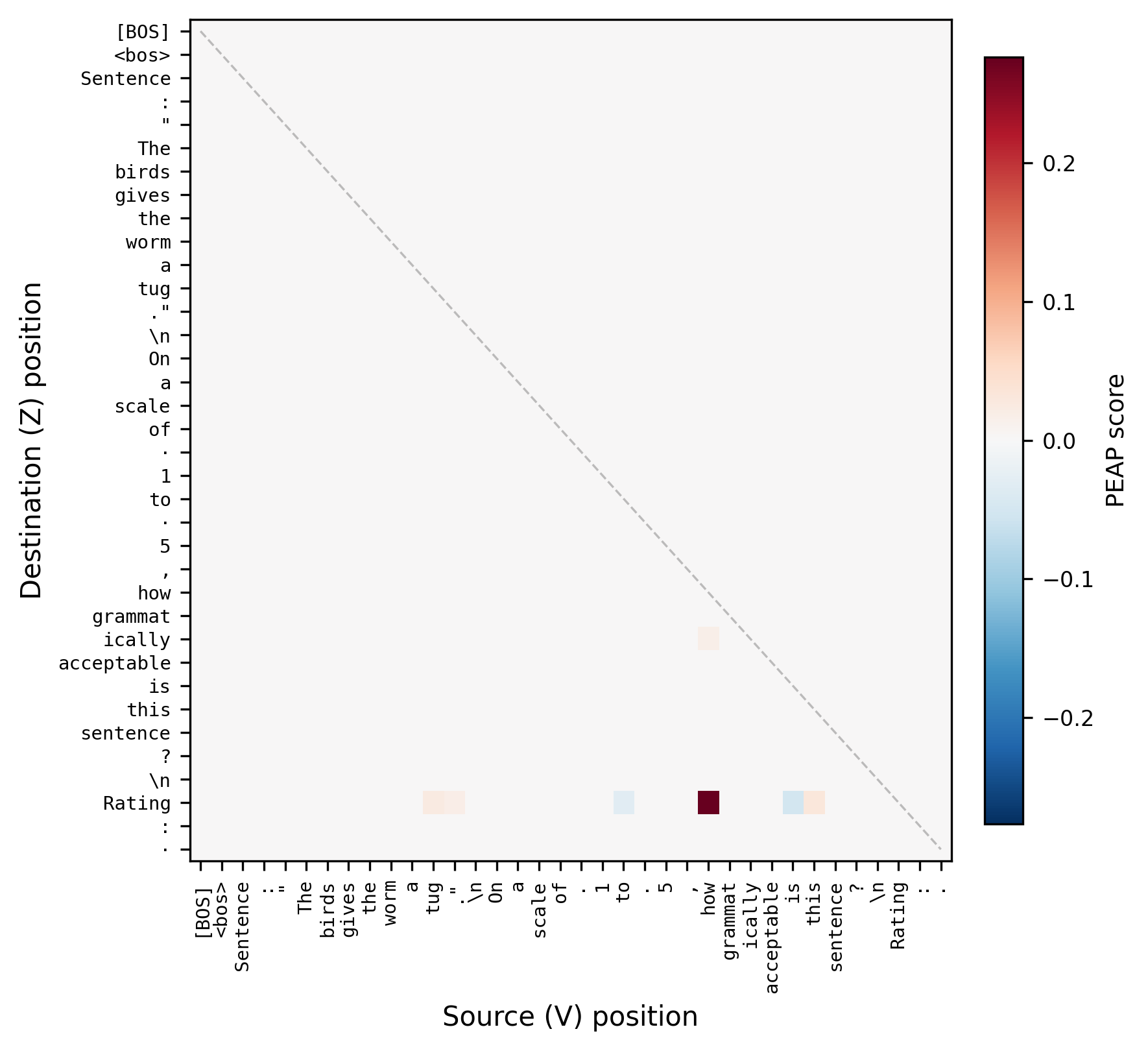} &
    \includegraphics[width=0.31\textwidth]{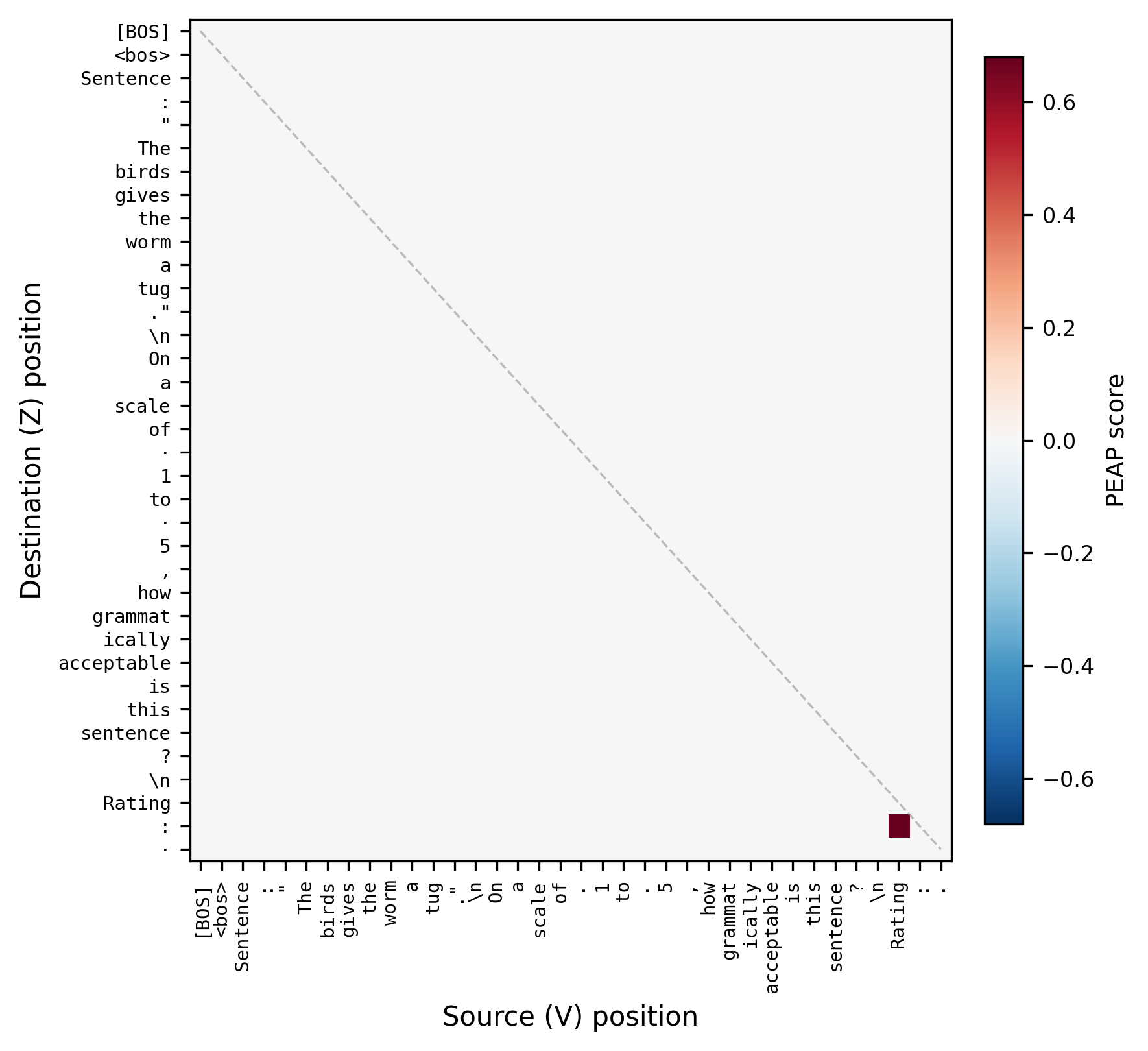} \\
    {\small L6H15 (rating formatter)} &
    {\small L25H12 (rating formatter)} &
    {\small L47H7 (shared evaluator)} \\[1em]
    \multicolumn{3}{l}{\textbf{STS-B}} \\[0.3em]
    \includegraphics[width=0.31\textwidth]{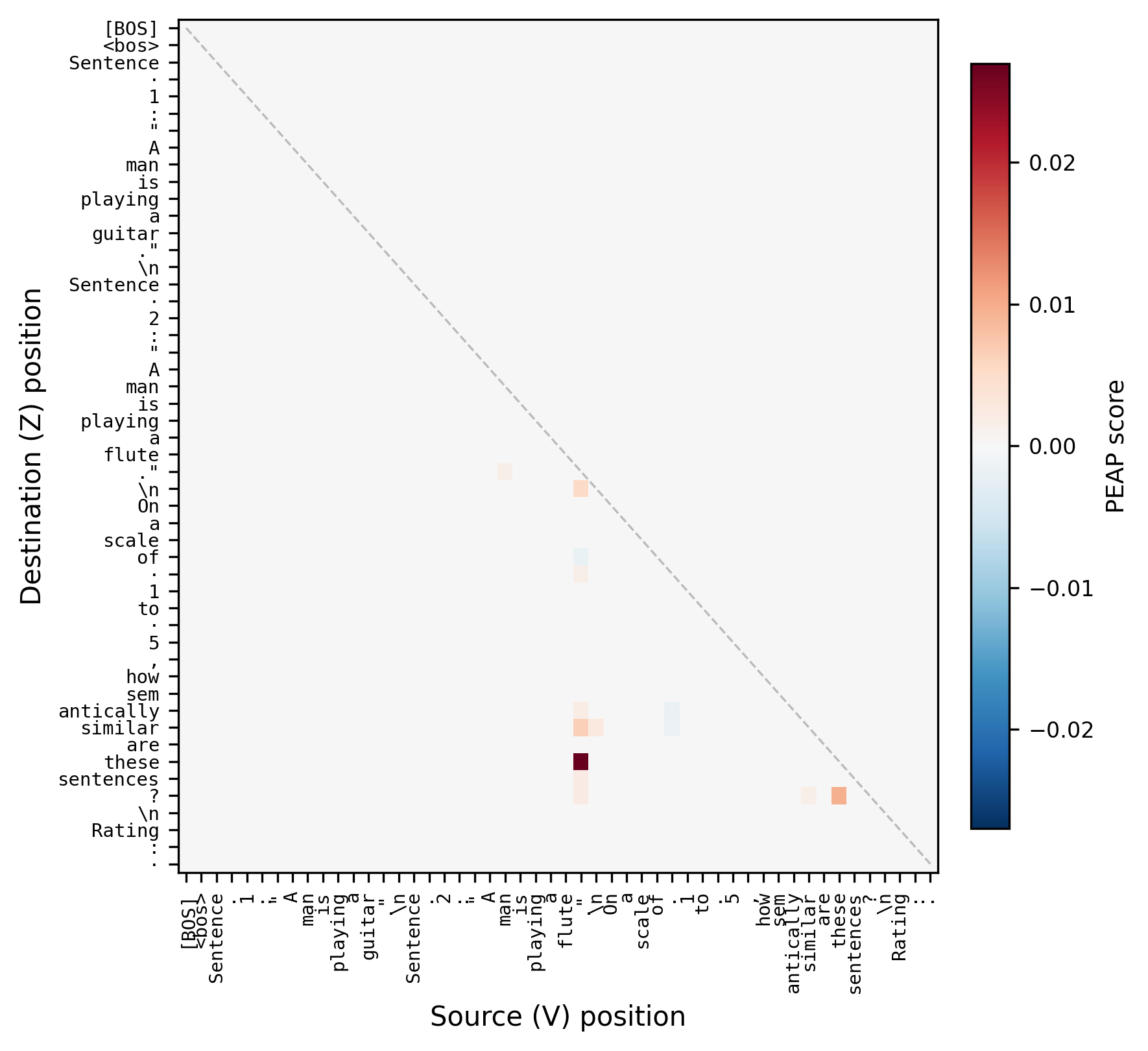} &
    \includegraphics[width=0.31\textwidth]{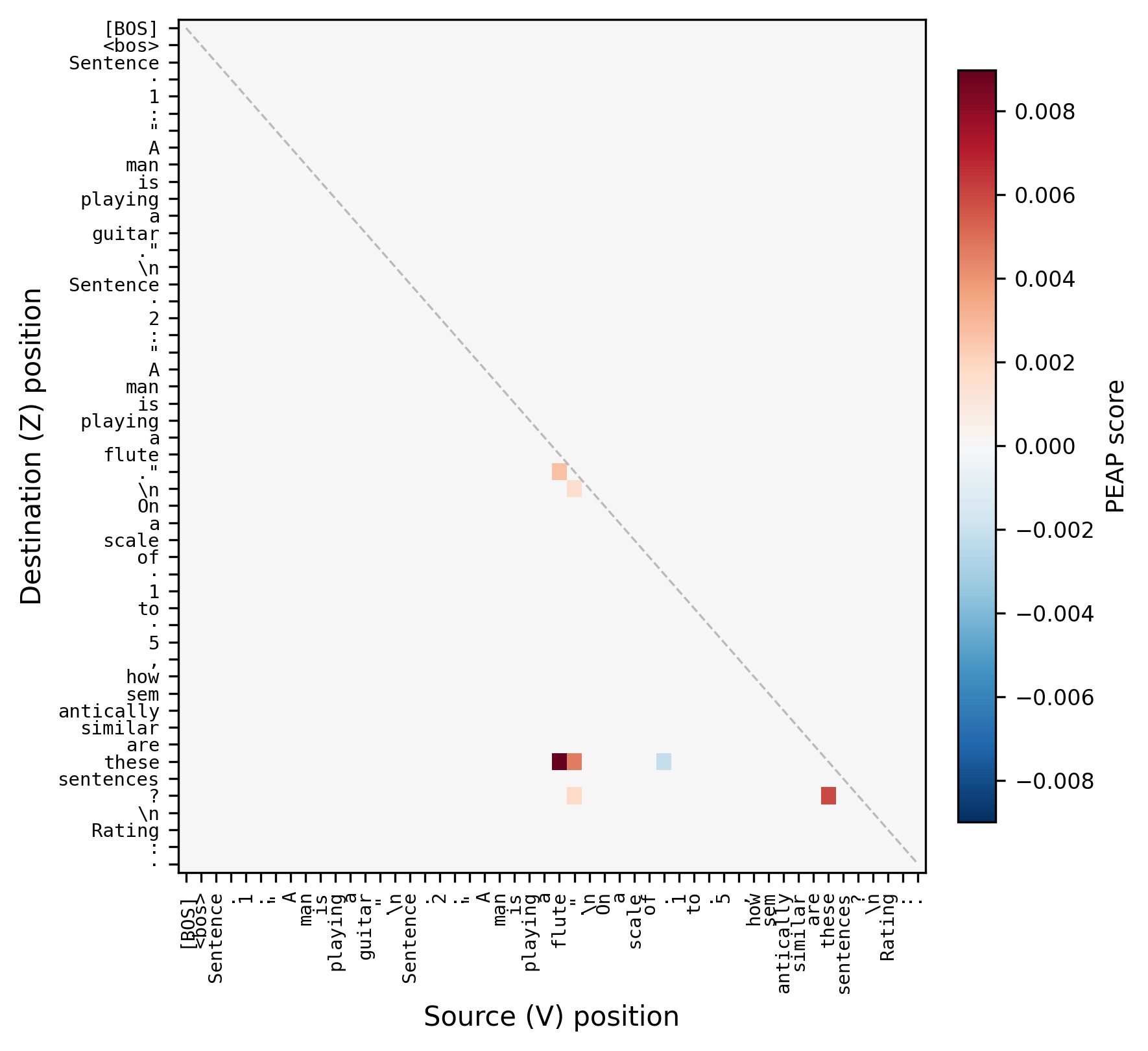} &
    \includegraphics[width=0.31\textwidth]{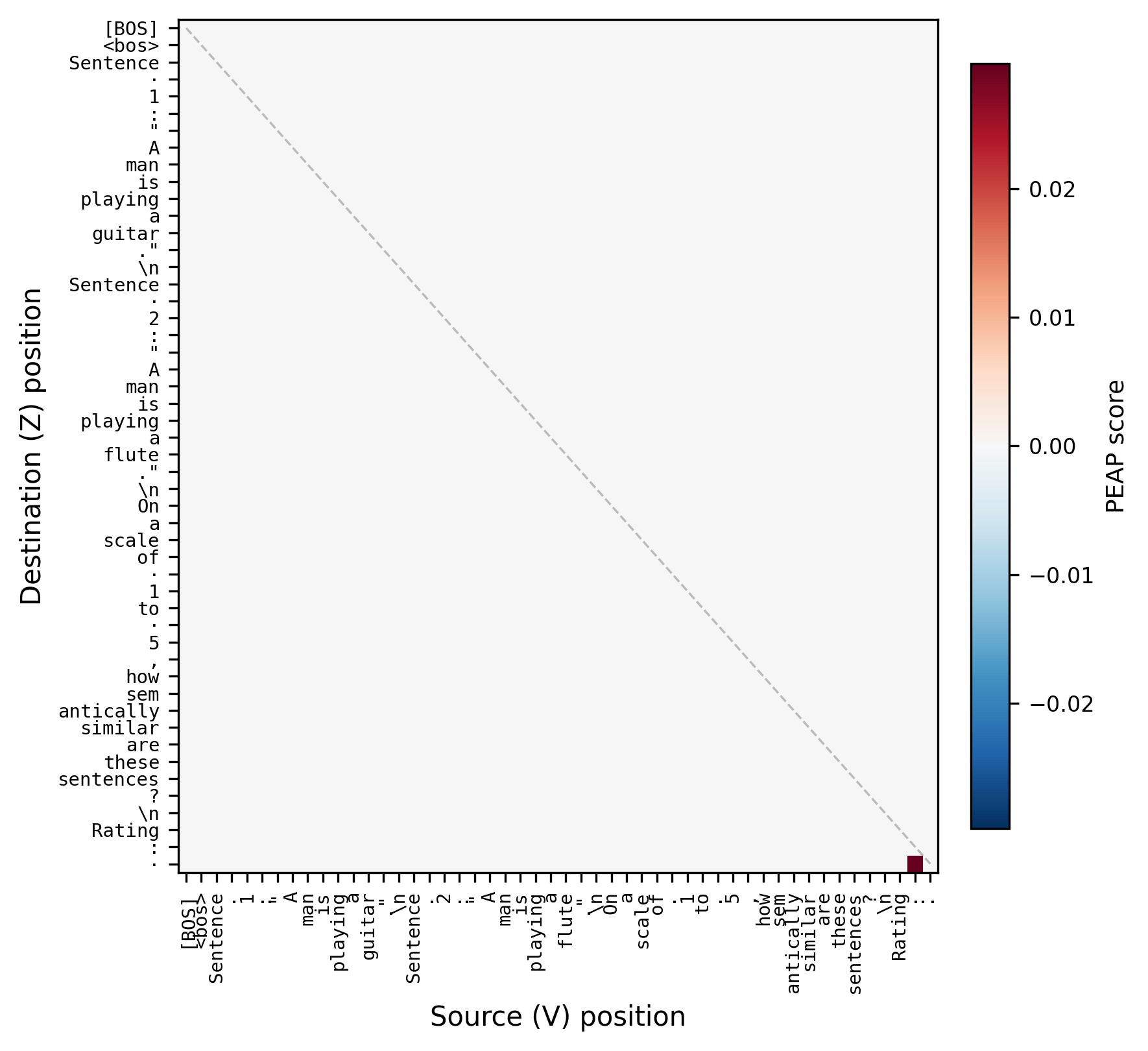} \\
    {\small L20H9 (shared evaluator)} &
    {\small L21H15 (class/rating formatter)} &
    {\small L47H7 (shared evaluator)} \\
    \end{tabular}
    \caption{
        Representative PEAP attribution heatmaps for selected attention heads in the \data{CoLA} (top) and \data{STS-B} (bottom) circuits. 
        Each cell encodes the causal attribution score from a source token position (x-axis) to a destination token position (y-axis), with red indicating positive and blue negative attribution. 
        Heads are ordered by layer depth from left to right. L47H7 appears in both circuits on matched pairs, demonstrating its consistent format-agnostic terminal routing behavior across tasks.
    }
    \label{fig:heatmaps}
\end{figure*}

\section{Full Format Transfer Injection Results}
\label{app:fti_full}

Table~\ref{tab:fti_full} reports the complete FTI panel behind the flip rates in Table~\ref{tab:fti_results}: per-cell base and patched probability of the positive target token, the flip rate, and its Wilson 95\% confidence interval.

\begin{table}[ht]
    \centering
    \caption{
        FTI probability shifts on all five tasks with Wilson 95\% CIs on the flip rate. Patching a $5$-star Latent Evaluator into a corrupted categorical classification prompt shifts probability mass toward the positive target token (Yes/Entailment) when the Task Formatter is geometrically compatible. $N$~is the post-filter pair count under the inclusion criteria (source rating EV $> 4$, corrupted base prediction $\notin\{$Yes, Entailment$\}$). Cells marked $^{\ast}$ ($N \leq 26$) are underpowered. Per-cell discussion in \S\ref{sec:fti_results}.
    }
    \label{tab:fti_full}
    \resizebox{\linewidth}{!}{%
    \begin{tabular}{clcccc}
        \toprule
        \textbf{Model} & \textbf{Dataset} & \textbf{Base Prob.} & \textbf{Patched Prob.} & \textbf{Flip Rate} & \textbf{Wls. 95\% CI} \\
        \midrule
        \multirow{5}{*}{Qwen2.5-7B}
        & \data{CoLA} ($N=26$)$^{\ast}$ & 17.4\% $\pm$ 14.9\% & 93.5\% $\pm$ 3.9\% & 100.0\% & [87.1, 100] \\
        & \data{MNLI} ($N=324$) & 3.4\% $\pm$ 6.3\% & 88.0\% $\pm$ 10.5\% & 99.1\% & [97.3, 99.7] \\
        & \data{STS-B} ($N=134$) & 1.6\% $\pm$ 2.2\% & 85.6\% $\pm$ 9.6\% & 100.0\% & [97.2, 100] \\
        & \data{Yelp} ($N=437$) & 0.1\% $\pm$ \phantom{0}0.4\% & 59.5\% $\pm$ 38.5\% & \phantom{0}66.4\% & [61.8, 70.6] \\
        & \data{RewardBench} ($N=84$) & 1.6\% $\pm$ \phantom{0}5.1\% & 96.9\% $\pm$ 10.1\% & \textbf{100.0\%} & [95.6, 100] \\
        \midrule
        \multirow{4}{*}{Llama-3.1-8B}
        & \data{CoLA} ($N=8$)$^{\ast}$ & 28.6\% $\pm$ \phantom{0}7.6\% & 41.7\% $\pm$ 13.9\% & 25.0\% & [7.1, 59.1] \\
        & \data{MNLI} ($N=120$) & 10.4\% $\pm$ 12.6\% & 13.1\% $\pm$ 17.0\% & \phantom{0}4.2\% & [1.8, 9.4] \\
        & \data{Yelp} ($N=94$) & \phantom{0}7.1\% $\pm$ \phantom{0}7.0\% & 11.1\% $\pm$ \phantom{0}5.4\% & \phantom{0}0.0\% & [0.0, 3.9] \\
        & \data{RewardBench} ($N=9$)$^{\ast}$ & 26.1\% $\pm$ 18.3\% & 39.6\% $\pm$ 20.8\% & 22.2\% & [6.3, 54.7] \\
        \midrule
        \multirow{5}{*}{Gemma-3-12B}
        & \data{CoLA} ($N=47$) & 6.1\% $\pm$ 11.8\% & 17.6\% $\pm$ 26.8\% & 14.9\% & [7.4, 27.7] \\
        & \data{MNLI} ($N=402$) & 6.3\% $\pm$ 11.0\% & 11.8\% $\pm$ 15.9\% & \phantom{0}3.7\% & [2.3, 6.1] \\
        & \data{STS-B} ($N=44$) & 15.1\% $\pm$ 13.6\% & 53.5\% $\pm$ 33.1\% & 50.0\% & [35.8, 64.2] \\
        & \data{Yelp} ($N=92$) & 0.0\% $\pm$ \phantom{0}0.1\% & \phantom{0}0.0\% $\pm$ \phantom{0}0.1\% & \phantom{0}0.0\% & [0.0, 4.0] \\
        & \data{RewardBench} ($N=26$)$^{\ast}$ & 7.2\% $\pm$ 12.4\% & 10.7\% $\pm$ 17.3\% & \phantom{0}7.7\% & [2.1, 24.1] \\
        \midrule
        \multirow{5}{*}{Qwen2.5-14B}
        & \data{CoLA} ($N=12$)$^{\ast}$ & 24.3\% $\pm$ 14.1\% & 69.0\% $\pm$ 18.7\% & 91.7\% & [64.6, 98.5] \\
        & \data{MNLI} ($N=125$) & 4.2\% $\pm$ \phantom{0}9.4\% & \phantom{0}4.9\% $\pm$ 14.2\% & \phantom{0}4.0\% & [1.7, 9.0] \\
        & \data{STS-B} ($N=156$) & 0.0\% $\pm$ \phantom{0}0.5\% & 16.0\% $\pm$ 16.0\% & \phantom{0}5.1\% & [2.6, 9.8] \\
        & \data{Yelp} ($N=189$) & 0.4\% $\pm$ \phantom{0}0.8\% & 41.0\% $\pm$ 16.9\% & 31.2\% & [25.0, 38.1] \\
        & \data{RewardBench} ($N=51$) & 1.5\% $\pm$ \phantom{0}5.0\% & 43.4\% $\pm$ 36.4\% & 41.2\% & [28.8, 54.8] \\
        \midrule
        \multirow{5}{*}{Gemma-3-27B}
        & \data{CoLA} ($N=54$) & 8.6\% $\pm$ 10.6\% & \phantom{0}9.9\% $\pm$ 11.2\% & \phantom{0}0.0\% & [0.0, 6.6] \\
        & \data{MNLI} ($N=151$) & 8.0\% $\pm$ 11.9\% & \phantom{0}8.0\% $\pm$ 11.9\% & \phantom{0}0.7\% & [0.1, 3.7] \\
        & \data{STS-B} ($N=31$) & 30.5\% $\pm$ \phantom{0}7.3\% & 78.1\% $\pm$ 13.7\% & \textbf{96.8\%} & [83.8, 99.4] \\
        & \data{Yelp} ($N=189$) & 1.3\% $\pm$ \phantom{0}3.2\% & \phantom{0}3.2\% $\pm$ \phantom{0}5.9\% & \phantom{0}1.1\% & [0.3, 3.8] \\
        & \data{RewardBench} ($N=19$)$^{\ast}$ & 20.8\% $\pm$ 18.1\% & 22.1\% $\pm$ 20.7\% & \phantom{0}5.3\% & [0.9, 24.6] \\
        \bottomrule
    \end{tabular}
    }
\end{table}

\section{Reverse Format Transfer Injection}
\label{app:reverse_fti}

Reverse FTI runs the transfer of \S\ref{sec:inconsistency} in the opposite direction: we capture the Latent Evaluator activations of a confidently positive \textit{classification} prompt (source probability of the positive label $\geq 0.8$) and inject them into the same model running the matched \textit{rating} prompt, restricted to instances whose baseline rating argmax is below $4$.
If the rating formatter accepted a blanket categorical judgment the way exposed classification formatters accept a rating judgment (Table~\ref{tab:fti_full}), the injection should lift the predicted rating to $4$ or $5$.
The attractor account of \S\ref{sec:fti_results} predicts the opposite: with five competing rating tokens, the injected mass should spread without settling on a single one.

\begin{table}[ht]
    \centering
    \caption{
        Reverse FTI: injecting a confidently positive classification Latent Evaluator into a rating prompt whose baseline argmax rating is below $4$.
        Flip rate is the fraction of pairs whose patched argmax rating reaches $4$ or $5$; $\Delta$EV${>}0$ is the fraction of pairs whose expected rating value moves upward.
        $^{\ast}$~marks underpowered cells ($N \leq 21$).
    }
    \label{tab:reverse_fti}
    \resizebox{\linewidth}{!}{%
    \begin{tabular}{llcccc}
        \toprule
        \textbf{Model} & \textbf{Dataset} & \textbf{Base $\rightarrow$ Patched EV} & \textbf{$\Delta$EV $\pm$ SD} & \textbf{Flip Rate} & \textbf{$\Delta$EV${>}0$} \\
        \midrule
        \multirow{5}{*}{Qwen2.5-7B}
        & \data{CoLA} ($N=114$)        & 2.29 $\rightarrow$ 1.65 & $-0.64 \pm 0.34$ & 0.9\%  & 1.8\% \\
        & \data{MNLI} ($N=21$)$^{\ast}$ & 1.07 $\rightarrow$ 1.05 & $-0.02 \pm 0.16$ & 0.0\%  & 66.7\% \\
        & \data{STS-B} ($N=174$)       & 1.59 $\rightarrow$ 1.78 & $+0.19 \pm 0.30$ & 1.7\%  & 69.5\% \\
        & \data{Yelp} ($N=483$)        & 1.06 $\rightarrow$ 1.08 & $+0.01 \pm 0.08$ & 0.0\%  & 61.7\% \\
        & \data{RewardBench} ($N=250$) & 1.57 $\rightarrow$ 1.84 & $+0.26 \pm 0.48$ & 6.4\%  & 79.2\% \\
        \midrule
        \multirow{5}{*}{Llama-3.1-8B}
        & \data{CoLA} ($N=50$)         & 2.35 $\rightarrow$ 2.47 & $+0.12 \pm 0.13$ & 2.0\%  & 86.0\% \\
        & \data{MNLI} ($N=7$)$^{\ast}$  & 2.31 $\rightarrow$ 2.54 & $+0.23 \pm 0.36$ & 0.0\%  & 85.7\% \\
        & \data{STS-B} ($N=177$)       & 2.19 $\rightarrow$ 2.20 & $+0.01 \pm 0.51$ & 2.8\%  & 41.8\% \\
        & \data{Yelp} ($N=96$)         & 1.17 $\rightarrow$ 1.24 & $+0.08 \pm 0.09$ & 0.0\%  & 91.7\% \\
        & \data{RewardBench} ($N=43$)  & 2.26 $\rightarrow$ 3.33 & $+1.07 \pm 0.50$ & \textbf{58.1\%} & 100.0\% \\
        \midrule
        \multirow{5}{*}{Gemma-3-12B}
        & \data{CoLA} ($N=121$)        & 1.67 $\rightarrow$ 1.77 & $+0.11 \pm 0.14$ & 0.0\%  & 100.0\% \\
        & \data{MNLI} ($N=65$)         & 1.51 $\rightarrow$ 1.43 & $-0.08 \pm 0.16$ & 3.1\%  & 9.2\% \\
        & \data{STS-B} ($N=177$)       & 2.42 $\rightarrow$ 2.92 & $+0.50 \pm 0.37$ & \textbf{37.9\%} & 96.6\% \\
        & \data{Yelp} ($N=95$)         & 1.01 $\rightarrow$ 1.01 & $-0.00 \pm 0.04$ & 0.0\%  & 92.6\% \\
        & \data{RewardBench} ($N=64$)  & 1.51 $\rightarrow$ 1.80 & $+0.29 \pm 0.63$ & 4.7\%  & 96.9\% \\
        \midrule
        \multirow{5}{*}{Qwen2.5-14B}
        & \data{CoLA} ($N=113$)        & 1.91 $\rightarrow$ 1.73 & $-0.18 \pm 0.17$ & 0.0\%  & 8.0\% \\
        & \data{MNLI} ($N=135$)        & 1.56 $\rightarrow$ 1.35 & $-0.21 \pm 0.24$ & 0.0\%  & 5.9\% \\
        & \data{STS-B} ($N=189$)       & 1.66 $\rightarrow$ 1.61 & $-0.06 \pm 0.13$ & 0.0\%  & 23.8\% \\
        & \data{Yelp} ($N=197$)        & 1.08 $\rightarrow$ 1.07 & $-0.01 \pm 0.06$ & 0.0\%  & 52.8\% \\
        & \data{RewardBench} ($N=103$) & 2.07 $\rightarrow$ 2.15 & $+0.08 \pm 0.26$ & 5.8\%  & 69.9\% \\
        \midrule
        \multirow{5}{*}{Gemma-3-27B}
        & \data{CoLA} ($N=125$)        & 1.62 $\rightarrow$ 1.62 & $+0.00 \pm 0.02$ & 0.0\%  & 56.0\% \\
        & \data{MNLI} ($N=66$)         & 1.38 $\rightarrow$ 1.38 & $-0.00 \pm 0.01$ & 0.0\%  & 53.0\% \\
        & \data{STS-B} ($N=89$)        & 2.08 $\rightarrow$ 2.08 & $-0.00 \pm 0.01$ & 0.0\%  & 64.0\% \\
        & \data{Yelp} ($N=197$)        & 1.11 $\rightarrow$ 1.13 & $+0.02 \pm 0.06$ & 0.0\%  & 97.0\% \\
        & \data{RewardBench}           & \multicolumn{4}{c}{not run (RewardBench classification trace unavailable)} \\
        \bottomrule
    \end{tabular}
    }
\end{table}

Table~\ref{tab:reverse_fti} matches this prediction on most cells: the expected rating value moves upward on a majority of pairs in $17$ of $24$ cells, but the argmax rating flips to $4$ or $5$ on fewer than $7\%$ of pairs everywhere except two cells.
The exceptions are Llama-3.1-8B on \data{RewardBench} and Gemma-3-12B on \data{STS-B}, where the upward shift is large enough to cross into the adjacent rating on a substantial share of pairs.
Three deviations are worth stating plainly.
First, Qwen2.5-7B on \data{CoLA} moves ratings \textit{down} ($\Delta$EV $-0.64$, with only $2\%$ of pairs moving up), so on this cell the blanket classification pattern actively conflicts with the rating computation instead of merely fragmenting.
Second, Qwen2.5-14B shows small negative shifts on four of five tasks.
Third, Gemma-3-27B is essentially inert in every cell ($|\Delta$EV$| \leq 0.02$), consistent with the insulation this model already shows under forward FTI (Table~\ref{tab:fti_full}).
The Gemma-3-27B \data{RewardBench} cell is missing because the classification-format PEAP trace for that pair has not been run.

\section{Structural Validation via Logit Lens}
\label{app:results_logit_lens}

While automated circuit discovery provides scalable methodologies for identifying active subgraph components, it heavily relies on dataset activations and external language models for semantic explanations \citep{golimblevskaia-2026-circuit-insights}.
To validate our findings, we employ Logit Lens\footnote{\url{https://www.lesswrong.com/posts/AcKRB8wDpdaN6v6ru}}: projecting the contrastive steering representations directly into the vocabulary space using the model's unembedding weights ($W_U$). 
This exposes the semantic content of the nodes without depending on black-box auto-interpretability.

To compare structure across architectures without an arbitrary threshold, we project the contrastive evaluator vectors onto the unembedding matrix and measure cosine similarity. 
Normalization removes noise from untrained tokens and reveals geometric alignment across the nodes. 
Figure~\ref{fig:semantic_bifurcation_depth} quantifies this alignment between ordinal (Rating) and categorical (Classification) evaluators across network depth.

The projections trace a shared geometry in the late-middle depth band ($0.50$ -- $0.85$).
Notably, \data{MNLI} shows weaker convergence with other tasks in this shared evaluation window, consistent with its three-way classification structure (entailment/neutral/contradiction) requiring a richer internal representation than a simple positive/negative judgment scalar. 
This suggests that the Latent Evaluator's 1D judgment abstraction generalizes best to binary or ordinal tasks, while multi-class judgment tasks partially bypass the shared trunk.
Applying a strict cross-architecture intersection to discard tokenizer-specific artifacts leaves a set of abstract evaluation tokens shared across all tasks. 
The highest-probability tokens at the evaluator nodes are the same across architectures: \{\emph{confirm}, \emph{verify}, \emph{validate}, \emph{identical}, \emph{perfectly}\}. 

Just before the terminal output-formatting boundary (depth $1.0$), however, this shared coherence collapses across networks. 
The rating tasks route their probability mass toward discrete number tokens (e.g., \emph{five}, \emph{5}, \emph{1}), abandoning the abstraction entirely.
The classification variants polarize into categorical literals (e.g., \emph{false}, \emph{true}, \emph{contradiction}) at the same depth.
This supports our Task Formatter hypothesis: the Latent Evaluator calculates a continuous judgment signal in the deep layers, and task-specific routers overwrite that geometry strictly only at the terminal formatting stage.
We emphasize that Logit Lens provides a correlational readout rather than a causal intervention: the tokens recovered through vocabulary projection represent directions that are linearly decodable from intermediate representations, which need not coincide with the representations the model actually uses for downstream computation \citep{yom-din-2024-jump-to-conclusions}.
We therefore treat these projections as supporting evidence that corroborates -- but does not independently prove -- the causal findings from PEAP and Boundless DAS.


\begin{table*}[t]
    \centering
    \caption{CoLA circuit attention head functional roles derived from per-head PEAP attribution analysis. Heads are grouped by role and sorted by layer within each group.}
    \label{tab:cola_heads}
    \small
    \begin{tabular}{lp{\dimexpr\linewidth-2cm}}
    \toprule
    \textbf{Head} & \textbf{Description} \\
    \midrule
    \multicolumn{2}{l}{\textit{Shared Evaluators}} \\
    \midrule
    L45H3 & Routes within the late instruction and output-adjacent span, occasionally aggregating from multiple nearby positions into the Rating token with mixed signs \\
    L46H12 & Consistently routes from the end of the instruction span directly to the Rating position \\
    L47H7 & Format-agnostic terminal router at the output position, forwarding signal to Rating regardless of output format; exhibits both positive and negative attribution across pairs \\
    \midrule
    \multicolumn{2}{l}{\textit{Rating Formatters}} \\
    \midrule
    L6H15 & Routes exclusively between neighboring tokens within the content span, performing local signal accumulation \\
    L16H1 & Local content routing with occasional forwarding toward the sentence boundary, likely accumulating syntactic and grammatical features \\
    L18H3 & Routes from content tokens toward the instruction region, beginning to bridge the judgment signal from content into the query template \\
    L19H0 & Routes from content tokens toward the instruction region similarly to L18H3 \\
    L20H9 & Takes signal from end of sentence and beginning of instruction, integrating it into later instruction positions \\
    L23H3 & Routes from both sentence content and mid-instruction toward the end of the instruction, progressively concentrating signal toward the output boundary \\
    L25H12 & Most integrative formatter, simultaneously aggregating signals from multiple source positions across both content and instruction spans into the output position \\
    L26H1 & Routes from sentence content and mid-instruction toward the output boundary \\
    L44H8 & Clean terminal forwarder routing from the late instruction span directly to the Rating position \\
    \midrule
    \multicolumn{2}{l}{\textit{Class Formatters}} \\
    \midrule
    L6H1 & Accumulates signal over the input content, occasionally routing toward the output; operates as early-stage format-specific preprocessor \\
    L6H14 & Operates similarly to L6H1, forming a complementary pair for early content signal accumulation \\
    L14H11 & Local content routing with forwarding toward the sentence boundary \\
    L27H9 & Pre-output processor routing within the late instruction span \\
    L42H6 & Pre-output processor routing within the late instruction span, rarely active \\
    L43H3 & Routes within the instruction span, focusing narrowly on the question mark \\
    L47H6 & Inhibitory terminal head at the output position with consistently negative attribution, dampening the rating-scale signal to allow the classification pathway to dominate \\
    \bottomrule
    \end{tabular}
\end{table*}

\section{Per-Head Causal Role Analysis}
\label{app:head_analysis}

To characterize the functional role of each attention head beyond its structural circuit membership, we visualize per-pair PEAP attribution heatmaps, where each cell encodes the causal attribution score from a sender token position to a receiver token position. 
This reveals which spans of the input each head routes information between, complementing the circuit intersection criterion with direct evidence of information flow. 
Figure~\ref{fig:heatmaps} illustrates representative attribution patterns for selected heads across early, intermediate, and terminal layers in both circuits.

\paragraph{CoLA Circuit Head Analysis.}
Table~\ref{tab:cola_heads} summarizes the functional roles of all attention heads in the \data{CoLA} circuit. 
The analysis reveals a depth-ordered processing pipeline where both rating and class formatters span the full network depth, from early content-processing heads to late terminal routers. 
The key structural difference between the two formatter types lies in the density of intermediate processing: rating formatters maintain a continuous relay of heads bridging content and instruction spans across layers 16 through 26, whereas class formatters have no equivalent intermediate stage, relying instead on weakly active late-layer heads for pre-output processing. 
The three shared evaluators (L45H3, L46H12, L47H7) are distinguished from formatters not by their layer position but by their format-agnostic routing behavior, consistently forwarding signal to the Rating position regardless of whether the output format is ordinal or categorical (see L47H7 in Figure~\ref{fig:heatmaps}, top row).

\paragraph{STS-B Circuit Head Analysis.}
To characterize attention head function in the \data{STS-B} circuit, we apply the same per-pair PEAP attribution heatmap analysis as for \data{CoLA}. Table~\ref{tab:stsb_heads} summarizes the results. 
The \data{STS-B} circuit is substantially more complex, containing 25 heads compared to 17 in \data{CoLA}, with ten shared evaluators versus three, and several heads carrying mixed roles across formatter categories. 
The earliest head activates at layer 18, ten layers deeper than \data{CoLA}'s earliest head at layer 6, reflecting the additional encoding depth required before cross-sentence comparison representations are available for circuit-level processing. 
The shared evaluators reveal a layered structure absent in \data{CoLA}: early shared evaluators (L20H9, L24H0, L25H8, L26H1) perform cross-sentence contrastive detection and instruction consolidation, visible in the distributed attribution pattern of L20H9 in Figure~\ref{fig:heatmaps} (bottom row).
Late shared evaluators (L45H3, L46H12, L47H7) replicate the same terminal forwarding behavior observed in the \data{CoLA} circuit, confirming their identity as stable cross-task output anchors. 
The inhibitory terminal heads L26H10, L27H8, and L47H6 collectively mirror the antagonistic role of L47H6 in \data{CoLA}, suggesting a consistent cross-task mechanism for format competition at the output boundary.

\section{MLP Layer Analysis via Transcoders}
\label{app:transcoder}

\begin{figure}[t]
    \centering
    \includegraphics[width=.9\columnwidth]{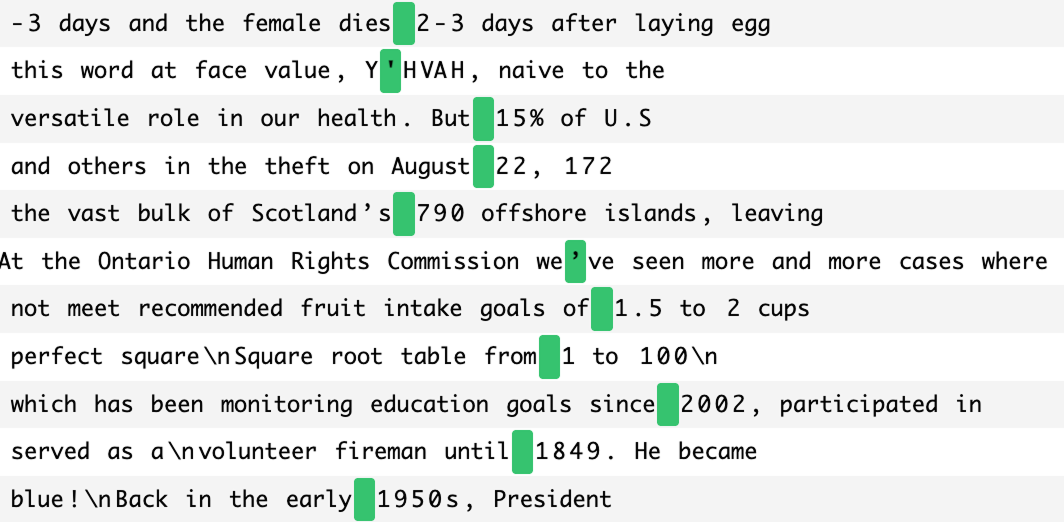}
    \caption{
        Activation pattern of transcoder feature F169 at layer 25 (position $-2$) in the \data{CoLA} circuit. 
        The feature fires consistently on the leading digit of numerical expressions across diverse corpus contexts, representing the format-level numerical output constraint active across all prompts.
    }
    \label{fig:transcoder_f169}
\end{figure}

\begin{figure}[t]
    \centering
    \includegraphics[width=.9\columnwidth]{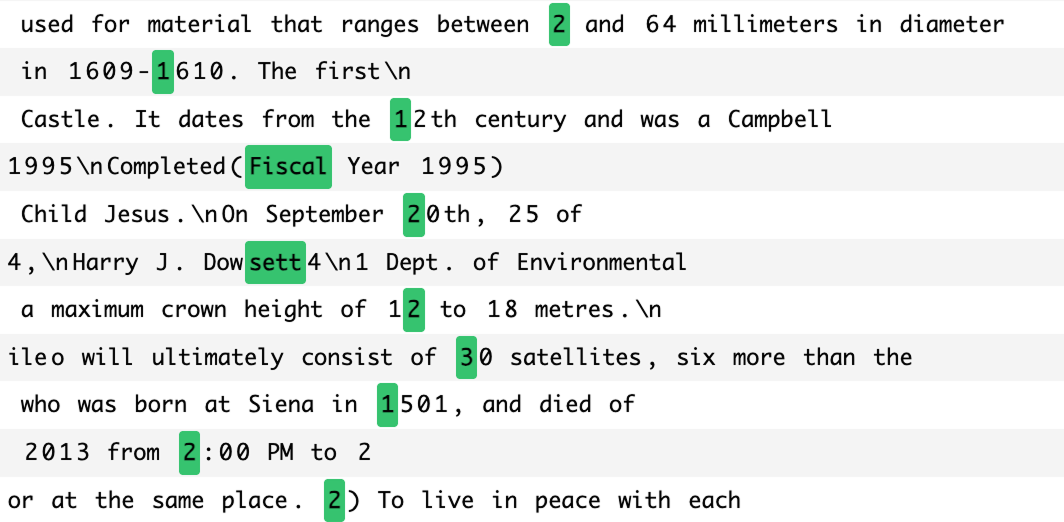}
    \caption{
        Transcoder feature F4 at layer 47 (position $-1$) in the \data{CoLA} circuit. 
        It activates on tokens 1--3 in general corpus contexts and is more active on low grammaticality prompts, illustrating the value-encoding group: 
        prompt-specific and judgment-sensitive, contrasting with format-level features that fire uniformly across all prompts.
    }
    \label{fig:transcoder_f4}
\end{figure}

To interpret the function of MLP layers identified as circuit components by PEAP, we analyze transcoder features~\citep{dunefsky-2024-transcoders} from a Replacement Model trained on Gemma-3-12B. 
Specifically, we use the 16k-feature transcoders from the Gemma-Scope-2 release~\citep{mcdougall2025gemmascope2}, noting that this feature dimensionality constrains the granularity of recoverable features; finer-grained semantic distinctions may require higher-capacity variants. 
Rather than running the transcoder-based circuit tracer, which would constitute a separate and methodologically distinct analysis requiring significant additional GPU compute, we instead directly analyze the most activating transcoder features at the token positions and layers at which PEAP identifies MLP nodes as circuit components. 
Feature activations are sampled over a 100M token subsample of the Pile~\citep{gao-2020-the-pile}, following the inverse-frequency approach of~\citet{golimblevskaia-2026-circuit-insights}. 
We do not apply automated interpretability~\citep{bills2023language, choi2024scaling, golimblevskaia-2026-circuit-insights} but instead manually inspect selected features, as no open-source feature descriptions or activation samples are publicly available for transcoders or Sparse Autoencoders (SAEs) trained on Gemma-3-12B. 
For each circuit-relevant MLP layer and token position, we identify features that activate consistently across all or most prompts and examine their activation patterns over the dataset to interpret their function.

\paragraph{CoLA MLP Analysis.}
Analysis of transcoder features across layers 25 through 47 at the pre-output position reveals two functional groups. 
The majority of top activated features fire consistently across all prompts and encode a general numerical output constraint, signaling that the next token will be a number regardless of the specific rating value.
This pattern persists through layer 40. 
Figure~\ref{fig:transcoder_f169} shows a representative example of this group: F169 at layer 25 activates on the leading digit of numerical expressions across diverse contexts, consistently firing at position $-2$ across all analyzed prompts. 
A second group of prompt-specific features encodes the actual rating value.
F1201 in layer 25 and F255 in layer 26 are representative examples: they tend to co-occur with low rating outputs but are polysemantic in general corpus contexts, activating on numbers in dates, counts, and list items. 
The persistence of the numerical constraint cluster across layers 25 through 40 suggests that the MLP chain is primarily engaged in progressively committing to and refining the numerical output. 
Layers 46 and 47 show a qualitative shift: activity moves to position $-1$, one step closer to final token selection, shared numerical features become more polysemantic, and judgment-sensitive features emerge. 
F4 in layer 47 (Figure~\ref{fig:transcoder_f4}) appears as a top feature for ungrammatical inputs but is absent from the top features for grammatical ones, suggesting it carries signal specifically for low rating outputs rather than just numerical format.

\paragraph{STS-B MLP Analysis.}
The STS-B MLP chain at the rating position progresses through three distinct functional regimes. 
Layers 15 through 21 show diffuse, largely uninterpretable feature activations spanning diverse token types across all input positions, consistent with general context aggregation rather than judgment-specific computation. 
Layers 22 through 24 mark a transitional regime where activity concentrates at the ``?'' token (position $-5$), the boundary between the instruction question and the output template, with features beginning to show numerical specificity.
This staging position is independently corroborated by the attention head analysis, where multiple heads were observed routing signal from ``?'' toward the Rating and ``:'' positions. 
From layer 25 onward the pattern stabilizes at position $-2$ with a cluster of features firing consistently across all prompts encoding a general numerical output constraint, alongside prompt-specific value-encoding features such as F255 in layer 26, which is biased toward low rating outputs but polysemantic in general corpus contexts. 
Layers 45 and 46 shift to position $-1$, with layer 47 replicating the terminal pattern seen in \data{CoLA}, including judgment-sensitive features that differentiate between high and low similarity pairs.

The difficulty of interpreting features in intermediate layers is consistent with broader findings in automated interpretability research~\citep{bills2023language, choi2024scaling, puri-etal-2025-fade, golimblevskaia-2026-circuit-insights}, where mid-network representations tend to encode distributed, context-dependent computations that resist clean decomposition into human-interpretable concepts.

Both circuits converge on the same two-level output preparation mechanism in their terminal MLP layers: a shared cluster of numerical format features firing across all prompts at position $-2$, and a smaller set of prompt-specific polysemantic features encoding the actual rating value. The key difference is the path to this regime. 
\data{CoLA} reaches it directly at layer 25 with no meaningful transitional stage, while \data{STS-B} requires an additional ten layers of diffuse context aggregation followed by a transitional consolidation at the ``?'' token before the same numerical output pattern emerges at layer 25. 
This difference in onset depth reflects the structural complexity gap between the two tasks: single-sentence grammaticality assessment can commit to a numerical output regime earlier, while cross-sentence semantic comparison requires more intermediate processing before a stable judgment signal is available for output formatting.

\section{Global Judge Circuit Topology}
\label{app:peap_circuit_topology}

\begin{figure*}[t]
    \centering
    \includegraphics[width=\textwidth]{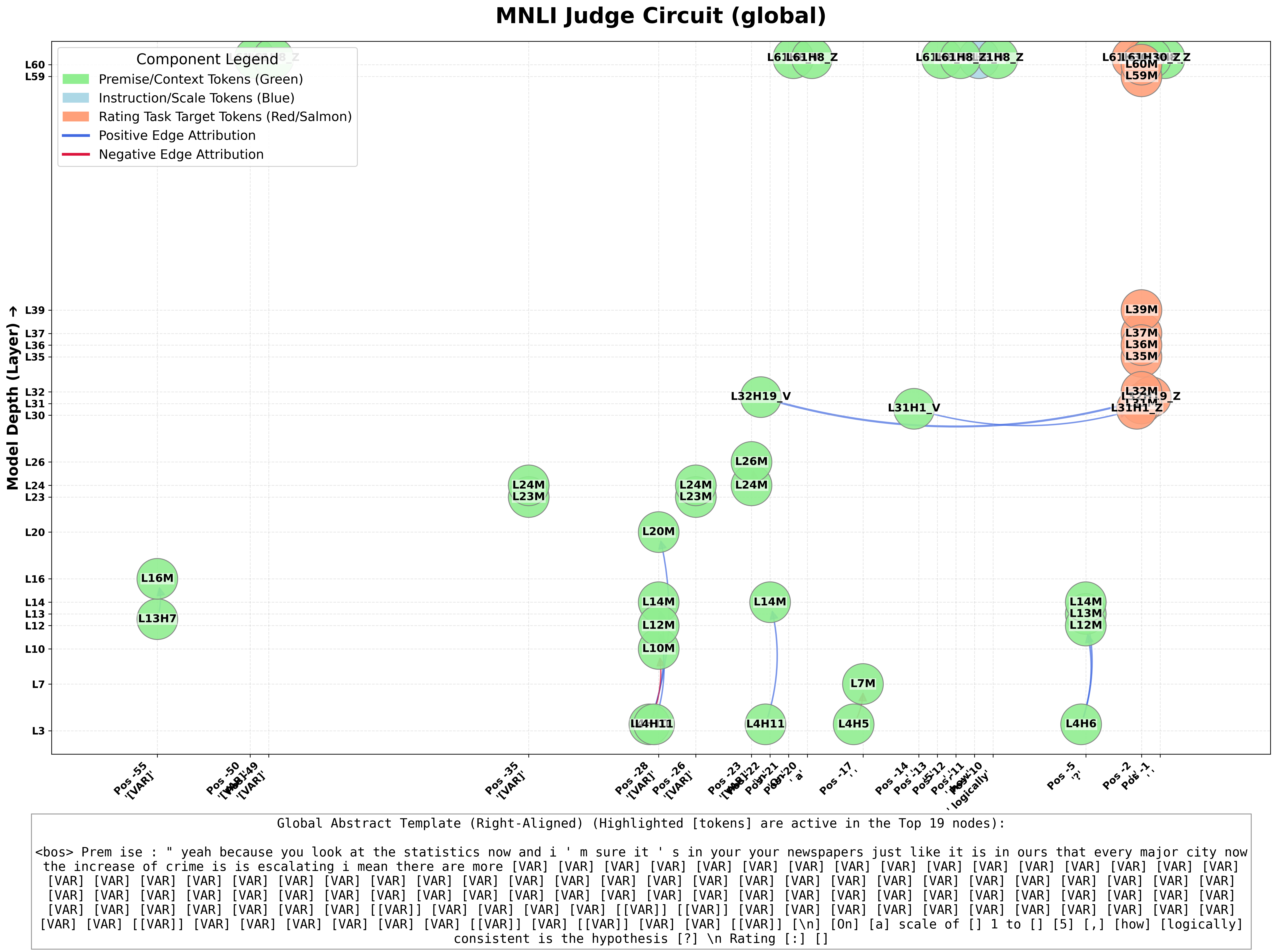}
    \caption{Global judge circuit for \data{MNLI} on Gemma-3-27B.}
    \label{fig:peap_circuit}
\end{figure*}

\begin{figure*}[t]
    \centering
    \includegraphics[width=\textwidth]{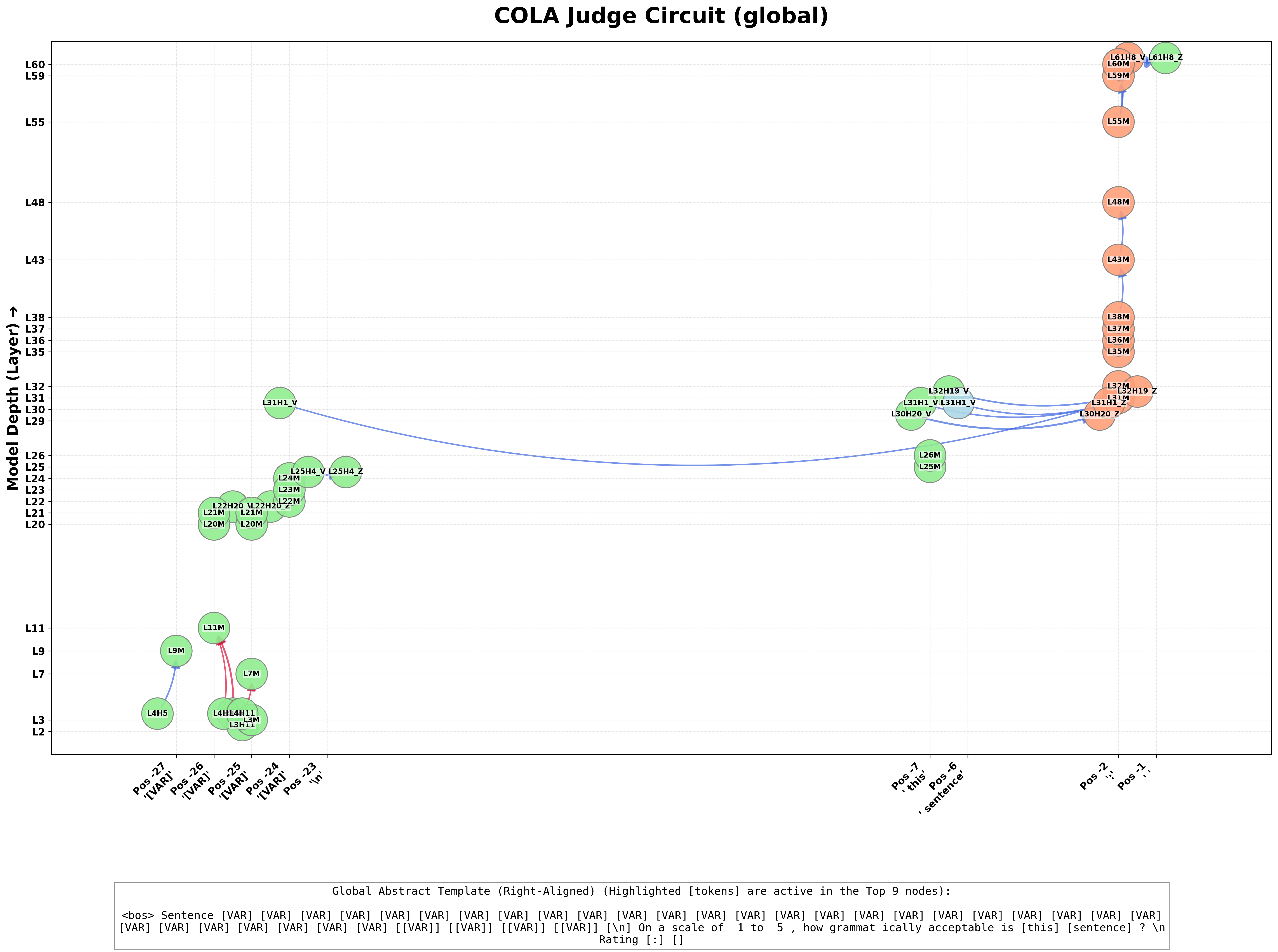}
    \caption{Global judge circuit for \data{CoLA} on Gemma-3-27B.}
    \label{fig:peap_circuit_cola_g27b}
\end{figure*}

\begin{figure*}[t]
    \centering
    \includegraphics[width=\textwidth]{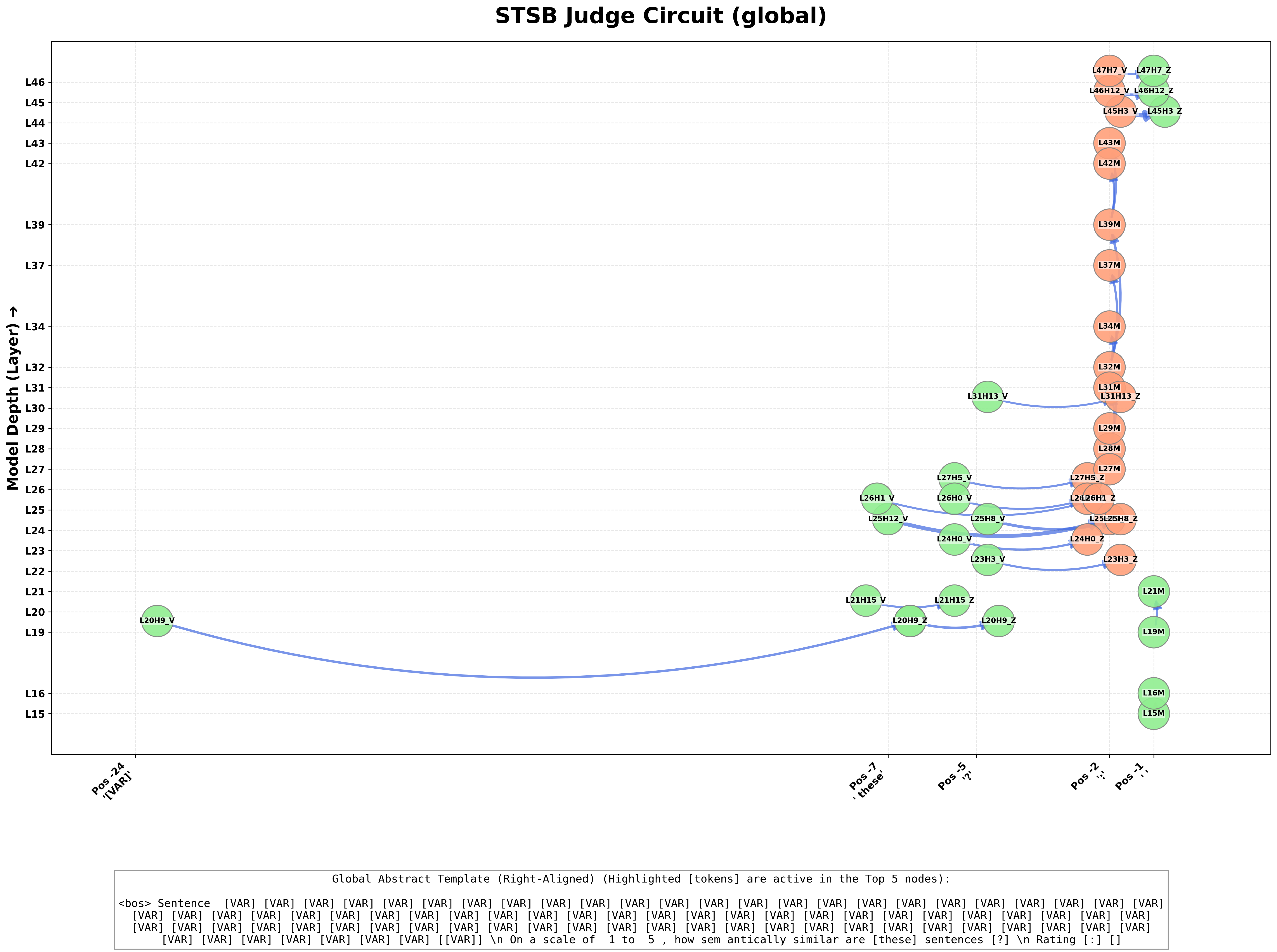}
    \caption{Global judge circuit for \data{STS-B} on Gemma-3-12B.}
    \label{fig:peap_circuit_stsb_g12b}
\end{figure*}

\begin{figure*}[t]
    \centering
    \includegraphics[width=\textwidth]{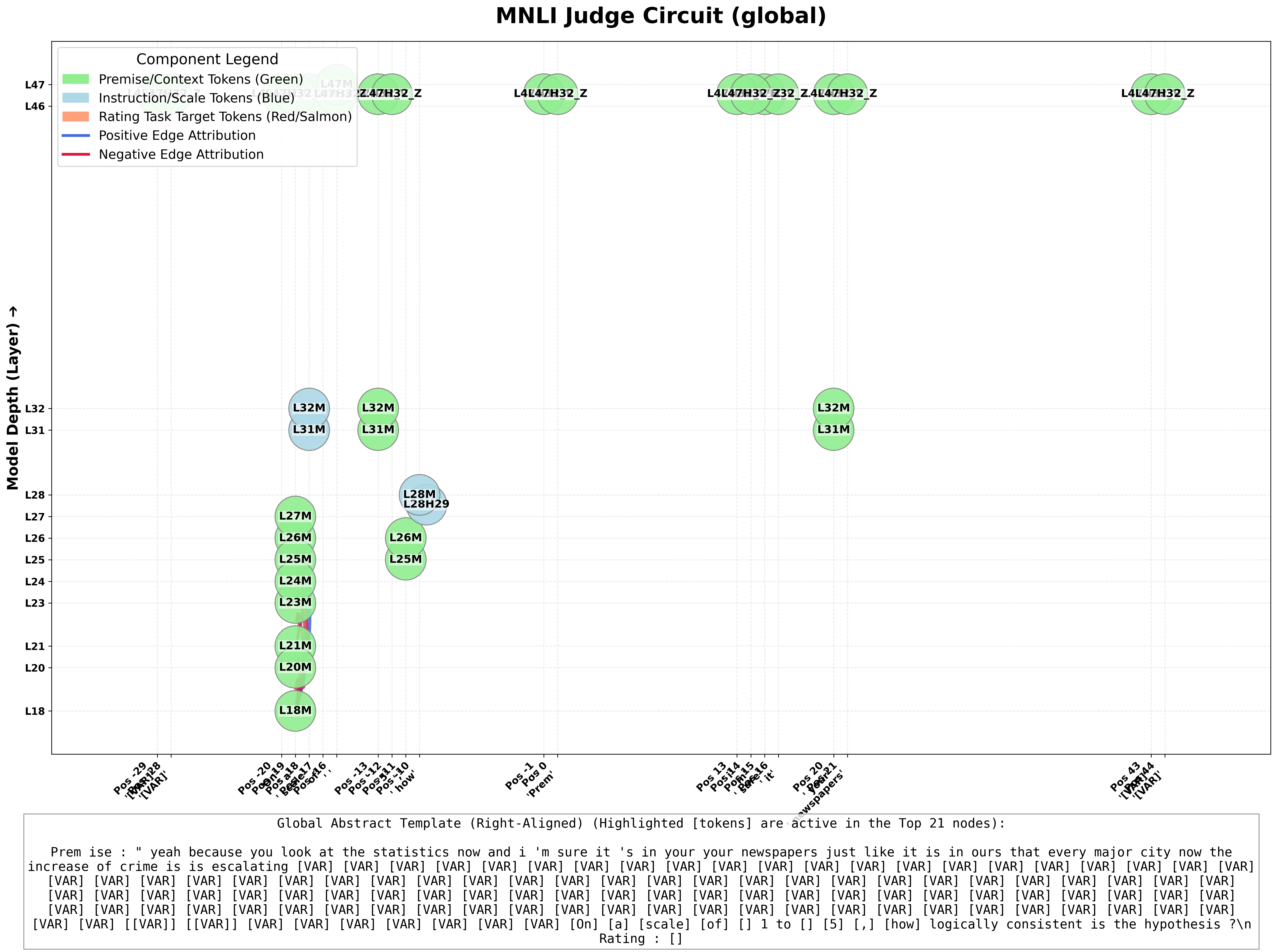}
    \caption{Global judge circuit for \data{MNLI} on Qwen2.5-14B.}
    \label{fig:peap_circuit_mnli_q14b}
\end{figure*}

To complement the structural-overlap and faithfulness summaries in the main body, we visualize the full PEAP-discovered judge circuit across multiple (model, task) pairs (Figures~\ref{fig:peap_circuit}--\ref{fig:peap_circuit_mnli_q14b}). Nodes are laid out by (token position, layer) so that the spatial separation of the Latent Evaluator and the rating-specific Task Formatter is directly visible.
We pair the canonical \data{MNLI} on Gemma-3-27B example with three additional circuits -- \data{CoLA} on the same model, \data{STS-B} on Gemma-3-12B, and \data{MNLI} on Qwen2.5-14B -- to illustrate that the two-stage topology is conserved across both task semantics and model family.
The remaining (\data{RewardBench}, \data{Yelp}) circuits and the unshown model variants are available in the code release and exhibit the same pattern.

Across all four panels, the \textit{Latent Evaluator} sub-circuit corresponds to the green content-token MLP cluster in the middle layers distributed across multiple token positions, while the rating-specific \textit{Task Formatter} corresponds to a concentrated salmon column of late-layer attention heads at the rating token position.
Node coloring encodes \textbf{token-role context} rather than circuit membership: green nodes sit on content tokens (premise/hypothesis or sentence spans), blue nodes on instruction/scale tokens (``scale'', ``how'', ``Sentence''), and salmon nodes on the terminal rating target tokens; edge color encodes PEAP attribution polarity (blue = positive, crimson = negative).
Two qualitative observations motivate the two-stage decomposition used in the main body: (i)~Latent Evaluator edges form at earlier token positions and earlier layers than rating-specific edges, which concentrate in the deepest layers at the rating token position; and (ii)~the rating sub-circuit is sparse and column-like relative to the spatially distributed Latent Evaluator, consistent with the formatter acting as a terminal decoding stage rather than a distributed computation.
Tokens not in the top-$k$ are rendered as \texttt{[VAR]} placeholders in the prompt template footer to avoid privileging any one instance.

\section{Split-Half Circuit Reliability}
\label{app:split_half}

\begin{table}[h!]
    \centering
    \caption{
        Split-half Edge IoU (\%) at top-$100$, mean $\pm$ standard deviation over 10 random partitions. 
        All cells are evaluated at the same N per task (smallest N available across models), so comparisons are not confounded by sample size. 
        Chance baseline is $< 7\%$ across all cells.
    }
    \label{tab:split_half}
    \resizebox{.9\columnwidth}{!}{%
    \begin{tabular}{lccccc}
        \toprule
        \textbf{Model} & \textbf{\data{CoLA}} & \textbf{\data{MNLI}} & \textbf{\data{STS-B}} & \textbf{\data{RB}} & \textbf{\data{Yelp}} \\
        \midrule
        Qwen2.5-7B   & $75.6 \pm 3.7$ & $61.7 \pm 5.1$ & $75.2 \pm 5.0$ & $43.5 \pm 4.7$ & $59.2 \pm 5.5$ \\
        Llama-3.1-8B & $72.4 \pm 3.8$ & --             & $77.1 \pm 4.1$ & --             & --             \\
        Qwen2.5-14B  & $58.0 \pm 5.3$ & $67.7 \pm 3.6$ & $73.8 \pm 7.4$ & $43.3 \pm 4.7$ & $77.5 \pm 2.6$ \\
        Gemma-3-12B  & $34.9 \pm 8.9$ & $64.8 \pm 3.3$ & $77.2 \pm 5.6$ & $25.6 \pm 5.1$ & $22.4 \pm 5.7$ \\
        Gemma-3-27B  & $33.6 \pm 6.3$ & $30.7 \pm 3.4$ & $58.9 \pm 5.1$ & $28.7 \pm 4.6$ & $34.3 \pm 3.5$ \\
        \bottomrule
    \end{tabular}
    }
    \vspace*{-1em}
\end{table}

A recurring concern with circuit-level interpretability is whether circuits discovered on modest sample sizes reflect stable causal structure or idiosyncratic features of the specific instances traced.
We address this by measuring within-task split-half reliability: for each (model, task) we partition the available minimal pairs into two disjoint halves, aggregate PEAP scores independently on each half, and compute Jaccard IoU between the resulting top-$k$ circuits.
We repeat this 10 times with different random partitions and report mean $\pm$ standard deviation.
IoU is computed on structural $(\text{sender}, \text{receiver})$ pairs using the same convention as \S\ref{sec:overlap} (\textit{i.e.}, position-specific edges are ranked first and then collapsed to structural pairs before computing IoU; early reading layers are excluded).
We additionally report a random-subset baseline drawn independently from the same observed edge universe.

Because the available pair counts per (model, task) vary ($N \in \{145, \dots, 500\}$), a naive comparison across cells would confound the reliability signal with statistical power.
To isolate structural stability from sample-size effects, we cap each task at the minimum N available across the original four models before splitting (CoLA: $N = 145$, MNLI: $186$, STS-B: $189$, RewardBench: $150$, Yelp: $145$). Table~\ref{tab:split_half} reports the resulting Edge IoU at $k = 100$.
Random-subset Edge IoU at $k = 100$ ranges between $0.5\%$ and $6.8\%$ across conditions, so all reported reliability values are at least several times above chance.

Four observations are worth emphasizing.
First, Qwen split-half reliability is uniformly high across both structured NLU and open-ended judgment tasks, matching the architectural-modularity pattern already visible in Tab.~\ref{tab:mmlu_modularity}: wherever a model exhibits functional modularity, its extracted circuits are also reliable.

Second, Gemma-3-27B yields \textit{lower} split-half Edge IoU on \data{MNLI} and \data{STS-B} than Gemma-3-12B does, even at matched N. We do not read this as instability.
Rather, it is consistent with a scale-dependent redundancy effect: once the Latent Evaluator is modular (Table~\ref{tab:mmlu_modularity}), the model can route judgment through multiple computationally equivalent sub-pathways, and different data halves select different-but-equivalent subsets of the top-$k$ edges.
The underlying Node IoU remains high on Gemma-3-27B ($66.5\%$ on \data{STS-B}, $65.4\%$ on \data{CoLA} at $k = 100$), indicating that the same set of components is recruited -- just at different attribution ranks within the top 100.

Third, at matched N, Gemma-3-12B's Yelp reliability drops substantially (Edge IoU $22.4\%$ vs. the $46.5\%$ we observe at its native $N = 500$). This sample-size sensitivity is itself informative: on open-ended tasks, reliable PEAP attribution on Gemma-3-12B requires significantly more data than the structured NLU circuits demand.
Qwen-14B, by contrast, maintains strong Yelp reliability ($77.5\%$) at $N = 145$, which matches Qwen's earlier-emergence-of-modularity pattern.

Fourth, the split-half numbers combined with the median MIB faithfulness results (Appendix~\ref{app:faithfulness}) yield a cleaner picture than the previous-draft interpretation.
Among the four models with full open-ended split-half coverage, \data{RewardBench} and \data{Yelp} are above chance on every model, and on three (both Qwens and Gemma-3-27B) the same sparse top-$k$ edge budget that suffices for structured NLU is sufficient to recover open-ended judgment behavior.
Only Gemma-3-12B exhibits reliable-but-unfaithful open-ended circuits (stable split-half IoU but MIB faithfulness near $0$), consistent with its entangled zero-ablation profile in Table~\ref{tab:mmlu_modularity}.
The original concern that open-ended judgment requires a denser circuit than structured NLU therefore reduces to a Gemma-3-12B-specific entanglement effect.

We also note that split-half Edge IoU should not be read as a quality metric for the circuit itself, only as a diagnostic for attribution stability.
Where a model's true circuit is distributed across many partially redundant paths (as we suspect is the case for Gemma-3-27B on \data{MNLI}/\data{STS-B}), a strict top-$k$ edge comparison understates the underlying structural agreement.
Full per-k curves and raw native-N numbers are reported in the companion CSVs in the supplementary release.

\section{Prompt-Paraphrase Robustness}
\label{app:paraphrase_robustness}

A natural concern about any circuit traced on templated prompts is that the recovered structure reflects the exact wording of the template.
To test this, we re-trace circuits with the identical PEAP pipeline on two paraphrase levels of each rating prompt, on Qwen2.5-7B and Gemma-3-12B for \data{CoLA}, \data{STS-B}, and \data{MNLI} (the three structured tasks with the largest minimal-pair sets):
\begin{itemize}[noitemsep,leftmargin=*]
    \item \textbf{PARA1} rewords the instruction while keeping the field labels, e.g., \data{CoLA}'s ``On a scale of 1 to 5, how grammatically acceptable is this sentence?'' becomes ``Rate the grammatical acceptability of this sentence on a scale from 1 to 5.''
    \item \textbf{PARA2} additionally renames the field labels and rephrases the question, e.g., ``Sentence:'' becomes ``Input:'' and \data{MNLI}'s ``Premise:''/``Hypothesis:'' become ``Statement A:''/``Statement B:'', with the instruction ``To what degree does Statement B follow from Statement A?''.
\end{itemize}
Table~\ref{tab:paraphrase_iou} reports the Jaccard IoU (\S\ref{sec:overlap}) between each paraphrased circuit and the original at three sparsity levels.
Overlap under PARA1 is comparable to the within-task split-half ceiling of Appendix~\ref{app:split_half} (e.g., $74.9\%$ edge IoU on Qwen2.5-7B \data{MNLI} at top-$200$, against a $65$--$70\%$ split-half ceiling and a random null below $5\%$), so rewording the instruction changes the circuit about as much as resampling the data does.
Renaming the field labels (PARA2) costs a further $10$--$25$ IoU points; the weakest cell (Qwen2.5-7B \data{STS-B} at top-$50$) falls to roughly the cross-task overlap floor, indicating that the field labels themselves carry a share of the position-anchored structure at high sparsity.
Node IoU stays above $45\%$ in every cell, so the same components are recruited even where the specific edges shift.

\begin{table}[ht]
    \centering
    \caption{
        Circuit overlap between original and paraphrased rating prompts at three top-$k$ thresholds.
        PARA1 rewords the instruction; PARA2 additionally renames the prompt's field labels.
        Reference points from Appendix~\ref{app:split_half}: within-task split-half edge IoU (the effective ceiling) is $65$--$80\%$ on these cells, the random-edge null is $0.7$--$5\%$, and cross-task overlap (the floor for related judgment tasks) is $24$--$49\%$.
    }
    \label{tab:paraphrase_iou}
    \small
    \begin{tabular}{llccc}
        \toprule
        & & \multicolumn{3}{c}{Edge IoU / Node IoU (\%)} \\
        \cmidrule(lr){3-5}
        \textbf{Model / Task} & \textbf{Variant} & top-$50$ & top-$100$ & top-$200$ \\
        \midrule
        \multirow{2}{*}{Qwen2.5-7B \data{CoLA}}  & PARA1 & 62.7 / 62.5 & 67.6 / 75.6 & 73.0 / 81.6 \\
                                                 & PARA2 & 38.8 / 52.7 & 50.0 / 65.5 & 55.7 / 63.2 \\
        \multirow{2}{*}{Qwen2.5-7B \data{STS-B}} & PARA1 & 62.1 / 81.4 & 59.6 / 73.6 & 66.5 / 79.7 \\
                                                 & PARA2 & 17.7 / 45.1 & 31.6 / 50.6 & 42.4 / 55.0 \\
        \multirow{2}{*}{Qwen2.5-7B \data{MNLI}}  & PARA1 & 67.3 / 84.1 & 74.0 / 82.7 & 74.9 / 86.0 \\
                                                 & PARA2 & 40.0 / 48.1 & 48.2 / 55.1 & 52.4 / 62.5 \\
        \midrule
        \multirow{2}{*}{Gemma-3-12B \data{CoLA}}  & PARA1 & 49.2 / 54.8 & 56.2 / 63.4 & 55.0 / 74.3 \\
                                                  & PARA2 & 31.0 / 38.5 & 34.1 / 56.1 & 42.4 / 58.3 \\
        \multirow{2}{*}{Gemma-3-12B \data{STS-B}} & PARA1 & 42.2 / 52.9 & 59.4 / 61.3 & 54.3 / 61.9 \\
                                                  & PARA2 & 42.9 / 48.6 & 45.6 / 60.0 & 49.7 / 58.5 \\
        \multirow{2}{*}{Gemma-3-12B \data{MNLI}}  & PARA1 & 54.2 / 50.9 & 53.6 / 57.7 & 53.5 / 71.6 \\
                                                  & PARA2 & 39.4 / 55.0 & 43.1 / 61.5 & 44.4 / 57.5 \\
        \bottomrule
    \end{tabular}
\end{table}

\section{Pooled-Directional Faithfulness}
\label{app:faithfulness_pooled}

As a sensitivity analysis on the per-instance MIB metric used in Appendix~\ref{app:faithfulness}, we additionally report a magnitude-weighted directional formulation:
\begin{equation*}
    \text{Faith}_{\text{pool}}(k) =
    \frac{\sum_{i=1}^{N} m_i \cdot \left(\text{EV}^{(i)}(\mathcal{C}_k) - \text{EV}^{(i)}_{\text{corr}}\right)}{\sum_{i=1}^{N} \left| \text{EV}^{(i)}_{\text{clean}} - \text{EV}^{(i)}_{\text{corr}} \right|},
\end{equation*}
with $m_i \in \{-1, +1\}$ the per-pair polarity sign.
This pooled formulation has a single aggregate denominator, which causes pairs with large $|\text{EV}_{\text{clean}} - \text{EV}_{\text{corr}}|$ to dominate the recovery score and can yield artifacts such as non-monotonic curves and implausibly high recovery at very small $k$.
On Gemma-3-12B the pooled curve peaks at $1.10$ on \data{MNLI} at $k=5$ (a single edge patching recovering 110\% of the gap is an aggregation artifact) and similarly overshoots at intermediate $k$ on \data{CoLA} and \data{STS-B}, before drifting downward at $k=200$.
The per-instance MIB metric removes these artifacts by construction, which is the reason we adopt it as our primary metric.
The two metrics agree on the qualitative structure-NLU vs.~open-ended-task distinction: both saturate near $1.0$ on \data{CoLA}, \data{MNLI}, and \data{STS-B} for Gemma-3-12B, and both remain below $0.5$ across the full $k$ range for \data{RewardBench} and \data{Yelp}.

\section{Cross-Method Validation via LRPEAP}
\label{app:method_independence}

PEAP belongs to the strongest circuit-localization family in the MIB benchmark \citep{mueller-2025-mib}, and the position-aware extension further improves the faithfulness-per-edge trade-off over position-invariant formulations \citep{haklay-2025-position-aware-automated-circuit-discovery}.
The faithfulness results (Appendix~\ref{app:faithfulness}) address metric fragility; this appendix addresses the complementary non-identifiability concern by comparing PEAP against a second attribution backbone.
On the (Qwen2.5-7B, Gemma-3-12B) $\times$ 10-task panel, top-200 PEAP and LRPEAP edge sets share $42\%$ mean Jaccard IoU on edges and $54\%$ on components (permutation null $p_{99} \leq 5.8\%$); the Latent Evaluator subgraph $\mathcal{C}_{\text{LE}} = \mathcal{C}_{\text{rate}} \cap \mathcal{C}_{\text{class}}$ computed under each method recovers at $0.54$ mean component IoU, peaking at $0.68$ on \data{RewardBench}.
The partial edge-overlap is consistent with computational redundancy, where multiple sparse subgraphs implement the same judgment behavior; the LE/TF decomposition is the structural intersection both methods converge on.

\subsection{Methodology}
\label{app:method_independence_methodology}

LRPEAP retains PEAP's position-aware edge attribution (Appendix~\ref{app:peap_formulas}) and per-pair aggregation but replaces the autograd backward with an LRP-rule backward, using the \texttt{LN-rule} / \texttt{Identity-rule} / \texttt{Half-rule} combination of RelP \citep{jafari-2025-relp}.
All other PEAP machinery -- candidate-edge set, top-$k$ capping, polarity correction $m = \mathrm{sgn}(\mathrm{EV}_{\text{clean}} - \mathrm{EV}_{\text{corr}})$ -- is unchanged, so LRPEAP and PEAP are comparable under our top-$k$ Jaccard IoU and faithfulness metrics.
LRPEAP is not equivalent to RelP itself: RelP's candidate-edge graph is component-level $(n_1, n_2) \in E$, whereas LRPEAP injects RelP's LRP-coefficient backward into PEAP's position-aware formulation.
LRPEAP runs on the same minimal-pair sets as the PEAP experiments (\S\ref{sec:setup_models}); the permutation null for each (model, task, $k$) cell samples $500$ random size-$k$ edge subsets from each method's edge pool and reports the $p_{99}$ Jaccard IoU.

\subsection{Results}
\label{app:method_independence_results}

\begin{table}[h!]
    \centering
    \caption{PEAP vs LRPEAP Jaccard IoU at $K{=}200$ (edge / component); null is the permutation $p_{99}$.}
    \label{tab:lrpeap_iou}
    \small
    \resizebox{.7\linewidth}{!}{%
    \begin{tabular}{lcccc}
    \toprule
    & \multicolumn{2}{c}{Qwen2.5-7B} & \multicolumn{2}{c}{Gemma-3-12B} \\
    \cmidrule(lr){2-3}\cmidrule(lr){4-5}
    Task & Edge & Comp. & Edge & Comp. \\
    \midrule
    \data{CoLA}              & 0.41 & 0.52 & 0.28 & 0.47 \\
    \data{CoLA\_CLASS}       & 0.44 & 0.60 & 0.20 & 0.42 \\
    \data{MNLI}              & 0.37 & 0.51 & 0.56 & 0.58 \\
    \data{MNLI\_CLASS}       & 0.50 & 0.57 & 0.52 & 0.63 \\
    \data{STS-B}             & 0.45 & 0.58 & 0.43 & 0.51 \\
    \data{STS-B\_CLASS}      & 0.44 & 0.60 & 0.38 & 0.58 \\
    \data{Yelp}              & 0.45 & 0.55 & 0.50 & 0.56 \\
    \data{Yelp\_CLASS}       & 0.42 & 0.53 & 0.28 & 0.47 \\
    \data{RewardBench}       & 0.40 & 0.53 & 0.37 & 0.47 \\
    \data{RewardBench\_CLASS}& 0.37 & 0.48 & 0.60 & 0.69 \\
    \midrule
    \textbf{Mean}            & \textbf{0.43} & \textbf{0.55} & \textbf{0.41} & \textbf{0.54} \\
    Null $p_{99}$            & 0.031 & --   & 0.058 & --   \\
    \bottomrule
    \end{tabular}
    }
\end{table}

Table~\ref{tab:lrpeap_iou} reports per-task PEAP$\leftrightarrow$LRPEAP IoU at $K{=}200$: mean edge IoU is $0.43$ on Qwen2.5-7B and $0.41$ on Gemma-3-12B against null $p_{99}$ of $0.031$ and $0.058$, a $4$--$22\times$ enrichment at $K{=}200$, with agreement above the null at every $k \in \{25,\dots,500\}$.
Cross-method agreement is comparable on the two models and uniform across tasks (edge IoU $0.37$--$0.50$ on Qwen2.5-7B).

\begin{table}[h]
    \centering
    \caption{
        Cross-method Latent Evaluator IoU at $K{=}200$ between PEAP's $\mathcal{C}_{\text{LE}}$ and LRPEAP's $\mathcal{C}_{\text{LE}}$, each computed as $\mathcal{C}_{\text{rate}} \cap \mathcal{C}_{\text{class}}$.
    }
    \label{tab:lrpeap_le_iou}
    \small
    \resizebox{.7\linewidth}{!}{%
    \begin{tabular}{lcccc}
        \toprule
        & \multicolumn{2}{c}{Qwen2.5-7B} & \multicolumn{2}{c}{Gemma-3-12B} \\
        \cmidrule(lr){2-3}\cmidrule(lr){4-5}
        Task pair & Edge & Comp. & Edge & Comp. \\
        \midrule
        \data{CoLA} pair          & 0.45 & 0.49 & 0.17 & 0.43 \\
        \data{MNLI} pair          & 0.46 & 0.54 & 0.42 & 0.62 \\
        \data{STS-B} pair         & 0.42 & 0.57 & 0.29 & 0.57 \\
        \data{Yelp} pair          & 0.42 & 0.60 & 0.20 & 0.44 \\
        \data{RewardBench} pair   & 0.31 & 0.49 & 0.42 & 0.68 \\
        \midrule
        \textbf{Mean}             & \textbf{0.41} & \textbf{0.54} & \textbf{0.30} & \textbf{0.55} \\
        \bottomrule
    \end{tabular}
    }
\end{table}

Restricting to the Latent Evaluator (Table~\ref{tab:lrpeap_le_iou}), the LE subgraph is recovered with $0.36$ edge / $0.54$ component IoU on average, peaking at $0.68$ on Gemma-3-12B \data{RewardBench}.

On Gemma-3-12B \data{CoLA}\,$\times$\,\data{CoLA\_CLASS}, LRPEAP's LE at $K{=}200$ includes $31$ distinct attention heads with V$\to$Z edges; L45H3, L46H12, and L47H7 (the three shared-evaluator heads from Appendix~\ref{app:head_analysis}) are all present.

\begin{figure*}[!t]
    \centering
    \includegraphics[width=\linewidth]{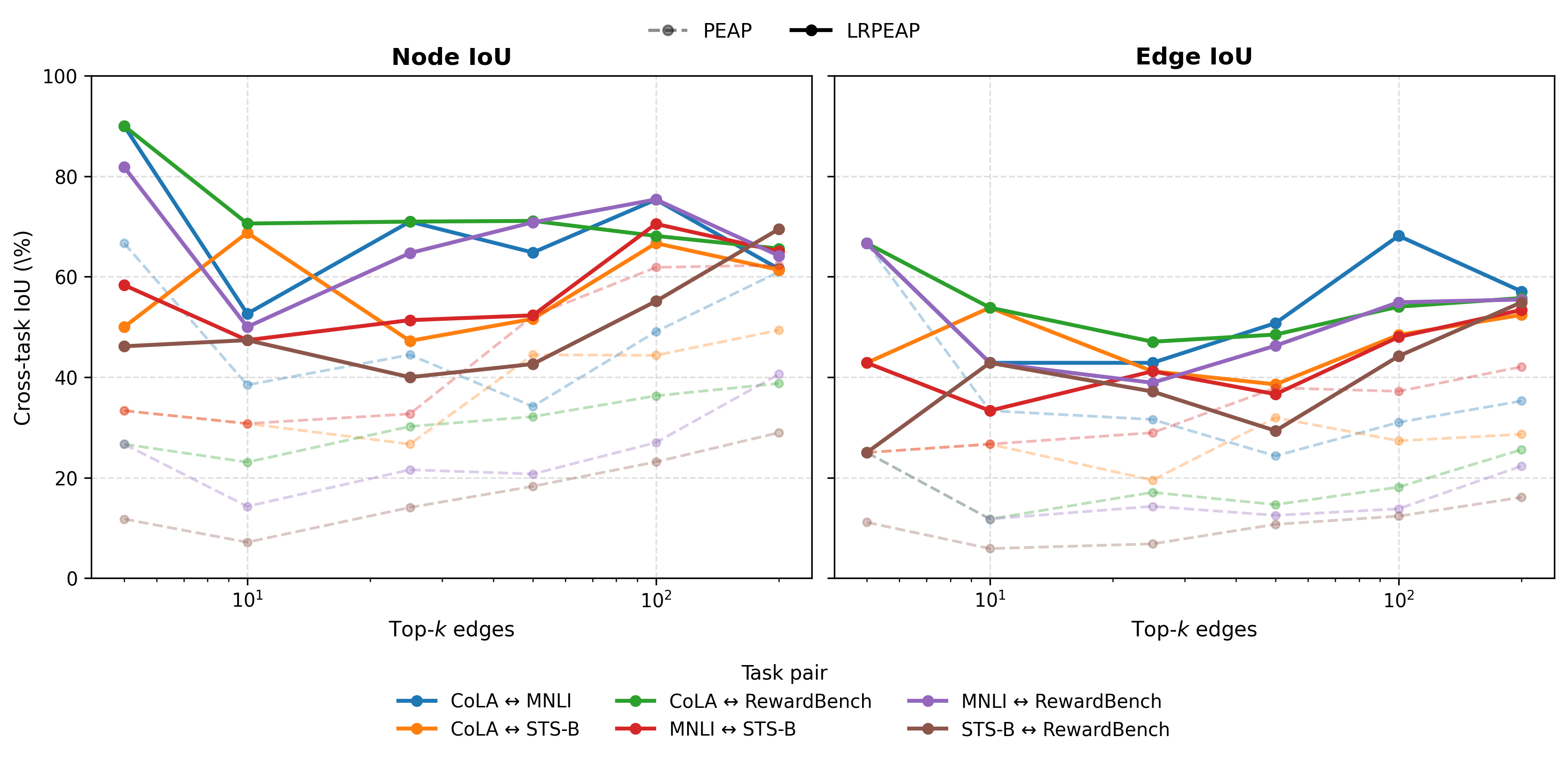}
    \caption{
        Cross-task structural overlap on Gemma-3-12B: LRPEAP (solid) overlaid on PEAP (dashed, faded) at matched task-pair color. Node IoU agrees on the structurally easy pairs (\data{CoLA}\,$\times$\,\data{MNLI}, \data{MNLI}\,$\times$\,\data{STS-B}); Edge IoU is consistently higher under LRPEAP, with both metrics diverging in LRPEAP's favor on pairs involving the open-ended \data{RewardBench} task.
    }
    \label{fig:lrpeap_cross_task}
    \vspace*{-1em}
\end{figure*}

The cross-task shared trunk of Finding~1 also reproduces under LRPEAP (Figure~\ref{fig:lrpeap_cross_task}): Gemma-3-12B Node IoU at top-$200$ is $61.5\%$ / $65.0\%$ / $65.6\%$ for \data{CoLA}\,$\times$\,\data{MNLI} / \data{MNLI}\,$\times$\,\data{STS-B} / \data{CoLA}\,$\times$\,\data{RewardBench}, matching or exceeding the PEAP numbers in \S\ref{sec:overlap} ($61.0\%$ / $62.3\%$ / $48.8\%$).
Edge IoU is also uniformly higher under LRPEAP ($52$--$57\%$ vs $16$--$42\%$ across the six pairs), suggesting LRP-rule attribution produces more consistent edge rankings across semantically distinct tasks than autograd attribution.

\begin{figure*}[t]
    \centering
    \includegraphics[width=\linewidth]{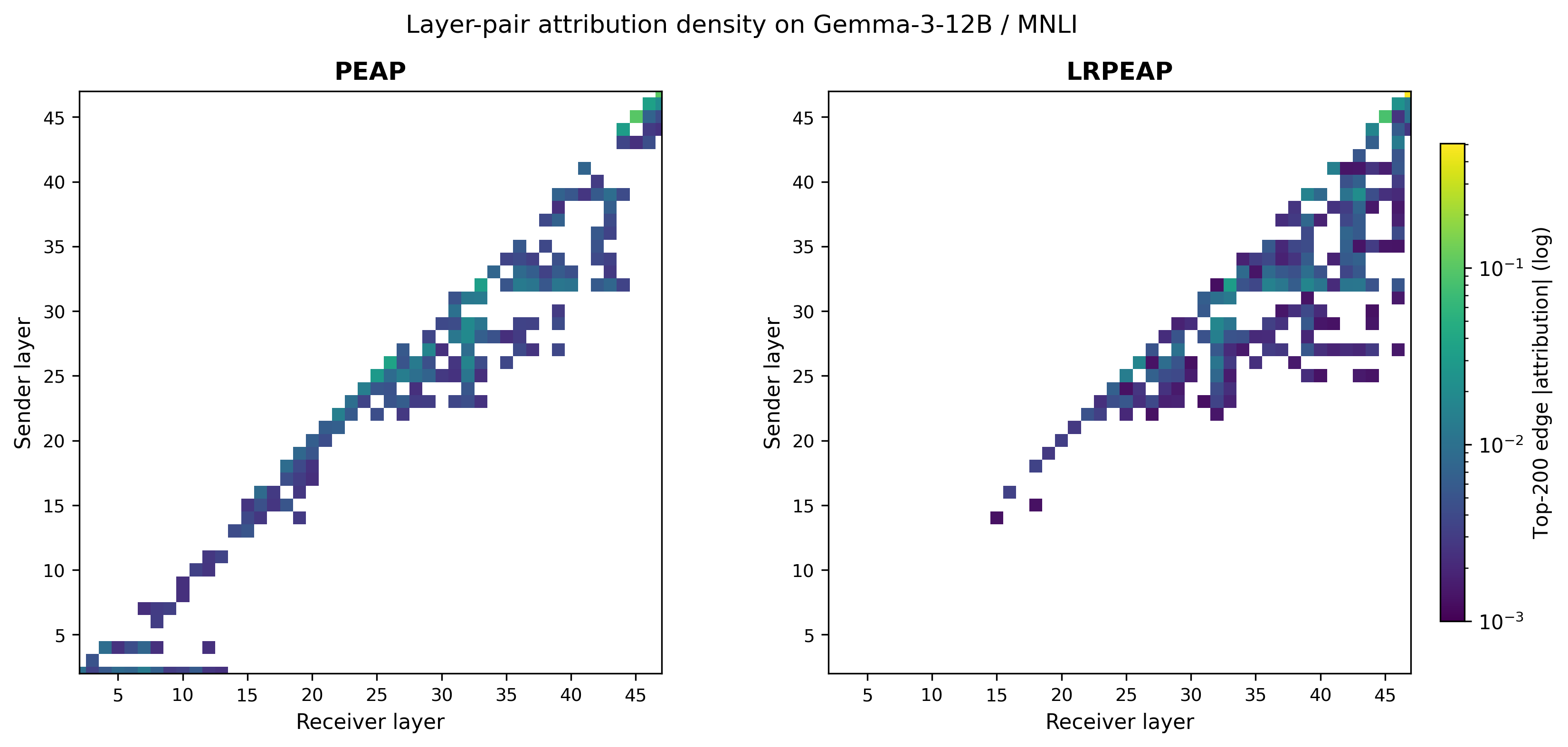}
    \caption{
        Layer-pair attribution density of the top-$200$ edges on Gemma-3-12B \data{MNLI} under PEAP (left) and LRPEAP (right). Both methods concentrate attribution in the same mid-to-late diagonal band, the LE region of \S\ref{sec:contrastive}. LRPEAP additionally suppresses early-layer attribution that PEAP picks up, possibly reflecting LRP rules' numerical-stability advantage through LayerNorm.
    }
    \label{fig:lrpeap_layer_pair}
    \vspace*{-1em}
\end{figure*}

Figure~\ref{fig:lrpeap_layer_pair} confirms architectural agreement: under both methods the top-$200$ \data{MNLI} edges on Gemma-3-12B concentrate in the same mid-to-late diagonal band (layers $\sim$20--47), exactly the LE region (\S\ref{sec:contrastive}); the only visible difference is some early-layer activity (layers $3$--$15$) that PEAP picks up but LRPEAP suppresses.

\begin{figure*}[t]
    \centering
    \includegraphics[width=\linewidth]{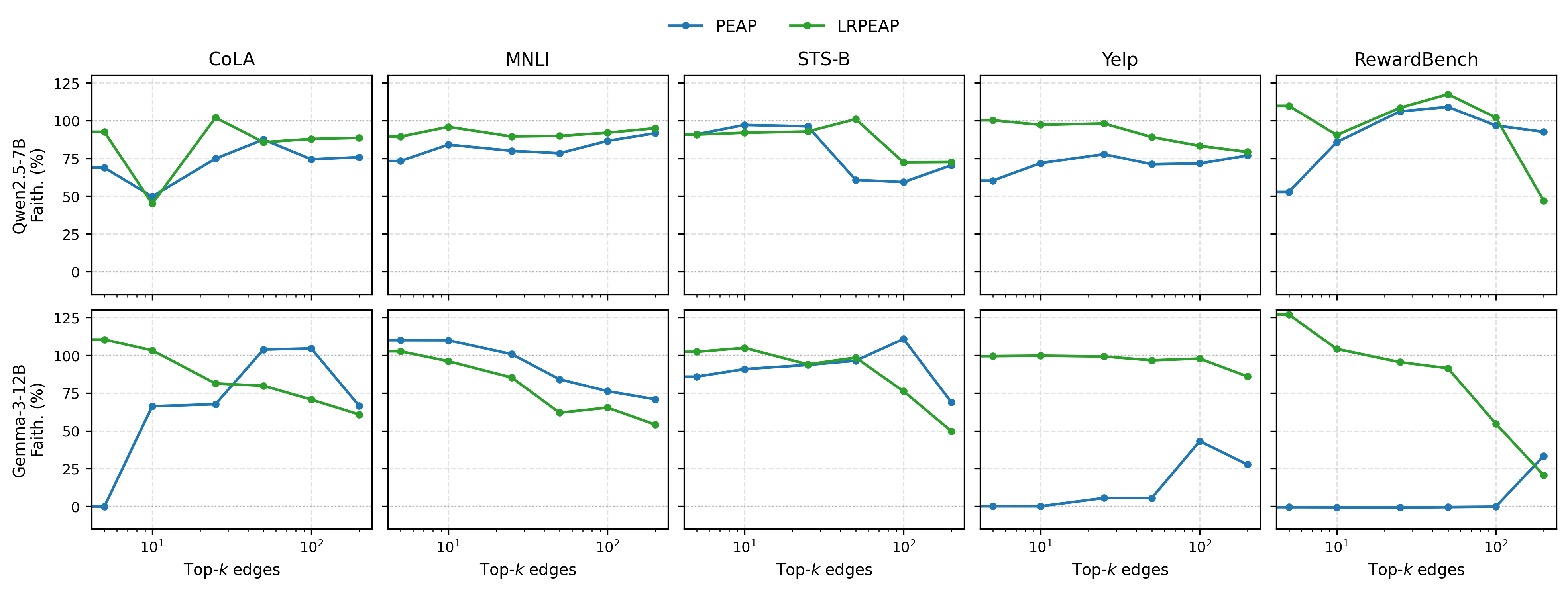}
    \caption{
        Sparse-circuit faithfulness with PEAP (blue) and LRPEAP (green) on Qwen2.5-7B and Gemma-3-12B across the five rating tasks.
    }
    \label{fig:lrpeap_grid}
    \vspace*{-1em}
\end{figure*}

Figure~\ref{fig:lrpeap_grid} overlays PEAP and LRPEAP faithfulness curves on the same panel as Figure~\ref{fig:faithfulness_master}; both backbones saturate at comparable $k$ on every cell where the PEAP circuit saturates.
The five cells where LRPEAP undershoots PEAP at $K{=}200$ (Qwen2.5-7B \data{MNLI\_CLASS} / \data{RewardBench}; Gemma-3-12B \data{CoLA\_CLASS} / \data{MNLI\_CLASS} / \data{STS-B\_CLASS}) all peak at $K \leq 100$ (e.g.~$86\%$, $117\%$, $107\%$ on the three cells that reach saturation).
The $K{=}200$ drop reflects sign-inverted edges entering the LRP ranking far down the tail on tasks with asymmetric output spaces.
There, LRP-rule relevance redistribution does not preserve the per-pair sign that PEAP's symmetric polarity correction (\S\ref{sec:method_patching}) handles by construction.

\subsection{Three-way Comparison Against ACDC}
\label{app:method_independence_acdc}

To test whether the PEAP$\leftrightarrow$LRPEAP overlap is specific to gradient-based attribution or would emerge between any two attribution backbones, we additionally trace ACDC \citep{conmy-2023-acdc} circuits on both Qwen2.5-7B and Gemma-3-12B for \data{MNLI} and \data{MNLI\_CLASS}, scoring all three methods on the same $10$-pair subset.
ACDC's iterative pruning is substantially more compute-intensive than the linear edge-attribution methods, which restricts the comparison to this pair count.

\begin{table}[h]
    \centering
    \caption{
        Three-way pairwise Jaccard IoU at $K{=}200$ between PEAP, LRPEAP, and ACDC on the shared $10$-pair subset, with each cell's enrichment over its permutation $p_{99}$ null in parentheses.
    }
    \label{tab:lrpeap_acdc_3way}
    \resizebox{.7\columnwidth}{!}{%
        \begin{tabular}{llcc}
            \toprule
            Model & Pair & Edge & Comp. \\
            \midrule
            \multirow{3}{*}{Qwen2.5-7B}
             & PEAP $\leftrightarrow$ LRPEAP & \textbf{0.31} (125$\times$) & \textbf{0.55} (5.9$\times$) \\
             & PEAP $\leftrightarrow$ ACDC   & 0.03 (11$\times$)  & 0.16 (1.5$\times$) \\
             & LRPEAP $\leftrightarrow$ ACDC & 0.03 (13$\times$)  & 0.14 (1.4$\times$) \\
            \midrule
            \multirow{3}{*}{Gemma-3-12B}
             & PEAP $\leftrightarrow$ LRPEAP & \textbf{0.21} (85$\times$) & \textbf{0.37} (3.8$\times$) \\
             & PEAP $\leftrightarrow$ ACDC   & 0.02 (7.9$\times$) & 0.31 (2.8$\times$) \\
             & LRPEAP $\leftrightarrow$ ACDC & 0.05 (18$\times$)  & 0.29 (2.9$\times$) \\
            \bottomrule
        \end{tabular}
    }
\end{table}

\begin{figure*}[h]
    \centering
        \includegraphics[width=\linewidth]{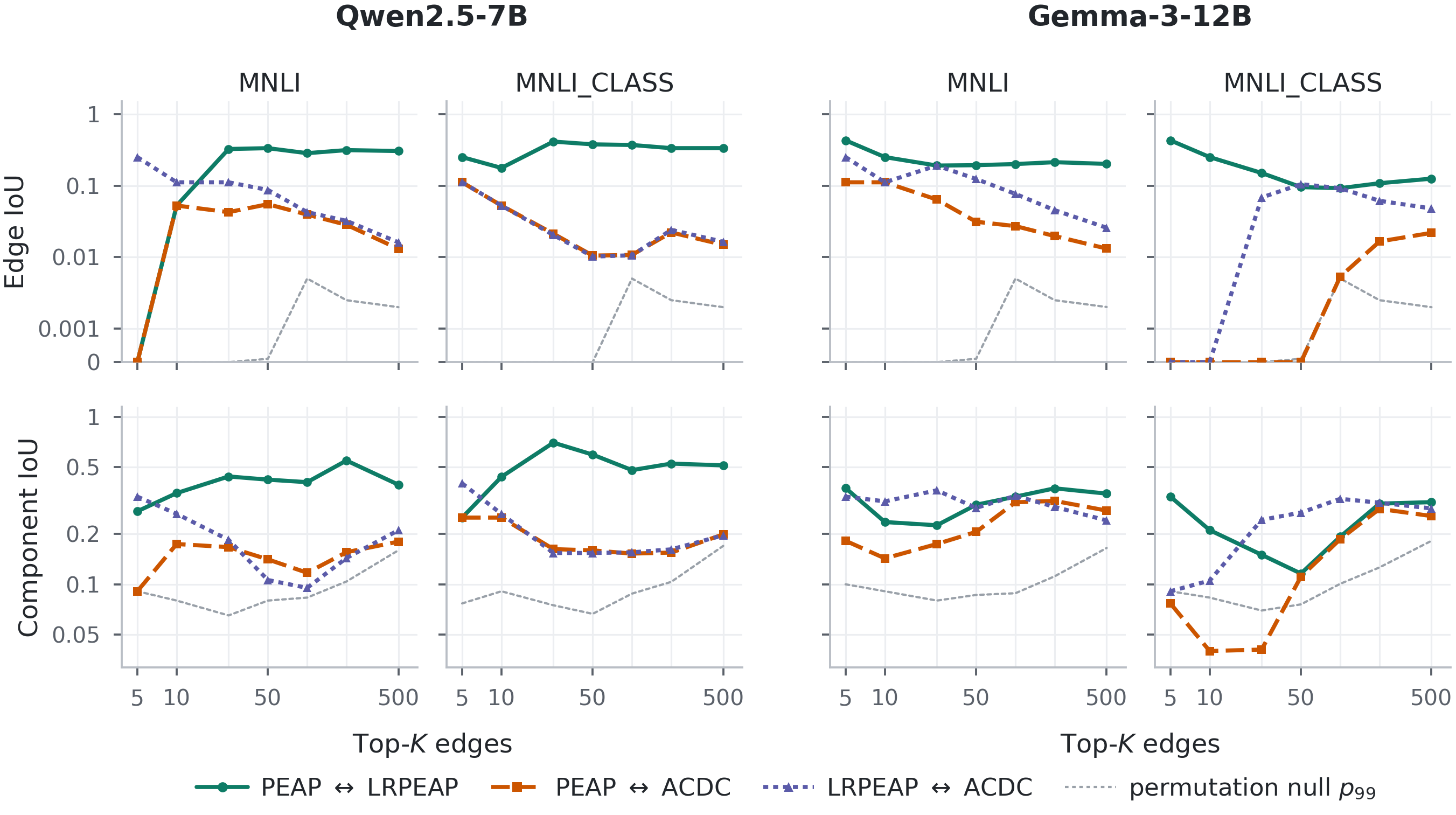}
    \caption{
        Pairwise Jaccard IoU as a function of $K$ for PEAP, LRPEAP, and ACDC on the shared $10$-pair subset, at the edge level (top) and the component level (bottom), for both models and both tasks. The dotted line is the permutation null $p_{99}$, which is exactly zero at small $K$. The edge-level axis is logarithmic above $0.001$ and linear below, so exact zeros are drawn at zero.
    }
    \label{fig:lrpeap_acdc_3way}
    \vspace*{-1em}
\end{figure*}

ACDC's overlap with each gradient method is an order of magnitude below the PEAP$\leftrightarrow$LRPEAP overlap at the edge level, but stays above chance throughout: $6.6$--$24.5\times$ the permutation null across the four cells (Table~\ref{tab:lrpeap_acdc_3way}, Figure~\ref{fig:lrpeap_acdc_3way}).
At the component level the gap narrows; on Gemma-3-12B ACDC agrees with both gradient methods at $0.28$--$0.31$ IoU ($2.2$--$3.0\times$ null) against the gradient-gradient pairing's $0.37$.
The triple intersection at $K{=}200$ is small ($6$--$10$ edges), with the shared components falling in the mid-to-late band that contains the LE (\S\ref{sec:contrastive}).
ACDC's threshold-swept iterative pruning lands on a structurally distinct sparse subgraph from either gradient-based method, which is expected given the algorithmic difference:
ACDC tests each candidate edge with its own forward pass and prunes it only if the resulting metric drop falls below a threshold $\tau$, whereas PEAP and LRPEAP score every edge from a single backward pass via the first-order gradient of the metric.

\section{Layer-Wise Decomposition of Edge IoU}
\label{app:layerwise_iou}

This appendix decomposes the Edge IoU of \S\ref{sec:overlap} along the (sender layer, receiver layer) grid and along network depth, examining where in the model the structural overlap the discovered circuits overlap.
 
\paragraph{Layer assignment for the Edge IoU.}
Each edge contributes one tagged item to its sender's layer bucket and one to its receiver's layer bucket; every edge therefore contributes exactly two layer participations, regardless of whether its endpoints share a layer. Formally, let
\vspace{-0.5em}
\begin{equation*}
\mathcal{B}_A[L] = \{ (s, r) \in \mathcal{E}_A : L \in \{\ell(s), \ell(r)\} \}
\end{equation*}
\vspace{-1.5em}
 
be the bucket of layer $L$ for task $A$, where $\ell(\cdot)$ denotes the layer of a component. 
Per-layer Edge IoU is then computed inside each layer using the same set definition as \S\ref{sec:overlap}:
\vspace{-0.5em}
\begin{equation*}
\mathrm{IoU}_{\mathrm{edge}}^{L}(\mathcal{E}_A, \mathcal{E}_B) = \mathrm{IoU}\bigl(\mathcal{B}_A[L], \mathcal{B}_B[L]\bigr).
\end{equation*}
\vspace{-1.5em}
 
For the (sender, receiver) decomposition we use the joint bucket
\vspace{-0.5em}
\begin{equation*}
\resizebox{.7\columnwidth}{!}{$
\mathcal{B}_A[(L_s, L_r)] = \{ (s, r) \in \mathcal{E}_A : \ell(s) = L_s,\ \ell(r) = L_r \},
$}
\end{equation*}
\vspace{-1.5em}
 
and per-cell Edge IoU is
\vspace{-0.5em}
\begin{equation*}
\resizebox{.7\columnwidth}{!}{$
\mathrm{IoU}_{\mathrm{edge}}^{(L_s, L_r)}(\mathcal{E}_A, \mathcal{E}_B) = \mathrm{IoU}\bigl(\mathcal{B}_A[(L_s, L_r)], \mathcal{B}_B[(L_s, L_r)]\bigr).
$}
\end{equation*}

\begin{figure*}[t]
    \centering
    \includegraphics[width=\linewidth]{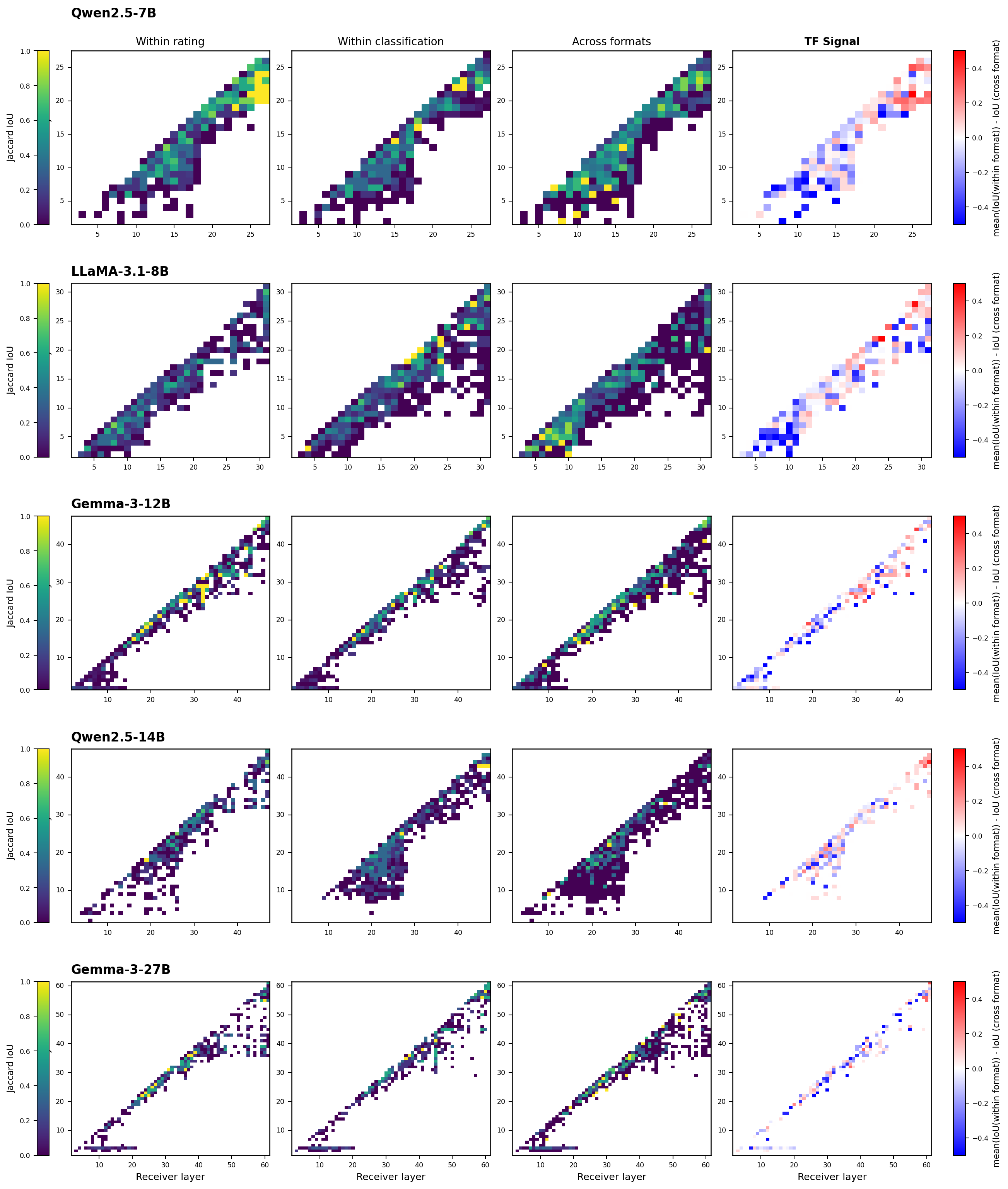}
    \caption{
        Per-model (sender, receiver) Edge IoU grids at top-$200$, one row per model. 
        The first three columns show mean Jaccard IoU within rating pairs, within classification pairs, and across tasks; the fourth column shows the TF signal (App.~\ref{app:layerwise_iou}). 
        Positive (red) cells in the TF signal panel localize edges shared by same-format pairs but absent from the matched cross-format pair, isolating the TF contribution.
    }
    \label{fig:le-tf-grid}
    \vspace*{-1em}
\end{figure*}

\paragraph{TF-attributable overlap}
Same-format pairs share both the LE and the TF; cross-format pairs of the same dataset share only the LE (\S\ref{sec:contrastive}). 
The cell-wise difference
\begin{equation*}
\Delta\mathrm{IoU}^{(L_s, L_r)} = \overline{\mathrm{IoU}^{(L_s, L_r)}_{\text{within-format}}} - \mathrm{IoU}^{(L_s, L_r)}_{\text{cross-format}}
\end{equation*}
isolates the contribution of the TF to the structural overlap. 
Within-format pairs are constructed pairwise from \data{CoLA}, \data{STS-B}, \data{MNLI}, and \data{RewardBench} and their classification variants; cross-format pairs match each dataset against its own counterpart. 
By construction, positive cells in the resulting heatmap (Figure~\ref{fig:le-tf-grid}) localize edges that two same-format circuits share but the matched cross-format pair does not, i.e., edges attributable to the TF.

Across the models, $\Delta\mathrm{IoU}^{(L_s, L_r)}$ concentrates on mid-to-late layers around the diagonal band, consistent with the LE/TF decomposition of \S\ref{sec:contrastive}. 
On the larger modular models (\lm{Qwen2.5-14B}, \lm{Gemma-3-27B}), the dispersion narrows and positive mass falls on a smaller set of (sender, receiver) pairs, consistent with the scale-dependent redundancy effect documented in Appendix~\ref{app:split_half}.

\begin{figure*}[t]
    \centering
    \includegraphics[width=\linewidth]{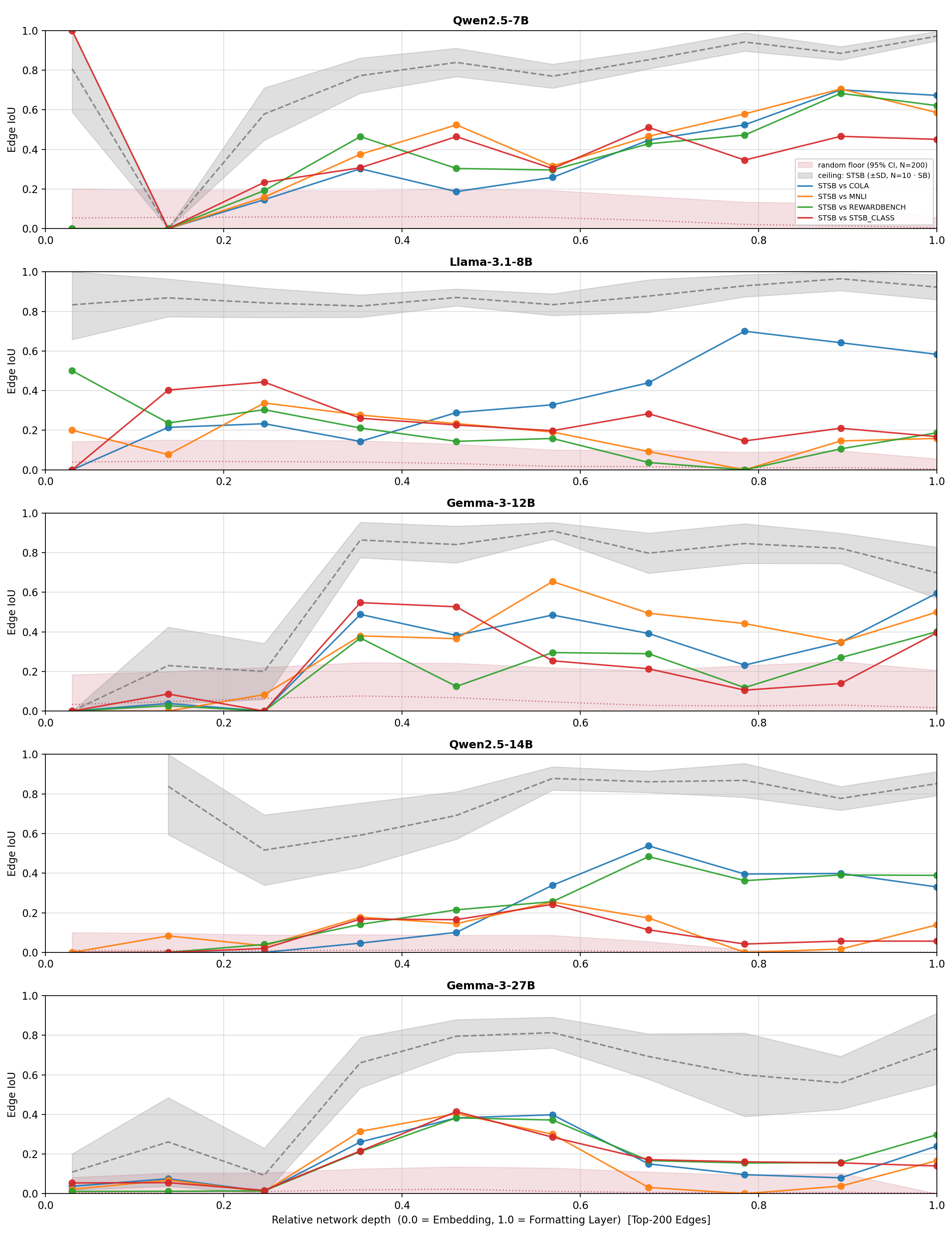}
    \caption{
        Per-layer Edge IoU at $K = 200$ between \data{STS-B} and every other task across the evaluated models, binned into 10 relative-depth bins. 
        The lower band is the random-edge baseline ($95\%$ CI over $200$ trials); the upper band is the within-task split-half ceiling. 
        The ceiling is undefined for \lm{Qwen2.5-14B} at the leftmost bin because no top-$K$ edge lands there in any partition.
    }
    \label{fig:layerwise-crosstask}
    \vspace*{-1em}
\end{figure*}

\paragraph{Cross-task layer-wise overlap.}
Figure~\ref{fig:layerwise-crosstask} reports per-layer Edge IoU at $K = 200$ between \data{STS-B} and every other task (\data{CoLA}, \data{MNLI}, \data{RewardBench}, and its own cross-format variant \data{STS-B\_CLASS}). 
Layers are binned into 10 relative-depth bins for cross-model comparison.

Two reference bands frame the cross-task lines. 
The lower band is the random-edge baseline (App.~\ref{app:split_half}): 
edges sampled uniformly from each task's full PEAP edge set, bucketed identically, with $95\%$ CI over $200$ trials. 
The upper band is the within-task split-half ceiling (App.~\ref{app:split_half}), extended per-layer with the same bucketing; 
it reports mean $\pm$ standard deviation across the 10 splits, Spearman--Brown corrected to project full-$N$ reliability.

Across the models, the rating-format cross-task curves (\data{STS-B} vs.\ \data{CoLA}, \data{MNLI}, \data{RewardBench}) rise above the random floor by the mid layers and remain elevated through the final layers, locating the shared-trunk overlap on the depth axis. 
The matched-dataset cross-format curve (\data{STS-B} vs.\ \data{STS-B\_CLASS}) tracks the same trajectory through the mid layers but ranks last among the four cross-task lines in the final layers.

\section{Knowledge-Probe Controls, Recall Reference Circuits, and LE Threshold Stability}
\label{app:knowledge_probes}

This appendix reports the full data behind the entanglement decomposition of \S\ref{sec:modularity_zero_ablation}: a heads-only ablation panel, a pure-recall cloze probe, factual-recall reference-circuit containment, and the Latent Evaluator's stability across the top-$k$ threshold.

\subsection{Heads-Only Ablation Panel}
\label{app:heads_only_ablation}

Table~\ref{tab:heads_only} isolates the contribution of the LE attention heads by ablating only the heads (leaving all LE MLPs intact), with a matched random-heads control.
On Gemma-3-12B, the three LE heads alone account for the \data{MMLU} verification damage ($14$--$18$ pp across subsets) while \data{StrategyQA} and \data{CREAK} stay within noise of the control; on Qwen2.5-7B the single LE head is inert on every probe.
Llama-3.1-8B, Qwen2.5-14B, and Gemma-3-27B have no attention heads in their merged LE, so their heads-only cells are vacuous by construction and are omitted.

\begin{table}[t]
    \centering
    \caption{
        Heads-only zero-ablation (\%) with a size-matched random-heads control (3 trials).
        Bold marks cells where damage exceeds the control.
        Gemma-3-12B's verification damage in Table~\ref{tab:mmlu_modularity} is carried by its three LE heads; models whose LE contains no heads are omitted (vacuous).
    }
    \label{tab:heads_only}
    \resizebox{.8\linewidth}{!}{%
    \begin{tabular}{llccc}
        \toprule
        \textbf{Model} & \textbf{Probe} & \textbf{Base} & \textbf{Heads abl.} & \textbf{Random heads} \\
        \midrule
        \multirow{5}{*}{\shortstack[l]{Gemma-3-12B\\(3 LE heads)}}
        & Clinical DB & 79.0 & \textbf{63.0} & 80.0 $\pm$ 0.8 \\
        & Abstract Alg. & 47.0 & \textbf{29.0} & 48.1 $\pm$ 3.0 \\
        & Physics & 52.0 & \textbf{38.0} & 51.7 $\pm$ 0.5 \\
        & \data{StrategyQA} & 65.0 & 64.0 & 64.8 $\pm$ 0.2 \\
        & \data{CREAK} & 80.5 & 78.0 & 80.8 $\pm$ 0.8 \\
        \midrule
        \multirow{5}{*}{\shortstack[l]{Qwen2.5-7B\\(1 LE head)}}
        & Clinical DB & 78.0 & 77.0 & 77.3 $\pm$ 0.9 \\
        & Abstract Alg. & 56.0 & 54.0 & 55.3 $\pm$ 0.9 \\
        & Physics & 53.0 & 55.0 & 51.7 $\pm$ 1.2 \\
        & \data{StrategyQA} & 66.5 & 66.0 & 65.2 $\pm$ 1.2 \\
        & \data{CREAK} & 81.5 & 82.0 & 81.8 $\pm$ 1.0 \\
        \bottomrule
    \end{tabular}
    }
    \vspace*{-1em}
\end{table}

\subsection{Pure-Recall Cloze Probe (PopQA)}
\label{app:cloze_probe}

The verification probes of Table~\ref{tab:mmlu_modularity} contain an evaluative step (deciding which answer is more correct), so damage to them cannot by itself distinguish knowledge entanglement from the probes recruiting judgment machinery.
We therefore run a declarative cloze completion on PopQA (e.g., \textit{``The capital of \{X\} is \_\_\_''}; no candidate answers, no verification step; LAMA-style first-token P@1; $N = 300$) under LE zero-ablation vs.\ the same size/type-matched random-ablation control (Table~\ref{tab:cloze_probe}).
The probe decomposes the Gemma-3-12B result of \S\ref{sec:modularity_zero_ablation}: full-LE ablation drops pure recall $13$ pp below the control (MLP-carried recall entanglement, also present at $27$B with $16$ pp), while ablating only the three LE heads -- which suffice to collapse \data{MMLU} (Table~\ref{tab:heads_only}) -- leaves pure recall untouched.
On Qwen2.5-7B/14B and Llama-3.1-8B, LE ablation never exceeds the control: recall does not route through their LEs.

\begin{table}[t]
    \centering
    \caption{
        Declarative PopQA cloze accuracy (first-token P@1, \%, $N = 300$) under LE zero-ablation vs.\ a size/type-matched random-ablation control (3 trials).
        Bold marks LE-ablation damage exceeding the control: an MLP-carried recall entanglement specific to the Gemma-3 family.
        Heads-only rows are omitted for the models whose LE contains no attention heads (vacuous).
    }
    \label{tab:cloze_probe}
    \resizebox{.8\linewidth}{!}{%
    \begin{tabular}{llccc}
        \toprule
        \textbf{Model} & \textbf{Ablated} & \textbf{Base} & \textbf{LE-abl.} & \textbf{Random ctrl.} \\
        \midrule
        Qwen2.5-7B & 1H+9M & 44.7 & 4.7 & 3.4 $\pm$ 1.3 \\
        Qwen2.5-7B & heads only (1H) & 44.7 & 38.7 & 41.2 $\pm$ 0.6 \\
        Llama-3.1-8B & 0H+8M & 62.7 & 5.7 & 11.5 $\pm$ 9.2 \\
        Gemma-3-12B & 3H+4M & 60.3 & \textbf{49.7} & 62.9 $\pm$ 4.5 \\
        Gemma-3-12B & heads only (3H) & 60.3 & 60.0 & 60.8 $\pm$ 0.4 \\
        Qwen2.5-14B & 0H+2M & 62.3 & 54.0 & 53.9 $\pm$ 16.9 \\
        Gemma-3-27B & 0H+3M & 59.3 & \textbf{46.3} & 62.3 $\pm$ 3.6 \\
        \bottomrule
    \end{tabular}
    }
    \vspace*{-1em}
\end{table}

\subsection{Factual-Recall Reference Circuits}
\label{app:recall_circuits}

As a mechanism-level criterion separating retrieval from judgment, we trace factual-recall circuits on \data{CREAK} (claim verification) and \data{StrategyQA} (implicit-reasoning QA) with the identical PEAP pipeline, at both $N = 100$ and $N = 200$ minimal pairs.
These circuits double as the non-judgment external reference whose absence \S\ref{sec:limitations} previously flagged.
For each recall circuit we measure the fraction of the merged Latent Evaluator (the intersection of the \data{CoLA}/\data{MNLI}/\data{STS-B} per-task LEs) that it recruits (Table~\ref{tab:recall_containment}).
Claim verification routes through the LE far more on the entangled Gemma-3-12B than on the modular Qwen2.5-7B at every threshold and sample size, a graded contrast consistent with the cloze-probe dissociation of Appendix~\ref{app:cloze_probe}; the permutation null for size-matched random subsets is $\approx 5\%$ throughout.
\data{StrategyQA}'s containment is high on \textit{both} models, consistent with implicit-reasoning QA itself recruiting evaluative machinery -- which is exactly why the cloze probe, with no verification step, is the discriminating control.

\begin{table}[t]
    \centering
    \caption{
        Merged-LE containment (\%): the share of the merged Latent Evaluator's components recruited by each factual-recall reference circuit, at two recall-circuit sample sizes ($N$ pairs) and two pruning thresholds (top-$k$ edges).
        The claim-verification contrast (bold vs.\ the Qwen column) is preserved at every threshold and sample size; permutation null $\approx 5\%$.
    }
    \label{tab:recall_containment}
    \resizebox{.8\linewidth}{!}{%
    \begin{tabular}{llcc|cc}
        \toprule
        & & \multicolumn{2}{c|}{\textbf{Gemma-3-12B}} & \multicolumn{2}{c}{\textbf{Qwen2.5-7B}} \\
        \textbf{Recall circuit} & \textbf{$N$} & top-100 & top-200 & top-100 & top-200 \\
        \midrule
        \multirow{2}{*}{\data{CREAK}}
        & 100 & 82.4 & \textbf{91.9} & 36.0 & 48.7 \\
        & 200 & 70.6 & \textbf{78.4} & 40.0 & 64.1 \\
        \midrule
        \multirow{2}{*}{\data{StrategyQA}}
        & 100 & 52.9 & 75.7 & 76.0 & 82.1 \\
        & 200 & 94.1 & 75.7 & 88.0 & 89.7 \\
        \bottomrule
    \end{tabular}
    }
    \vspace*{-1em}
\end{table}

\subsection{LE Stability Across the Top-$k$ Threshold}
\label{app:le_stability}

Because the Latent Evaluator is defined as an intersection of circuits pruned at a fixed top-$k$, we verify that its identity is not an artifact of the chosen threshold.
Computing $\mathcal{C}_{\text{LE}}(k)$ for $k \in \{50, 100, 200, 500, 1000\}$ from the cached attribution scores, the LE is \textit{strictly nested}: $\mathcal{C}_{\text{LE}}(k) \subseteq \mathcal{C}_{\text{LE}}(1000)$ with $100\%$ containment at every $k$, for all of \data{CoLA}/\data{MNLI}/\data{STS-B} on both Gemma-3-12B and Qwen2.5-7B (Table~\ref{tab:le_stability}).
Varying $k$ grows the intersection but never swaps out components, so the \textit{identity} of the shared evaluator is threshold-stable (raw IoU across $k$ differs only because set sizes differ), and the named shared-evaluator heads (L45H3, L46H12, L47H7) are present at every $k$ from $50$ to $1000$ on all three tasks.
Our operating point $k = 200$ is where faithfulness saturates for $21$ of $25$ cells (Figure~\ref{fig:faithfulness_master}); the top-$1000$ instance in Figure~\ref{fig:worked_example} is a visualization superset of the same nested hierarchy.

\begin{table}[t]
    \centering
    \caption{
        Latent Evaluator size $|\mathcal{C}_{\text{LE}}(k)|$ (components) across pruning thresholds.
        Every set is fully contained in the next larger one ($100\%$ nesting along each row), and the named shared-evaluator heads appear at every $k$ on all Gemma-3-12B tasks.
    }
    \label{tab:le_stability}
    \resizebox{.8\linewidth}{!}{%
    \begin{tabular}{llccccc}
        \toprule
        \textbf{Model} & \textbf{Task} & $k{=}50$ & $k{=}100$ & $k{=}200$ & $k{=}500$ & $k{=}1000$ \\
        \midrule
        \multirow{3}{*}{Gemma-3-12B}
        & \data{CoLA} & 28 & 54 & 77 & 124 & 177 \\
        & \data{MNLI} & 34 & 46 & 70 & 107 & 151 \\
        & \data{STS-B} & 39 & 61 & 88 & 137 & 201 \\
        \midrule
        \multirow{3}{*}{Qwen2.5-7B}
        & \data{CoLA} & 23 & 43 & 77 & 143 & 222 \\
        & \data{MNLI} & 23 & 35 & 55 & 122 & 209 \\
        & \data{STS-B} & 24 & 45 & 82 & 140 & 227 \\
        \bottomrule
    \end{tabular}
    }
    \vspace*{-1em}
\end{table}

\begin{table*}[t]
    \centering
    \caption{STS-B circuit attention head functional roles derived from per-head PEAP attribution analysis. Heads are grouped by role and sorted by layer within each group. Heads with multiple roles (L20H9, L21H15, L24H0, L25H12, L26H0, L26H1) appear in their primary role group.}
    \label{tab:stsb_heads}
    \small
    \begin{tabular}{lp{\dimexpr\linewidth-2cm}}
    \toprule
    \textbf{Head} & \textbf{Description} \\
    \midrule
    \multicolumn{2}{l}{\textit{Shared Evaluators}} \\
    \midrule
    L20H9 & Contrastive token detector; attends to semantically differentiating tokens in the second sentence and simultaneously distributes the resulting comparison signal into all three downstream pathways (shared evaluation, rating-specific, and classification-specific) \\
    L24H0 & Routes within the late instruction span, consolidating information across the question template tokens without forwarding to the Rating position \\
    L25H8 & Routes from the sentence-final question mark toward the output boundary region, propagating the consolidated judgment signal rather than contributing directly to the Rating token \\
    L25H12 & Aggregates from both the second sentence and the full instruction span toward the Rating position, serving as a key bridge between the second sentence representation and the final output \\
    L26H0 & Routes within the instruction span toward the late boundary region around the question mark, functioning as a straightforward instruction-level consolidator \\
    L26H1 & Routes from both the late instruction span and the second sentence toward the output boundary region spanning the question mark, Rating token, and colon \\
    L44H8 & Routes within the late instruction span toward the output boundary with minimal variation \\
    L45H3 & Consistently routes from the end of the instruction span to the Rating position; exhibits the same terminal forwarding behavior in both CoLA and STS-B circuits \\
    L46H12 & Mirrors L45H3 in consistently routing from the late instruction span to the Rating position across both tasks \\
    L47H7 & Format-agnostic terminal router, consistently forwarding from the late instruction region to the Rating position regardless of task or assigned role \\
    \midrule
    \multicolumn{2}{l}{\textit{Rating Formatters}} \\
    \midrule
    L18H10 & Routes from the second sentence span toward mid-instruction positions, with occasional local intra-sentence activity \\
    L19H0 & Routes primarily from semantically relevant tokens within the second sentence toward the instruction span, with very occasional contributions from the first sentence \\
    L21H15 & Routes between neighboring tokens within the second sentence while also forwarding toward the instruction boundary, performing local semantic integration within the second sentence \\
    L23H3 & Routes from the second sentence span toward the instruction region while also routing within the instruction span toward its later parts \\
    L26H5 & Routes within the mid-instruction span toward the late instruction region with minimal attribution \\
    L27H5 & Routes from the instruction span toward the end of the instruction and the start of the output boundary region \\
    L31H13 & Routes from the late instruction span toward the output boundary region spanning the question mark and Rating token \\
    \midrule
    \multicolumn{2}{l}{\textit{Class Formatters}} \\
    \midrule
    L20H2 & Routes from both the first sentence content tokens and the instruction span toward the "sentences" boundary token, consolidating semantic content at the final instruction boundary \\
    L21H15 & Routes between neighboring tokens within the second sentence while forwarding toward the instruction boundary; dual role reflects format-agnostic intra-sentence integration \\
    L24H13 & Routes within the late instruction span toward the output boundary, with occasional contributions from the second sentence \\
    L24H15 & Distributes signal across various instruction tokens without a strong directional preference toward the output \\
    L26H0 & Routes within the instruction span toward the late boundary region; see shared evaluator entry \\
    L26H10 & Routes from the mid-instruction span toward the output position with predominantly negative attribution \\
    L27H8 & Routes from the end of the instruction span toward the Rating position with consistently negative attribution, functioning as an inhibitory terminal head \\
    L27H9 & Routes within the mid-to-late instruction span with minimal attribution \\
    L43H3 & Selectively picks up an accumulated signal from a specific second sentence position alongside instruction tokens, rarely active \\
    L47H6 & Inhibitory terminal head applying consistently negative attribution at the Rating position, dampening competing format signals at the output boundary \\
    \bottomrule
    \end{tabular}
\end{table*}

\end{document}